\documentclass{article} 
\usepackage{iclr2027_conference,times}

\usepackage{amsmath,amsfonts,bm}

\def\eqref#1{equation~\ref{#1}}

\def\1{\bm{1}}

\DeclareMathAlphabet{\mathsfit}{\encodingdefault}{\sfdefault}{m}{sl}
\SetMathAlphabet{\mathsfit}{bold}{\encodingdefault}{\sfdefault}{bx}{n}

\usepackage{hyperref}
\usepackage{url}

\usepackage{amssymb}
\usepackage{amsmath}
\usepackage{booktabs}
\usepackage{multirow}
\usepackage{wrapfig}
\usepackage{enumitem}

\usepackage[table]{xcolor}
\usepackage{caption}

\usepackage{graphicx}
\usepackage{pifont}

\usepackage{titletoc}

\newcommand{\cmark}{\ding{51}}

\title{Demiurge: Forging Navigable 4D Worlds via Autoregressive Video Rerendering}

\author{\textbf{Hanxin Zhu}$^{1,2*}$,\quad
\textbf{Jianqi Ma}$^{2*}$,\quad
\textbf{Cong Wang}$^{3}$,\quad
\textbf{Zhihao Hu}$^{2}$,\quad
\textbf{Junyao Gao}$^{4}$,\\[3pt]
\textbf{Tianyu He}$^{5}$,\quad
\textbf{Junyan Lin}$^{6}$,\quad
\textbf{Jiayin Xiao}$^{1}$,\quad
\textbf{Jiateng Shou}$^{1}$,\\[3pt]
\textbf{Sen Liang}$^{1}$,\quad
\textbf{Weidong Zhang}$^{2}$,\quad
\textbf{Zhibo Chen}$^{1\dagger}$,\quad
\textbf{Jun Zhang}$^{2\dagger}$\\[8pt]
$^{1}$USTC \quad
$^{2}$Tencent \quad
$^{3}$CASIA \quad
$^{4}$Tongji University \quad
$^{5}$Microsoft \quad
$^{6}$PolyU\\[3pt]
$^{*}$Equal contribution\qquad
$^{\dagger}$Corresponding authors
}

\iclrfinalcopy

\begin{document}

\maketitle
\lhead{Preprint}

\begin{abstract}
Recent advancements in video diffusion models have substantially empowered generative video rerendering, where novel-view videos can be synthesized from a single input video, facilitating a broad spectrum of applications such as world modeling, embodied intelligence, and robotic simulation. However, despite their impressive photorealism, existing video rerendering methods typically treat each camera trajectory in isolation, lacking the cross-trajectory consistency required to form a coherent and traversable 4D world. To address this limitation, we introduce \textbf{\textit{Demiurge}}, a novel autoregressive video rerendering pipeline for navigable 4D world generation. Built upon a dynamic 4D point-cloud memory cache, Demiurge promotes cross-trajectory consistency through \textbf{four key components:} \textbf{1)} an appearance video diffusion model that rerenders photorealistic novel-view videos along target camera trajectories conditioned on the 4D memory; \textbf{2)} a memory-consistent video depth completion model that aligns known-region depths with cached geometry while completing missing regions, mitigating scale ambiguity and discontinuities during memory updates; \textbf{3)} a memory-aware dual-stream warping strategy that provides high-fidelity visual guidance and occlusion cues for plausible novel-view video synthesis; \textbf{4)} a geometry-aware video refinement model that further enhances the perceptual quality of synthesized videos. Extensive experiments show that Demiurge achieves state-of-the-art performance with \textbf{three key capabilities:} \textbf{1)} coherent spatiotemporal lattice construction over dense multi-view trajectories; \textbf{2)} large-parallax video rerendering under challenging camera motions, such as $180^\circ$ rotations with no input-view overlap; \textbf{3)} geometry-faithful outputs that can be directly optimized into high-quality 3D Gaussian Splatting representations with minimal artifacts. The project page is available at
\url{https://hanxinzhu-lab.github.io/demiurge_project_page/}.

\end{abstract}    
\section{Introduction}
\label{sec:intro}

\begin{figure}[h]
    \centering
    \begin{minipage}{0.60\textwidth}
        \centering
        \includegraphics[width=\linewidth]{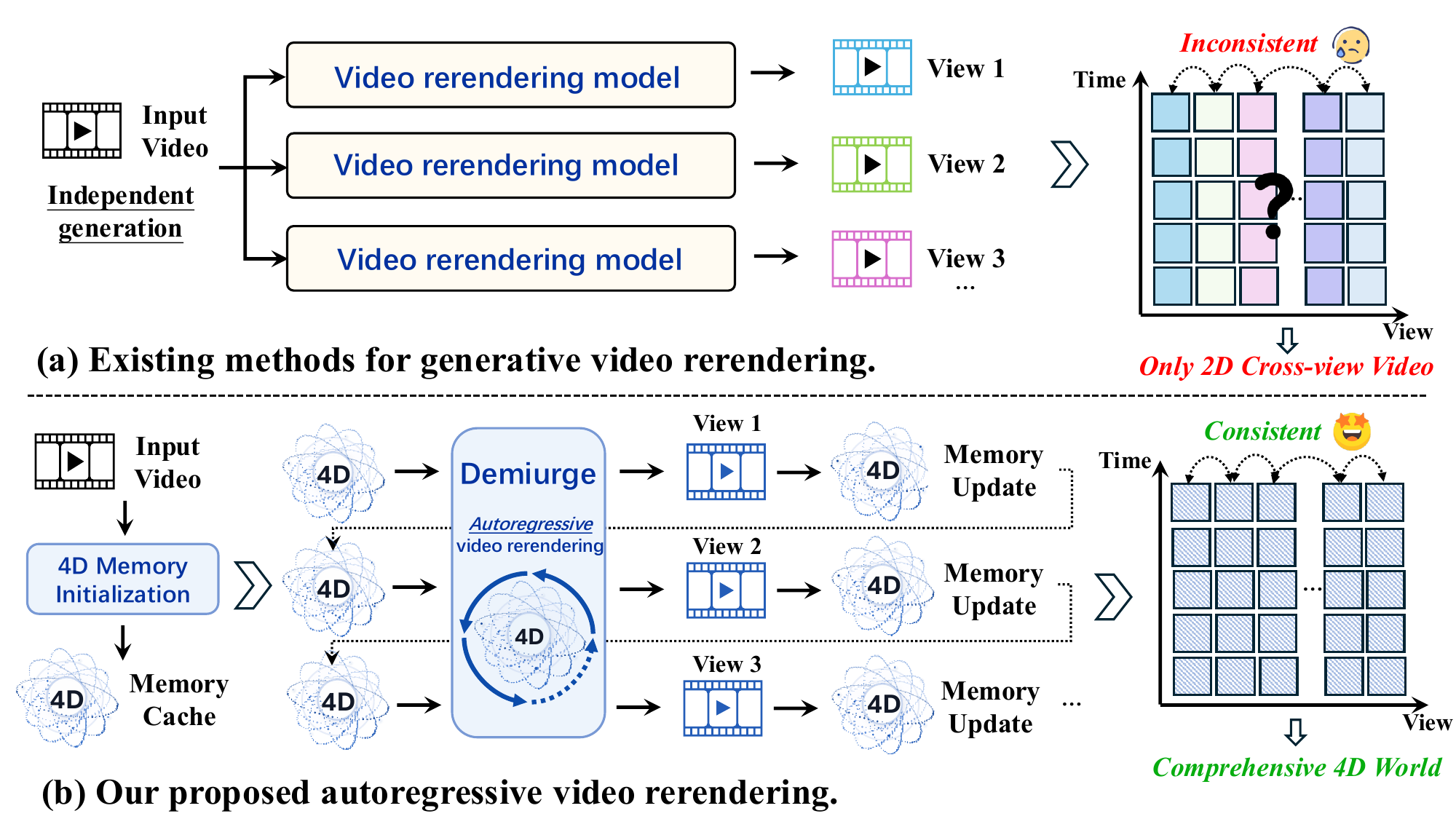}
    \end{minipage}\hfill
    \begin{minipage}{0.37\textwidth}
        \caption{\textbf{Comparison of our proposed method with existing
        methods.} \textbf{(a)} Existing methods independently synthesize videos
        along different target trajectories, resulting in cross-trajectory
        inconsistencies and only isolated 2D cross-view videos. \textbf{(b)} In
        contrast, Demiurge autoregressively generates each trajectory with a
        shared 4D memory, maintaining cross-trajectory consistency and
        thereby enabling coherent 4D world construction.}
        \label{fig:fig1}
    \end{minipage}
\end{figure}

As a fundamental medium for representing evolving visual content, videos provide a natural record of real-world dynamics, serving as a cornerstone for information sharing, content creation, and scene understanding~\cite{tong2022videomae,ho2022video}. However, despite such informative spatiotemporal cues, a single video alone still falls short of fully representing the underlying physical world, as it provides only 2D projections captured along a specific camera trajectory, preserving partial observations of the scene structure while leaving occluded regions unobserved. This inherent limitation has spurred growing interest in generative video rerendering, which aims to synthesize novel-view videos from a single input video along target camera trajectories, offering a promising pathway towards world modeling~\cite{yu2024viewcrafter,yu2025trajectorycrafter}, embodied intelligence~\cite{tu2026embody4d}, and robotic simulation~\cite{zhou2024robodreamer}.

To achieve this goal, leveraging recent advances in video diffusion models, existing methods typically adopt a camera-conditioned video generation paradigm, incorporating target camera trajectories through intrinsics/extrinsics embeddings~\cite{wang2024motionctrl}, Plücker ray maps~\cite{bahmani2025vd3d}, or depth-based warping cues~\cite{lin2026vista4d,zhu2026gta}. 
However, as shown in Fig.~\ref{fig:fig1}(a), while photorealistic novel-view videos can be obtained along individual camera paths, the output for each trajectory is synthesized independently, leaving scene geometry, appearance, and dynamics across different views unconstrained. Consequently, the synthesized observations cannot be consolidated into a globally consistent spatiotemporal lattice, precluding the formation of a coherent and traversable 4D world.

\begin{figure*}[h]
    \centering
    \includegraphics[width=1\linewidth]{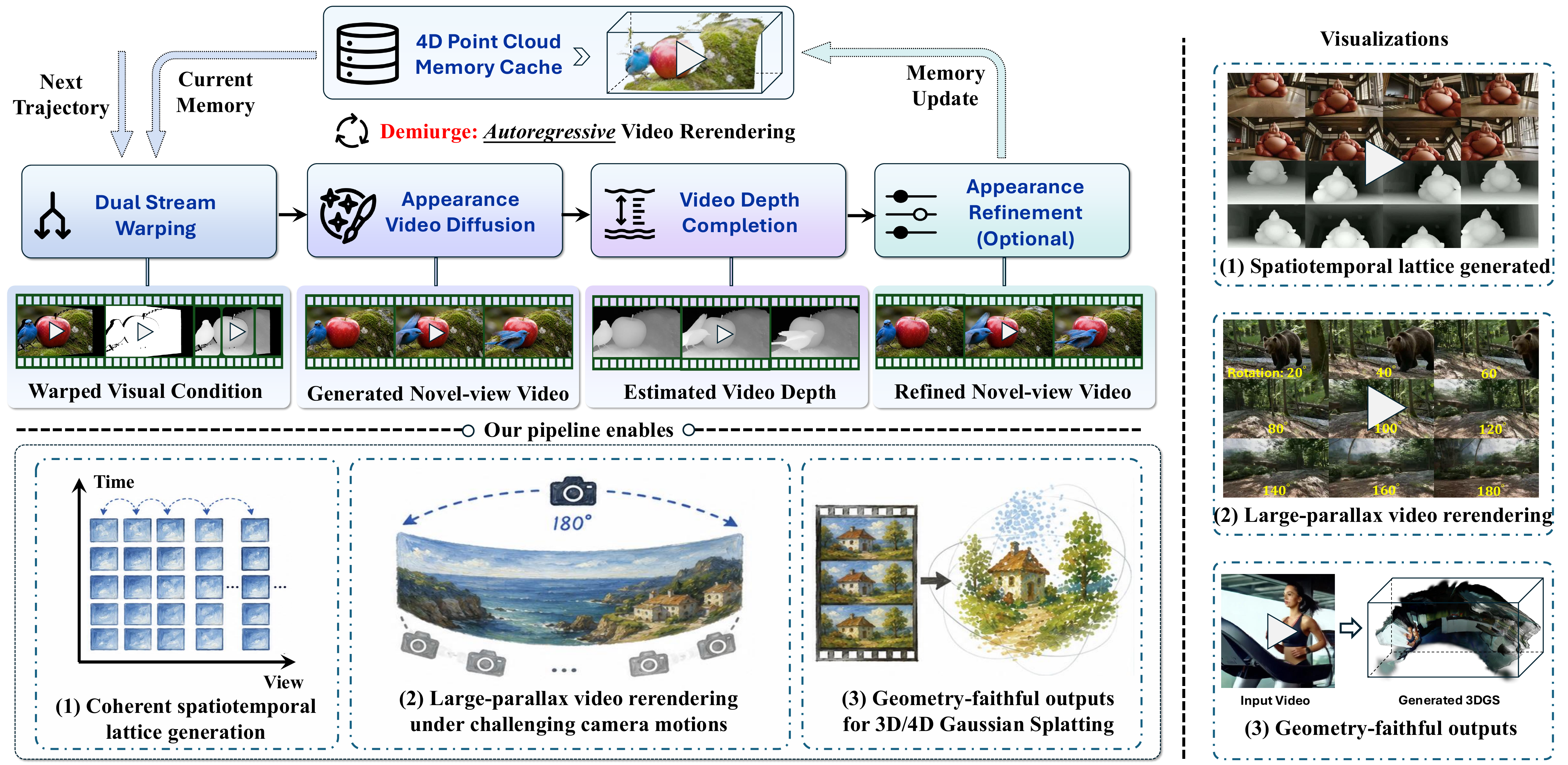}
    \caption{\textbf{Overview of Demiurge.} Demiurge performs autoregressive video rerendering with an evolving 4D point-cloud memory through a \textbf{generate--recover--update cycle}. By integrating dual-stream warping, appearance generation, depth completion, and optional refinement, it enables coherent 4D world construction, large-parallax rerendering, and geometry-faithful 3D/4D reconstruction.}
    \label{fig:fig2}
\end{figure*}

To move beyond trajectory-wise synthesis towards navigable 4D world construction, as shown in Fig.~\ref{fig:fig1}(b), we introduce \textbf{\textit{Demiurge}}, an autoregressive video rerendering framework underpinned by a dynamic 4D point-cloud memory. At each autoregressive step, a novel-view video is synthesized along a target camera trajectory using the current memory as conditioning, with its appearance and recovered geometry subsequently fused back into the 4D memory to guide subsequent rerendering and thus maintain cross-trajectory consistency. Building upon this formulation, as shown in Fig.~\ref{fig:fig2}, \textbf{Demiurge comprises four key components:} \textbf{1) a memory-guided appearance video diffusion model} that synthesizes photorealistic novel-view videos along target camera trajectories using point-cloud renderings from the 4D memory, complemented by a lightweight noise-monitoring mechanism that detects and suppresses conspicuous outliers in the rendered guidance to provide cleaner conditioning signals; \textbf{2) a memory-consistent video depth completion model} that recovers temporally coherent dense depth maps for each synthesized novel-view video under the guidance of partial depth renderings from the 4D memory, anchoring predictions in previously observed regions to cached metric geometry while completing newly revealed regions, thereby allowing the recovered depths to be back-projected into dynamic point clouds and reliably fused into the memory for geometrically consistent memory updates; \textbf{3) a memory-aware dual-stream warping strategy} that derives high-fidelity appearance guidance from a selectively updated point cloud and reliable occlusion cues from a densely updated counterpart, fusing the two according to their depth discrepancies to preserve stable visual content and guide plausible synthesis in newly exposed regions, thereby suppressing geometric artifacts and cumulative color drift over autoregressive iterations; \textbf{4) an optional geometry-aware video refinement model} that leverages temporally consistent target-view depth maps alongside partial RGB renderings to attenuate subtle texture flicker and noise, yielding refined outputs with enhanced visual fidelity and temporal coherence.

Extensive evaluations spanning diverse scenes and viewing paths demonstrate that Demiurge attains state-of-the-art results in perceptual realism, geometric accuracy, and cross-trajectory consistency. Notably, it delivers \textbf{three capabilities} essential to navigable 4D worlds: \textbf{1)} assembling coherent spatiotemporal lattices over densely sampled viewpoints; \textbf{2)} sustaining robust rerendering under extreme parallax, including $180^\circ$ rotations without input-view overlap; and \textbf{3)} producing geometry-faithful videos that can be directly optimized into high-quality 3D Gaussian Splatting representations with minimal artifacts.

Our main contributions are summarized as follows:

\begin{itemize}[leftmargin=1em, itemsep=3pt, topsep=0pt, parsep=0pt]
    \item We propose \textbf{\textit{Demiurge}}, an autoregressive video rerendering
    framework with a dynamic 4D point-cloud memory for
    cross-trajectory-consistent and navigable 4D world generation.

    \item To ensure high-fidelity synthesis and consistent memory evolution,
    \textbf{we introduce four coor\-dinated designs:}\ an appearance video
    diffusion model for photorealistic novel-view synthesis; a video depth
    completion model for reliable geometry recovery and memory updates; a
    dual-stream warping strategy for appearance-preserving occlusion handling
    without cumulative color drift; and an optional video refinement model for
    attenuating texture flicker and noise.

    \item Extensive experiments demonstrate that Demiurge achieves
    state-of-the-art perceptual realism and cross-trajectory consistency,
    supports robust large-parallax rerendering, and yields geometry-faithful
    outputs for high-quality 3D Gaussian Splatting reconstruction.
\end{itemize}

\vspace{-2mm}
\section{Related Works}
\label{sec:related}

\vspace{-3mm}
\subsection{Camera-Controllable Video Generation}
\vspace{-3mm}

Early methods condition video diffusion on parameterized camera poses, either through a dedicated camera-motion module~\cite{wang2024motionctrl} or via
pixel-aligned Pl\"ucker ray embeddings~\cite{he2024cameractrl,xu2024camco,
bahmani2025ac3d,bahmani2025vd3d,he2025cameractrl,zheng2024cami2v}. However, since viewpoint transformations are learned
implicitly, they are prone to geometric distortion and content drift under large
camera motion. Recently, geometry-grounded methods~\cite{gu2025diffusion,ma2025controllable} instead lift source observations into 3D
and reproject them into target views. For example, ViewCrafter~\cite{yu2024viewcrafter} and
GEN3C~\cite{ren2025gen3c} condition on partial point-cloud renderings, FlexWorld~\cite{chen2025flexworld} and Voyager~\cite{huang2025voyager} progressively expand persistent 3D scenes with newly generated observations, and
GTA~\cite{zhu2026gta} decouples geometry generation from appearance synthesis. More recently, camera control has been repurposed as a generative renderer for dynamic scenes~\cite{zheng2026versecrafter,yang2026neoverse,song2026taming,lin2026vista4d,park2026zero4d}, re-rendering monocular videos along novel trajectories while
preserving their original dynamics, either from encoded
extrinsics~\cite{bai2025recammaster} or by explicitly warping the source video with estimated geometry~\cite{yu2025trajectorycrafter}. Both, however, synthesize each trajectory independently and retain nothing for subsequent views. In contrast, Demiurge maintains an explicit 4D memory of generated appearance and recovered geometry, enforcing long-range consistency across diverse camera trajectories.

\vspace{-3mm}
\subsection{Memory-Augmented Video Generation}
\vspace{-3mm}

To mitigate context forgetting and visual drift in long-horizon generation, one
line of work retrieves or reuses historical frames according to their relevance
to the current state~\cite{xiao2026worldmem,yu2025context,
schneider2025worldexplorer,chen2025deepverse}, while another compresses the
generation history into hierarchical, multi-scale, or sliding-window latent
memories~\cite{liu2026worldweaver,li2025hunyuan,mao2025yume,cai2026mixture,
team2026inspatio}. Such appearance- and feature-space memories offer limited
explicit geometric grounding, causing spatial drift under large camera motion.
Geometry-aware approaches instead organize history in explicit 3D space, using a
surfel-indexed view memory~\cite{li2025vmem}, point-cloud spatial
context~\cite{wu2026video}, an incrementally expanded RGB-D
cache~\cite{huang2025voyager}, or a persistent point cloud with dynamic--static
separation~\cite{zhao2026spatia}; yet they predominantly preserve static geometry
rather than the complete time-varying scene state. Most closely related to ours,
PlenopticDreamer~\cite{fu2026plenoptic} banks previously generated
video--camera pairs and retrieves the top-$k$ by 3D field-of-view overlap. Its
fixed context capacity, however, forces selective retrieval or trajectory fusion
once the number of trajectories exceeds $k$, discarding fine-grained detail and
tying long-range consistency to retrieval quality. In contrast, Demiurge
consolidates reliable appearance and recovered geometry from all generated
trajectories into a unified 4D memory, avoiding a fixed video-context bottleneck.

\vspace{-2mm}
\section{Method}
\label{sec:method}

\vspace{-2mm}
\paragraph{Overview.}
Given an input video $\mathbf{V}^{0}=\{\mathbf{I}^{0}_{t}\}_{t=1}^{T}$ and its camera trajectory $\mathbf{C}^{0}=\{\mathbf{C}^{0}_{t}\}_{t=1}^{T}$, our goal is to rerender the observed scene along $N$ target trajectories
$\{\mathbf{C}^{i}\}_{i=1}^{N}$, where $t$ indexes the $T$ video frames and $i$ indexes the target trajectories. To this end, Demiurge maintains an evolving 4D point-cloud memory
$\mathcal{M}_{i}=\{\mathcal{P}_{i,t}\}_{t=1}^{T}$ as a persistent world state that retains historical geometry and appearance across autoregressive iterations, where $\mathcal{P}_{i,t}$ denotes the colored 3D point cloud accumulated at time $t$ after processing the $i$-th trajectory. The initial memory $\mathcal{M}_{0}$ is reconstructed from the input video using a monocular video depth estimator. As illustrated in Fig.~\ref{fig:fig2}, at iteration $i$, the current memory $\mathcal{M}_{i-1}$ is rendered (via dual-stream warping in Sec.~\ref{subsec: Dual-stream Warping}) along $\mathbf{C}^{i}$ to guide the synthesis of a novel-view video
$\mathbf{V}^{i}=\{\mathbf{I}^{i}_{t}\}_{t=1}^{T}$ (via appearance video diffusion in Sec.~\ref{methos:Memory-guided Appearance Video Diffusion}) and its temporally coherent dense depth maps $\mathbf{D}^{i}=\{\mathbf{D}^{i}_{t}\}_{t=1}^{T}$ (via video depth completion in Sec.~\ref{method:Memory-consistent Video Depth Completion}). The generated appearance and recovered geometry are subsequently back-projected and fused into the memory to obtain $\mathcal{M}_{i}$, which guides the next trajectory. By iterating this generate--recover--update process, Demiurge progressively constructs a navigable 4D world. Further details on the 4D memory representation and point-cloud warping are provided in Appendix~\ref{sec:supp_4d_point_cloud}.

\vspace{-2mm}
\subsection{Memory-guided Appearance Video Diffusion}
\label{methos:Memory-guided Appearance Video Diffusion}
\vspace{-3mm}

Constructing a navigable 4D world requires progressively expanding the observed scene along new camera trajectories while preserving the appearance and geometry accumulated from previous observations. This, in turn, calls for a generative model capable of synthesizing photorealistic videos along a target trajectory while remaining grounded in the current world state. To this end, at the $i$-th autoregressive step, we first render the 4D memory $\mathcal{M}_{i-1}$ along the target trajectory $\mathbf{C}^{i}$ to obtain a sequence of memory-derived visual conditions:
\begin{equation}
\left(
\widetilde{\mathbf{V}}^{i},
\widetilde{\mathbf{D}}^{i},
\widetilde{\mathbf{M}}^{i}
\right)
=
\mathcal{R}_{\mathrm{dual}}
\left(
\mathcal{M}_{i-1},
\mathbf{C}^{i}
\right),
\label{eq:cache_rendering}
\end{equation}
where $\mathcal{R}_{\mathrm{dual}}$ denotes our dual-stream warping operator presented in Sec.~\ref{subsec: Dual-stream Warping} (Eq.~\ref{eq:dual_stream_operator}), and $\widetilde{\mathbf{V}}^{i}$, $\widetilde{\mathbf{D}}^{i}$, and $\widetilde{\mathbf{M}}^{i}$ respectively denote the resulting partial RGB video, partial depth video, and visibility masks, providing complementary appearance, geometry, and visibility cues. 

However, discrete point-cloud rendering may inherently produce holes and unstable projections (as shown in Fig.~\ref{fig:ablation_qualitaive_results}(a)), particularly near occlusion boundaries, which can accumulate over successive autoregressive iterations. To prevent such corrupted observations from conditioning the subsequent generation process, we introduce a lightweight, non-learned noise monitor that identifies potentially unreliable pixels from the rendered depth. Specifically, for each pixel $\mathbf{p}$ in the depth frame $\widetilde{\mathbf{D}}^{i}_{t}$, we define
\begin{equation}
\begin{aligned}
\mathbf{O}^{i}_{t}(\mathbf{p})
=
\mathbb{I}\Big[
\widetilde{\mathbf{D}}^{i}_{t}(\mathbf{p}) < \tau_{0}
\;\lor\;
\left\|
\nabla\widetilde{\mathbf{D}}^{i}_{t}(\mathbf{p})
\right\|_{2} > \tau_{g}
\lor\;
\operatorname{Var}_{\mathbf{q}\in\mathcal{N}_{3}(\mathbf{p})}
\left(
\widetilde{\mathbf{D}}^{i}_{t}(\mathbf{q})
\right) > \tau_{v}
\Big],
\end{aligned}
\label{eq:noise_monitor}
\end{equation}
where $\mathcal{N}_{3}(\mathbf{p})$ denotes the $3\times3$ neighborhood centered at $\mathbf{p}$, and $\tau_{0}$, $\tau_{g}$, and $\tau_{v}$ are thresholds for invalid depth, depth-gradient magnitude, and local depth variance, respectively. The resulting sequence $\mathbf{O}^{i}=\{\mathbf{O}^{i}_{t}\}_{t=1}^{T}$ captures missing depth values, abrupt depth variations, and locally unstable projections. We consistently invalidate these pixels across all rendered conditions:
\begin{equation}
\left(
\overline{\mathbf{V}}^{i},
\overline{\mathbf{D}}^{i},
\overline{\mathbf{M}}^{i}
\right)
=
\left(1-\mathbf{O}^{i}\right)
\odot
\left(
\widetilde{\mathbf{V}}^{i},
\widetilde{\mathbf{D}}^{i},
\widetilde{\mathbf{M}}^{i}
\right).
\label{eq:condition_cleaning}
\end{equation}
Given the cleaned conditions, our memory-guided appearance video diffusion model $\mathcal{F}_{\mathrm{app}}$ synthesizes the target novel-view video as
\begin{equation}
\mathbf{V}^{i}
=
\mathcal{F}_{\mathrm{app}}
\left(
\overline{\mathbf{V}}^{i},
\overline{\mathbf{D}}^{i},
\overline{\mathbf{M}}^{i}
\right),
\label{eq:appearance_generation}
\end{equation}
where $\mathbf{V}^{i}=\{\mathbf{I}^{i}_{t}\}_{t=1}^{T}$ denotes the generated RGB video along $\mathbf{C}^{i}$. Internally, the three conditions are concatenated with the noisy video latent during denoising. In this way, $\mathcal{F}_{\mathrm{app}}$ preserves reliable memory-supported content while regenerating unreliable and newly exposed regions with photorealistic details. Further details are provided in Appendix~\ref{sec:supp_noise_monitor} and ~\ref{sec:supp_appearance_diffusion}.

\vspace{-2mm}
\subsection{Memory-consistent Video Depth Completion}
\label{method:Memory-consistent Video Depth Completion}
\vspace{-3mm}

Given the generated novel-view RGB video $\mathbf{V}^{i}$, to incorporate its newly observed content into the 4D memory and support the next autoregressive step, we need to recover a dense video depth that is both temporally coherent and geometrically consistent with the existing memory. A straightforward solution is to apply an off-the-shelf monocular video depth estimator to $\mathbf{V}^{i}$. However, monocular depth is inherently scale-ambiguous and may disagree with the cached geometry even in previously observed regions. Directly back-projecting such predictions can introduce misaligned surfaces and geometric discontinuities into the memory (as shown in Fig.~\ref{fig:ablation_qualitaive_results}(b)), which may further accumulate over subsequent iterations. To address this issue, we therefore recover dense geometry from the generated video through a memory-consistent, diffusion-based video depth completion model.

Specifically, our model takes the generated video $\mathbf{V}^{i}$ and the cleaned partial cached depth $\overline{\mathbf{D}}^{i}$ as inputs. The partial cached depth anchors the prediction to the accumulated geometry in previously observed regions, while the generated RGB video provides spatial and temporal cues for estimating the geometry of newly revealed content. Since the model is built upon latent video diffusion, we first convert the partial depth into disparity and normalize its valid values to $[0,1]$:
\begin{equation}
\mathbf{U}^{i}
=
1 / ({\overline{\mathbf{D}}^{i}+\epsilon}),
\
\overline{\mathbf{U}}^{i}
=
\mathcal{N}
\left(
\mathbf{U}^{i};
\overline{\mathbf{M}}^{i}
\right),
\label{eq:depth_to_disparity}
\end{equation}
where $\epsilon$ ensures numerical stability, and $\mathcal{N}(\cdot)$ denotes min--max normalization over valid pixels.

Conditioned on the generated RGB video and normalized partial disparity, our memory-consistent video depth completion model $\mathcal{F}_{\mathrm{dep}}$ predicts a temporally coherent dense relative disparity video:
\begin{equation}
\widehat{\mathbf{U}}^{i}
=
\mathcal{F}_{\mathrm{dep}}
\left(
\mathbf{V}^{i},
\overline{\mathbf{U}}^{i}
\right).
\label{eq:depth_completion}
\end{equation}
To preserve memory-derived geometry throughout denoising, we employ a dual-path conditioning design within $\mathcal{F}_{\mathrm{dep}}$. The partial cached disparity anchors the denoising process to previously observed geometry, while the generated RGB video provides complementary spatial and temporal cues for completing newly revealed regions. Detailed architecture is provided in the supplementary material.

While $\widehat{\mathbf{U}}^{i}$ captures the relative geometric
structure of the target video, it does not preserve the metric scale
of the memory. We therefore employ a two-stage global affine alignment.
Specifically, we first align the prediction to the normalized partial
cached depth and then recover its metric scale using the original cached
depth. Both transformations are estimated over all memory-supported
pixels and shared across frames, avoiding temporal scale fluctuations.
We denote the resulting dense metric-depth prediction by
$\mathbf{G}^{i}$.

Despite the global alignment, small local deviations may still remain in previously observed regions, while directly replacing them with cached depths can introduce boundary discontinuities. We therefore perform Poisson refinement~\cite{perez2023poisson} in metric-depth space:
\begin{equation}
\begin{aligned}
    \mathbf{D}^{i}
    =
    \underset{\mathbf{X}^{i}}{\arg\min}\quad
    \sum_{t=1}^{T}
    \left\|
        \nabla\mathbf{X}_{t}^{i}
        -
        \nabla\mathbf{G}_{t}^{i}
    \right\|_{2}^{2},
    \mathrm{s.t.}\quad
    \mathbf{X}_{t}^{i}(\mathbf{p})
    =
    \overline{\mathbf{D}}_{t}^{i}(\mathbf{p}),
    \quad
    \forall\,\mathbf{p}
    \ \text{with}\
    \overline{\mathbf{M}}_{t}^{i}(\mathbf{p})=1.
\end{aligned}
\label{eq:poisson_refinement}
\end{equation}
This refinement preserves the spatial gradients of the aligned dense prediction while enforcing exact agreement with the cached geometry. The resulting dense depth video $\mathbf{D}^{i}$ is subsequently fused into the 4D memory together with $\mathbf{V}^{i}$. Further alignment and solver details are provided in the supplementary material. We defer further details to Appendix~\ref{sec:supp_depth_completion}.

\vspace{-2mm}
\subsection{Memory-aware Dual-stream Warping}\label{subsec: Dual-stream Warping}
\vspace{-3mm}

After recovering the RGB-D video $(\mathbf{V}^{i},\mathbf{D}^{i})$, we next consider how to update the point-cloud memory so that it can provide reliable conditions for the subsequent target trajectory. A natural strategy is to register only newly generated pixels while retaining the points accumulated in previously observed regions. Such a selective update prevents stable colors from being repeatedly overwritten by diffusion-generated appearance, thereby preserving sharp and visually consistent renderings over successive autoregressive iterations.

However, selective updates may leave the point cloud insufficiently dense for reliable visibility reasoning (as shown in the leftmost RGB--depth pair of
Fig.~\ref{fig:ablation_qualitaive_results}(d)
). Under large viewpoint changes, sparse foreground points may fail to fully occlude the background during z-buffer rendering, causing background content that should remain hidden to leak into foreground regions. This failure is not captured by the visibility mask, which only records whether a pixel originates from cached or newly generated content, rather than whether its projection remains geometrically valid under the target viewpoint. Consequently, these erroneous projections may be treated as reliable appearance guidance, propagated into the diffusion output, and eventually written back into the memory.

An alternative is to register all pixels from each generated RGB-D video. The resulting dense point cloud provides more complete foreground geometry and enables more reliable occlusion reasoning under large parallax. Nevertheless, this strategy also repeatedly incorporates the generated appearance of previously observed regions. Small texture deviations and high-frequency artifacts can therefore accumulate over multiple iterations, eventually producing blurred renderings and color drift (as shown in the middle RGB--depth pair of
Fig.~\ref{fig:ablation_qualitaive_results}(d)).

To reconcile the competing requirements of appearance preservation and geometric completeness, we introduce \textbf{dual-stream warping}. Conceptually, the global 4D memory comprises two complementary memory states:
\begin{equation}
\mathcal{M}_{i}
=
\left(
\mathcal{M}_{i}^{\mathrm{app}},
\mathcal{M}_{i}^{\mathrm{geo}}
\right),
\label{eq:dual_memory}
\end{equation}
where the appearance memory $\mathcal{M}_{i}^{\mathrm{app}}$ selectively incorporates newly synthesized regions to preserve stable colors, while the geometry memory $\mathcal{M}_{i}^{\mathrm{geo}}$ densely incorporates all valid RGB-D observations to support reliable occlusion reasoning. Both memories are initialized from the same input video and subsequently evolve under their respective update policies:
\begin{equation}
\begin{aligned}
\mathcal{M}_{i}^{\mathrm{app}}
&=
\operatorname{Fuse}
\left(
\mathcal{M}_{i-1}^{\mathrm{app}},
\operatorname{BackProj}
\left(
\mathbf{V}^{i},
\mathbf{D}^{i},
\mathbf{C}^{i};
1-\overline{\mathbf{M}}^{i}
\right)
\right),\\
\mathcal{M}_{i}^{\mathrm{geo}}
&=
\operatorname{Fuse}
\left(
\mathcal{M}_{i-1}^{\mathrm{geo}},
\operatorname{BackProj}
\left(
\mathbf{V}^{i},
\mathbf{D}^{i},
\mathbf{C}^{i};
\mathbf{1}
\right)
\right).
\end{aligned}
\label{eq:dual_memory_update}
\end{equation}
Here, $\operatorname{BackProj}(\cdot)$ lifts the selected RGB-D pixels into world-space colored point clouds using the corresponding camera parameters, while $\operatorname{Fuse}(\cdot)$ appends the recovered points to the accumulated memory. The last argument of $\operatorname{BackProj}(\cdot)$ denotes the pixel-selection mask. 

For the next target trajectory $\mathbf{C}^{i+1}$, we render the two memory states in parallel, producing an appearance warping stream and a geometry warping stream:
\begin{equation}
\begin{aligned}
\left(
\mathbf{V}_{\mathrm{app}}^{i+1},
\mathbf{D}_{\mathrm{app}}^{i+1},
\mathbf{M}_{\mathrm{app}}^{i+1}
\right)
=
\mathcal{R}
\left(
\mathcal{M}_{i}^{\mathrm{app}},
\mathbf{C}^{i+1}
\right),
\left(
\mathbf{V}_{\mathrm{geo}}^{i+1},
\mathbf{D}_{\mathrm{geo}}^{i+1},
\mathbf{M}_{\mathrm{geo}}^{i+1}
\right)
=
\mathcal{R}
\left(
\mathcal{M}_{i}^{\mathrm{geo}},
\mathbf{C}^{i+1}
\right),
\end{aligned}
\label{eq:dual_stream_rendering}
\end{equation}
where $\mathcal{R}$ denotes point cloud rendering.

The appearance stream provides sharp and stable visual guidance, whereas the geometry stream offers denser geometric coverage for more reliable occlusion reasoning. We identify viewpoint-dependent projection conflicts by comparing their rendered depths:
\begin{equation}
\boldsymbol{\Delta}_{t}^{i+1}(\mathbf{p})
=
\left|
\mathbf{D}_{\mathrm{app},t}^{i+1}(\mathbf{p})
-
\mathbf{D}_{\mathrm{geo},t}^{i+1}(\mathbf{p})
\right|, \
\mathbf{Q}_{t}^{i+1}(\mathbf{p})
=
\mathbb{I}
\left[
\boldsymbol{\Delta}_{t}^{i+1}(\mathbf{p})
>
\tau_{\mathrm{occ}}
\right],
\label{eq:dual_stream_conflict}
\end{equation}
where $\mathbf{p}$ denotes a pixel location in the target image plane, $\boldsymbol{\Delta}_{t}^{i+1}(\mathbf{p})$ measures the depth discrepancy between the two streams at frame $t$, $\mathbf{Q}^{i+1}$ is the resulting conflict mask, and $\tau_{\mathrm{occ}}$ is the discrepancy threshold. In regions without detected conflicts, we retain the RGB-D conditions from the appearance stream. At locations indicated by $\mathbf{Q}^{i+1}$, we instead adopt the geometry-stream conditions, whose denser point distribution provides reliable visibility reasoning:
\begin{equation}
\begin{aligned}
\widetilde{\mathbf{V}}^{i+1}
&=
\left(1-\mathbf{Q}^{i+1}\right)
\odot
\mathbf{V}_{\mathrm{app}}^{i+1}
+
\mathbf{Q}^{i+1}
\odot
\mathbf{V}_{\mathrm{geo}}^{i+1},\\
\widetilde{\mathbf{D}}^{i+1}
&=
\left(1-\mathbf{Q}^{i+1}\right)
\odot
\mathbf{D}_{\mathrm{app}}^{i+1}
+
\mathbf{Q}^{i+1}
\odot
\mathbf{D}_{\mathrm{geo}}^{i+1},\\
\widetilde{\mathbf{M}}^{i+1}
&=
\left(1-\mathbf{Q}^{i+1}\right)
\odot
\mathbf{M}_{\mathrm{app}}^{i+1}
+
\mathbf{Q}^{i+1}
\odot
\mathbf{M}_{\mathrm{geo}}^{i+1}.
\end{aligned}
\label{eq:dual_stream_fusion}
\end{equation}

Equivalently, denoting the pixel-wise fusion defined in Eqs.~\ref{eq:dual_stream_conflict} and~\ref{eq:dual_stream_fusion} by $\mathcal{H}$, the complete dual-stream warping operator can be summarized as
\begin{equation}
\begin{aligned}
\left(
\widetilde{\mathbf{V}}^{i+1},
\widetilde{\mathbf{D}}^{i+1},
\widetilde{\mathbf{M}}^{i+1}
\right)
=
\mathcal{R}_{\mathrm{dual}}
\left(
\mathcal{M}_{i},
\mathbf{C}^{i+1}
\right)
=
\mathcal{H}
\Big(
\mathcal{R}
\left(
\mathcal{M}_{i}^{\mathrm{app}},
\mathbf{C}^{i+1}
\right),
\mathcal{R}
\left(
\mathcal{M}_{i}^{\mathrm{geo}},
\mathbf{C}^{i+1}
\right)
\Big).
\end{aligned}
\label{eq:dual_stream_operator}
\end{equation}

The fused conditions $(\widetilde{\mathbf{V}}^{i+1},\widetilde{\mathbf{D}}^{i+1},\widetilde{\mathbf{M}}^{i+1})$ can be subsequently passed to $\mathcal{F}_{\mathrm{app}}$ for the next autoregressive generation step. This pixel-wise selection preserves the sharp appearance produced by selective warping over most regions while exploiting dense geometry where depth disagreement indicates a potential occlusion conflict. In this way, dual-stream warping suppresses both background leakage under large viewpoint changes and appearance degradation over successive iterations.
Notably, while formulated as two logical memory states, the two streams share the same accumulated RGB-D observations, camera poses, and per-trajectory provenance masks. In practice, they are implemented through two warping passes with different source-mask configurations: the appearance pass respects the stored provenance masks, whereas the geometry pass treats all source pixels as valid. Thus, no duplicated point-cloud memory is required. Kindly refer to more details in Appendix~\ref{sec:supp_dual_stream_warping}.

\vspace{-4mm}
\subsection{Geometry-aware Video Refinement}
\vspace{-3mm}

With the novel-view RGB video $\mathbf{V}^{i}$, its dense depth video $\mathbf{D}^{i}$, and the memory update mechanism described above, the core autoregressive loop is already sufficient to support the generation of subsequent target trajectories. However, we observe that $\mathbf{V}^{i}$ may occasionally exhibit subtle texture flicker and high-frequency noise, which degrade its perceptual quality (as shown in Fig.~\ref{fig:ablation_qualitaive_results}(c)). Fortunately, our video depth completion model remains robust to these mild appearance artifacts and produces temporally stable and spatially smooth depth predictions. This observation motivates an optional geometry-aware refinement stage that leverages the recovered depth as a reliable structural prior to further improve the generated appearance.

Specifically, we introduce a diffusion-based, geometry-aware video refinement model $\mathcal{F}_{\mathrm{ref}}$, which takes the partial RGB rendering $\overline{\mathbf{V}}^{i}$ and the completed target-view depth $\mathbf{D}^{i}$ as inputs:
\begin{equation}
\mathbf{V}_{\mathrm{ref}}^{i}
=
\mathcal{F}_{\mathrm{ref}}
\left(
\overline{\mathbf{V}}^{i},
\mathbf{D}^{i}
\right),
\label{eq:geometry_aware_refinement}
\end{equation}
where $\mathbf{V}_{\mathrm{ref}}^{i}=\{\mathbf{I}_{\mathrm{ref},t}^{i}\}_{t=1}^{T}$ denotes the refined novel-view video. The partial RGB rendering provides appearance guidance from the 4D memory, while the dense depth supplies complete and temporally coherent geometric guidance across the target views. Conditioned on these complementary signals, $\mathcal{F}_{\mathrm{ref}}$ reduces texture flicker and noise while preserving the recovered scene structure, yielding refined videos with improved visual fidelity and temporal coherence. As an optional module, this refinement stage enhances the final video quality without affecting the core autoregressive world-expansion process. More details are provided in Appendix~\ref{sec:supp_geometry_refinement}.

\vspace{-3mm}
\section{Experiments}
\label{sec:experiments}
\vspace{-3mm}

\begin{table*}[b]
    \centering
    \caption{\textbf{Quantitative comparison of generated novel-view videos.} Demiurge achieves the best perceptual quality, cross-trajectory consistency, and camera controllability. Red, orange, and yellow denote the best, second-best, and third-best results, respectively.}
    \label{tab:main_results}
    \setlength{\tabcolsep}{5.5pt}
    \renewcommand{\arraystretch}{1.15}
    \resizebox{\textwidth}{!}{%
    \begin{tabular}{lccccccccc}
        \toprule
        \multirow{2}{*}{Method}
        & \multirow{2}{*}{FVD $\downarrow$}
        & \multicolumn{2}{c}{Q-Align-Video $\uparrow$}
        & \multicolumn{2}{c}{Q-Align-Image $\uparrow$}
        & \multirow{2}{*}{P-MSE $\downarrow$}
        & \multirow{2}{*}{R-Err $\downarrow$}
        & \multirow{2}{*}{T-Err $\downarrow$}
        & \multirow{2}{*}{View Sync $\uparrow$} \\
        \cmidrule(lr){3-4}
        \cmidrule(lr){5-6}
        & & Quality & Aesthetics & Quality & Aesthetics & & & & \\
        \midrule

        TrajCrafter~\cite{yu2025trajectorycrafter}
        & 597.652
        & 4.262
        & 3.242
        & 3.950
        & 3.035
        & \cellcolor{orange!25}0.007
        & \cellcolor{orange!25}4.106
        & \cellcolor{yellow!25}2.403
        & 21.426 \\

        InspatioWorld~\cite{team2026inspatio}
        & \cellcolor{yellow!25}369.577
        & \cellcolor{orange!25}4.413
        & \cellcolor{orange!25}3.366
        & \cellcolor{orange!25}4.196
        & \cellcolor{orange!25}3.135
        & 0.017
        & 7.119
        & \cellcolor{orange!25}2.159
        & \cellcolor{yellow!25}40.607 \\

        ReCamMaster~\cite{bai2025recammaster}
        & 478.897
        & 4.189
        & 3.162
        & 3.956
        & 2.966
        & 0.046
        & \cellcolor{yellow!25}4.248
        & 2.934
        & 39.624 \\

        Gen3C~\cite{ren2025gen3c}
        & \cellcolor{orange!25}363.400
        & \cellcolor{yellow!25}4.277
        & 3.266
        & 4.060
        & 3.074
        & \cellcolor{yellow!25}0.008
        & 6.645
        & 2.647
        & \cellcolor{orange!25}46.562 \\

        Voyager~\cite{huang2025voyager}
        & 839.415
        & 4.081
        & 3.141
        & 3.774
        & 2.950
        & 0.031
        & 10.803
        & 3.715
        & 22.106 \\

        Spatia~\cite{zhao2026spatia}
        & 477.889
        & 4.234
        & \cellcolor{yellow!25}3.347
        & \cellcolor{yellow!25}4.093
        & \cellcolor{yellow!25}3.125
        & 0.023
        & 6.798
        & 3.172
        & 33.171 \\

        \midrule
        \textbf{Ours}
        & \cellcolor{red!25}334.518
        & \cellcolor{red!25}4.549
        & \cellcolor{red!25}3.507
        & \cellcolor{red!25}4.331
        & \cellcolor{red!25}3.291
        & \cellcolor{red!25}0.001
        & \cellcolor{red!25}3.655
        & \cellcolor{red!25}1.666
        & \cellcolor{red!25}64.893 \\
        \bottomrule
    \end{tabular}%
    }
\end{table*}

\begin{table*}[t]
    \centering
    \caption{\textbf{Quantitative comparison of 3DGS reconstructed from generated videos.} Average is the geometric mean of $\mathrm{MSE}=10^{-\mathrm{PSNR}/10}$, $\sqrt{1-\mathrm{SSIM}}$, and LPIPS. Red, orange, and yellow denote the best, second-best, and third-best results, respectively.}
    \label{tab:reconstruction_results}
    \setlength{\tabcolsep}{5.5pt}
    \renewcommand{\arraystretch}{1.15}
    \resizebox{\textwidth}{!}{%
    \begin{tabular}{lccccccccc}
        \toprule
        \multirow{2}{*}{Method}
        & \multirow{2}{*}{FVD $\downarrow$}
        & \multicolumn{2}{c}{Q-Align-Video $\uparrow$}
        & \multicolumn{2}{c}{Q-Align-Image $\uparrow$}
        & \multirow{2}{*}{PSNR $\uparrow$}
        & \multirow{2}{*}{SSIM $\uparrow$}
        & \multirow{2}{*}{LPIPS $\downarrow$}
        & \multirow{2}{*}{Average $\downarrow$} \\
        \cmidrule(lr){3-4}
        \cmidrule(lr){5-6}
        & & Quality & Aesthetics & Quality & Aesthetics & & & & \\
        \midrule

        TrajCrafter~\cite{yu2025trajectorycrafter}
        & 1258.756
        & 3.430
        & 2.703
        & 2.837
        & 2.481
        & 20.153
        & 0.671
        & 0.368
        & 0.127 \\

        InspatioWorld~\cite{team2026inspatio}
        & \cellcolor{orange!25}1141.013
        & \cellcolor{orange!25}3.590
        & \cellcolor{orange!25}2.747
        & \cellcolor{orange!25}3.098
        & \cellcolor{orange!25}2.5213
        & 21.767
        & 0.706
        & \cellcolor{yellow!25}0.296
        & 0.102 \\

        ReCamMaster~\cite{bai2025recammaster}
        & 1318.239
        & 3.343
        & 2.547
        & 2.810
        & 2.322
        & \cellcolor{orange!25}22.212
        & \cellcolor{orange!25}0.752
        & 0.327
        & \cellcolor{yellow!25}0.099 \\

        Gen3C~\cite{ren2025gen3c}
        & \cellcolor{yellow!25}1143.709
        & 3.390
        & 2.623
        & 2.865
        & 2.399
        & \cellcolor{yellow!25}22.161
        & \cellcolor{yellow!25}0.727
        & \cellcolor{orange!25}0.282
        & \cellcolor{orange!25}0.096 \\

        Voyager~\cite{huang2025voyager}
        & 1677.240
        & 2.970
        & 2.385
        & 2.263
        & 2.174
        & 16.116
        & 0.568
        & 0.552
        & 0.207 \\

        Spatia~\cite{zhao2026spatia}
        & 1217.144
        & \cellcolor{yellow!25}3.485
        & \cellcolor{yellow!25}2.726
        & \cellcolor{yellow!25}2.922
        & \cellcolor{yellow!25}2.515
        & 21.496
        & 0.706
        & 0.343
        & 0.110 \\

        \midrule
        \textbf{Ours}
        & \cellcolor{red!25}1128.544
        & \cellcolor{red!25}3.625
        & \cellcolor{red!25}2.836
        & \cellcolor{red!25}3.146
        & \cellcolor{red!25}2.617
        & \cellcolor{red!25}24.360
        & \cellcolor{red!25}0.769
        & \cellcolor{red!25}0.237
        & \cellcolor{red!25}0.075 \\
        \bottomrule
    \end{tabular}%
    }
\end{table*}

\begin{figure*}[ht]
    \centering
    \includegraphics[width=1\linewidth]{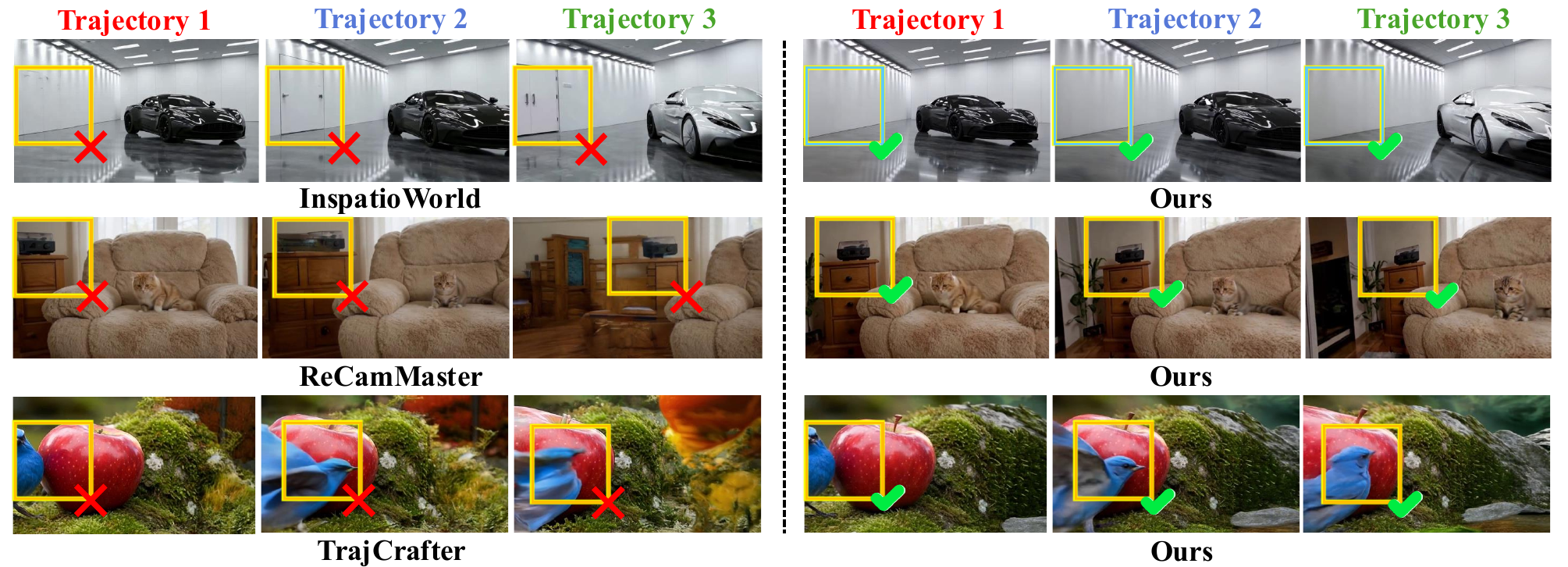}
    \caption{\textbf{Qualitative comparison.} Demiurge better preserves appearance and geometry across camera trajectories, reducing hallucinations, content drift, and structural distortions.}
    \label{fig:qualitative_comparison}
\end{figure*}

\subsection{Experimental setups}
\vspace{-3mm}
\paragraph{Implementation Details.}
All diffusion-based stages of our framework share Wan2.1-1.3B~\cite{wan2025wan} as the generative backbone. The models operate on 81-frame video clips at a resolution of $480\times832$. Our training corpus comprises 236K samples, including 136K samples derived from the MultiCamVideo dataset~\cite{bai2025recammaster} and 100K samples from the OpenVid dataset~\cite{nan2025openvid}. All experiments are conducted using eight NVIDIA H20 GPUs. For systematic evaluation under unconstrained real-world scenarios, we establish a benchmark comprising 128 diverse in-the-wild monocular videos. For each input sequence, we synthesize novel views along a fixed suite of 14 camera trajectories: eight azimuthal trajectories at $\{\pm10^\circ,\pm20^\circ,\pm30^\circ,\pm40^\circ\}$, four elevation trajectories at $\{\pm10^\circ,\pm20^\circ\}$, and two radial zoom-out trajectories with normalized displacements of 0.5 and 1.0. The orbital radius is fixed to unit length for all rotational trajectories. Metrics are computed over all $128\times14=\mathbf{1{,}792}$ generated videos. See more details in Appendix~\ref{sec:supp_training_data}.

\vspace{-4mm}
\paragraph{Baselines and metrics.}
We compare our method against six representative camera-controlled video generation approaches: TrajCrafter~\cite{yu2025trajectorycrafter}, InSpatioWorld~\cite{team2026inspatio}, ReCamMaster~\cite{bai2025recammaster}, Gen3C~\cite{ren2025gen3c}, Voyager~\cite{huang2025voyager}, and Spatia~\cite{zhao2026spatia}.  We evaluate all methods along two complementary dimensions: video quality and consistency. \textit{Video Quality.} We report FVD~\cite{unterthiner2018towards} and Q-Align~\cite{wu2023q} to assess the perceptual quality and aesthetics of generated videos. \textit{Consistency.} We introduce Projection MSE (P-MSE) to measure input--output consistency by projecting the input video onto the generated camera trajectory and computing the MSE over valid overlapping regions. We further use PSNR~\cite{huynh2008scope}, SSIM~\cite{wang2004image}, and LPIPS~\cite{zhang2018unreasonable} to assess the rendering quality of 3DGS representations reconstructed from the generated multi-view videos, and View Sync~\cite{edstedt2024roma} to quantify cross-view correspondence by counting matched pixels above a confidence threshold. Finally, we report rotation error (R-err) and translation error (T-err)~\cite{he2024cameractrl} between the recovered and target camera trajectories to evaluate trajectory adherence. We provide more details in Appendix~\ref{sec:supp_evaluation}.

\vspace{-4mm}
\subsection{Main results}
\vspace{-3mm}
As shown in Tab.~\ref{tab:main_results}, we first evaluate the quality and consistency of the generated novel-view videos. Demiurge achieves the best results across all reported metrics, obtaining the lowest FVD of 334.518 and the highest quality and aesthetics scores under both Q-Align-Video and Q-Align-Image. Compared with the strongest baselines, it reduces P-MSE by 85.7\% and improves View Sync by 39.4\%, demonstrating substantially better content preservation and cross-trajectory consistency. It further reduces R-err and T-err by 11.0\% and 22.8\%, respectively, confirming more accurate adherence to the target camera trajectories. We additionally reconstruct a 3D Gaussian Splatting (3DGS) representation from the videos generated by each method and evaluate its rendering quality in Tab.~\ref{tab:reconstruction_results}. Demiurge again performs best across all metrics, achieving a PSNR of 24.360, an SSIM of 0.769, and an LPIPS of 0.237. In particular, it improves PSNR by 2.148\,dB and reduces LPIPS and the aggregate reconstruction error by 16.0\% and 21.9\%, respectively, indicating that its generated views form a more coherent 3D representation. The qualitative comparisons in Fig.~\ref{fig:qualitative_comparison} further show that competing methods often introduce viewpoint-dependent hallucinations, appearance changes, and geometric distortions, whereas Demiurge better preserves object identity, fine-grained appearance, and spatial layout across diverse viewpoints. Together, these results demonstrate that our proposed method enables high-quality video generation while maintaining long-range consistency across camera trajectories and supporting geometrically coherent 3DGS reconstruction.
Due to space constraints, additional results on key capabilities, qualitative
comparisons, compatibility with existing methods, and visualizations are
provided in Appendices~\ref{appendix: Key capabilities},
\ref{appendix: more qualitative comparisons},
\ref{appendix: Compatibility with Existing Method}, and
\ref{appendix: Additional visualizations}.

\vspace{-4mm}
\subsection{Ablation studies}
\vspace{-3mm}

\begin{table*}[t]
    \centering
    \caption{\textbf{Component ablation of Demiurge.} NM, VDC, DSW, and Ref. denote noise monitoring, video depth completion, dual-stream warping, and geometry-aware refinement, respectively. \emph{M. Est.} directly uses monocular video depth estimation.
    }
    \label{tab:component_ablation}
    \setlength{\tabcolsep}{4.2pt}
    \renewcommand{\arraystretch}{1.15}
    \resizebox{\textwidth}{!}{%
    \begin{tabular}{lcccccccccccc}
        \toprule
        \multirow{2}{*}{Variant}
        & \multicolumn{4}{c}{Components}
        & \multicolumn{4}{c}{Generated Videos}
        & \multicolumn{4}{c}{Reconstructed 3DGS} \\
        \cmidrule(lr){2-5}
        \cmidrule(lr){6-9}
        \cmidrule(lr){10-13}
        & NM
        & Depth Module
        & DSW
        & Ref.
        & FVD$\downarrow$
        & Q-Align-V$\uparrow$
        & P-MSE$\downarrow$
        & View Sync$\uparrow$
        & PSNR$\uparrow$
        & SSIM$\uparrow$
        & LPIPS$\downarrow$
        & Avg.$\downarrow$ \\
        \midrule

        \textit{w/o} Noise Monitor
        & --
        & VDC.
        & \cmark
        & \cmark
        & \cellcolor{orange!25}339.238
        & \cellcolor{yellow!25}4.460
        & \cellcolor{orange!25}0.0013
        & \cellcolor{orange!25}62.575
        & \cellcolor{yellow!25}24.217
        & \cellcolor{orange!25}0.757
        & \cellcolor{orange!25}0.239
        & \cellcolor{yellow!25}0.076 \\

        \textit{w/} Monocular Depth
        & \cmark
        & M. Est.
        & \cmark
        & \cmark
        & 388.597
        & 4.447
        & \cellcolor{yellow!25}0.0020
        & 54.226
        & 24.132
        & 0.749
        & 0.255
        & 0.079 \\

        \textit{w/o} Dual-stream Warping
        & \cmark
        & VDC.
        & --
        & \cmark
        & 383.525
        & 4.412
        & 0.0049
        & 51.498
        & 23.886
        & 0.744
        & \cellcolor{yellow!25}0.254
        & 0.081 \\

        \textit{w/o} Refinement
        & \cmark
        & VDC.
        & \cmark
        & --
        & \cellcolor{yellow!25}347.922
        & \cellcolor{orange!25}4.521
        & 0.0042
        & \cellcolor{yellow!25}60.680
        & \cellcolor{orange!25}24.325
        & \cellcolor{yellow!25}0.756
        & \cellcolor{orange!25}0.239
        & \cellcolor{orange!25}0.076 \\
        \midrule

        \textbf{Full model}
        & \cmark
        & VDC.
        & \cmark
        & \cmark
        & \cellcolor{red!25}334.518
        & \cellcolor{red!25}4.549
        & \cellcolor{red!25}0.0010
        & \cellcolor{red!25}64.893
        & \cellcolor{red!25}24.360
        & \cellcolor{red!25}0.769
        & \cellcolor{red!25}0.237
        & \cellcolor{red!25}0.075 \\
        \bottomrule
    \end{tabular}%
    }
\end{table*}

\begin{figure*}[t]
    \centering
    \includegraphics[width=1\linewidth]{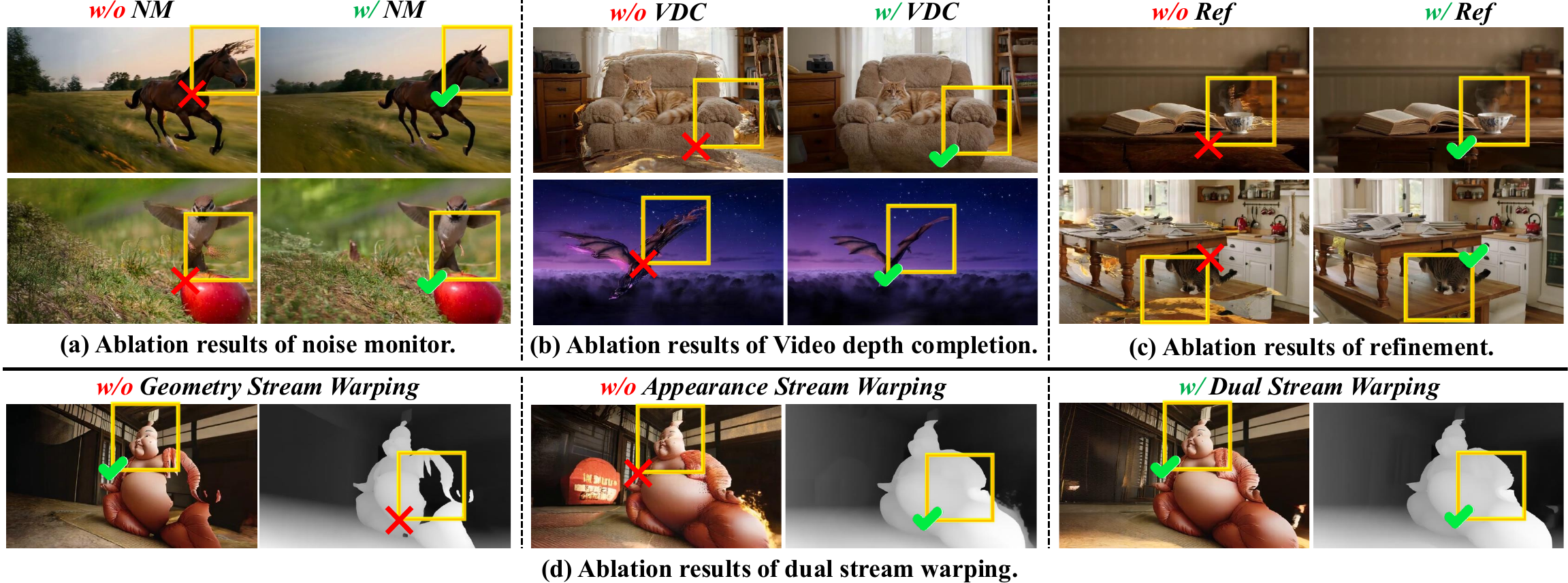}
    \caption{\textbf{Qualitative component ablation.} NM suppresses projection artifacts, VDC improves geometric consistency, dual-stream warping preserves both appearance and occlusion geometry, and refinement removes residual visual artifacts.}
    \label{fig:ablation_qualitaive_results}
\end{figure*}

Tab.~\ref{tab:component_ablation} and
Fig.~\ref{fig:ablation_qualitaive_results} validate the complementary roles of
our components. The noise monitor suppresses unreliable memory projections and
the resulting boundary artifacts. Replacing video depth completion with
monocular depth estimation increases P-MSE from 0.0010 to 0.0020 and decreases
View Sync from 64.893 to 54.226, demonstrating the importance of
memory-consistent geometry recovery. Dual-stream warping contributes most to
cross-trajectory consistency: removing it raises P-MSE to 0.0049 and reduces
View Sync to 51.498. As visualized in Fig.~\ref{fig:ablation_qualitaive_results}(d),
its appearance and geometry streams respectively preserve visual details and
reliable occlusion geometry. Finally, geometry-aware refinement further removes
residual artifacts and improves local details. Together, these components yield
high-quality generation with consistent appearance and geometry. Additional visualizations are provided in Appendix~\ref{appendix:Additional ablation results}.


\vspace{-4mm}
\section{Conclusion}
\vspace{-3mm}
We introduced \textbf{\textit{Demiurge}}, an autoregressive video rerendering
framework for constructing navigable 4D worlds from monocular videos. Unlike
existing methods that synthesize each camera trajectory independently, Demiurge
continuously consolidates generated appearance and recovered geometry into an
evolving 4D point-cloud memory through a generate--recover--update cycle. This
explicit memory enables reliable reuse of historical observations and preserves
long-range consistency across camera trajectories. Extensive experiments
establish state-of-the-art performance in perceptual quality, trajectory
accuracy, and cross-trajectory consistency, as well as demonstrating robust
large-parallax rerendering and geometry-faithful 3DGS reconstruction.

\subsection*{AI use statement}
Generative AI tools were used solely for grammar correction and linguistic refinement of author-written text. They were not used for research conceptualization, method development, code generation, experimental design, data analysis, or the formulation of scientific claims. All AI-assisted edits were carefully reviewed by the authors to preserve the intended technical meaning. The authors take full responsibility for the accuracy and integrity of the final manuscript.

\subsection*{Ethics statement}
Demiurge is developed for research on controllable video rerendering and 4D scene representation. As with other generative video methods, it may be misused to produce misleading content or process identifiable individuals without
consent, and it may inherit biases from pretrained models and training data. We encourage responsible use, disclosure of generated content, respect for
privacy and data licenses, and the adoption of provenance and watermarking mechanisms in practical deployments.

\subsection*{Reproducibility statement}
We provide detailed descriptions of the model architecture, training and inference procedures, 4D memory construction, depth alignment, dual-stream warping, and evaluation protocols in the main paper and supplementary
material. The supplementary material further specifies the metric implementations, runtime settings, and additional qualitative and ablation
results. Training data composition, model backbones, video resolution, sequence
length, and computational resources are also reported. Additional videos and supporting materials are available on our anonymous project page. We intend to release the source code, model checkpoints, and evaluation scripts to facilitate future reproduction and comparison.

\bibliography{iclr2027_conference}
\bibliographystyle{iclr2027_conference}

\appendix
\newpage
\startcontents
\section*{\LARGE Appendix for Demiurge: Forging Navigable 4D Worlds via Autoregressive Video Rerendering}

\printcontents{}{1}{
    \setcounter{tocdepth}{2}
}

\section{4D Point-Cloud Memory}
\label{sec:supp_4d_point_cloud}

\subsection{4D Point-Cloud Representation}
\label{sec:supp_4d_point_cloud_representation}

As introduced in the main paper, Demiurge represents a dynamic scene
using an evolving 4D point-cloud memory
$\mathcal{M}_{i}=\{\mathcal{P}_{i,t}\}_{t=1}^{T}$, where
$\mathcal{P}_{i,t}$ denotes the colored 3D point cloud accumulated at
time $t$ after processing the $i$-th target trajectory. Each
$\mathcal{P}_{i,t}$ represents an instantaneous 3D snapshot of the
scene, while their temporally ordered collection constitutes the
complete 4D representation.

Let $\mathbf{D}^{0}=\{\mathbf{D}_{t}^{0}\}_{t=1}^{T}$ denote the
depth video estimated from the input video
$\mathbf{V}^{0}=\{\mathbf{I}_{t}^{0}\}_{t=1}^{T}$. For a pixel
$\mathbf{p}=(u,v)$, we define its homogeneous image coordinate as
\begin{equation}
    \overline{\mathbf{p}}
    =
    \begin{bmatrix}
        u & v & 1
    \end{bmatrix}^{\mathsf T}.
\label{eq:supp_homogeneous_pixel}
\end{equation}
Following the camera convention used by our warping implementation,
a source-camera point is transformed into the shared world coordinate
system by the inverse camera transformation. Therefore, the
world-space position associated with pixel $\mathbf{p}$ in the
$t$-th input frame is
\begin{equation}
    \begin{bmatrix}
        \mathbf{X}_{t}^{0}(\mathbf{p})\\
        1
    \end{bmatrix}
    =
    \left(
        \mathbf{C}_{t}^{0}
    \right)^{-1}
    \begin{bmatrix}
        \mathbf{D}_{t}^{0}(\mathbf{p})
        \mathbf{K}^{-1}
        \overline{\mathbf{p}}\\
        1
    \end{bmatrix},
\label{eq:supp_initial_backprojection}
\end{equation}
where $\mathbf{K}$ is the camera intrinsic matrix. The initial colored
point cloud at time $t$ is then constructed as
\begin{equation}
    \mathcal{P}_{0,t}
    =
    \left\{
        \left(
            \mathbf{X}_{t}^{0}(\mathbf{p}),
            \mathbf{I}_{t}^{0}(\mathbf{p})
        \right)
        \;\middle|\;
        \mathbf{p}\in\Omega_{t}^{0}
    \right\},
\label{eq:supp_initial_point_cloud}
\end{equation}
where $\Omega_{t}^{0}$ denotes the set of valid pixels in the input
RGB-D frame. The initial 4D memory is consequently
\begin{equation}
    \mathcal{M}_{0}
    =
    \left\{
        \mathcal{P}_{0,t}
    \right\}_{t=1}^{T}.
\label{eq:supp_initial_4d_memory}
\end{equation}

After generating the $i$-th novel-view video
$\mathbf{V}^{i}=\{\mathbf{I}_{t}^{i}\}_{t=1}^{T}$ and recovering its
dense depth video
$\mathbf{D}^{i}=\{\mathbf{D}_{t}^{i}\}_{t=1}^{T}$, we similarly
back-project its valid RGB-D observations. For every
$\mathbf{p}\in\Omega_{t}^{i}$, the corresponding world-space point is
\begin{equation}
    \begin{bmatrix}
        \mathbf{X}_{t}^{i}(\mathbf{p})\\
        1
    \end{bmatrix}
    =
    \left(
        \mathbf{C}_{t}^{i}
    \right)^{-1}
    \begin{bmatrix}
        \mathbf{D}_{t}^{i}(\mathbf{p})
        \mathbf{K}^{-1}
        \overline{\mathbf{p}}\\
        1
    \end{bmatrix},
\label{eq:supp_generated_backprojection}
\end{equation}
where $\Omega_{t}^{i}$ denotes the set of valid pixels selected by
the memory-update procedure. These colored points are accumulated into
the point-cloud snapshot at the same timestamp:
\begin{equation}
\begin{aligned}
    \mathcal{P}_{i,t}
    =
    \mathcal{P}_{i-1,t}
    \uplus
    \left\{
        \left(
            \mathbf{X}_{t}^{i}(\mathbf{p}),
            \mathbf{I}_{t}^{i}(\mathbf{p})
        \right)
        \;\middle|\;
        \mathbf{p}\in\Omega_{t}^{i}
    \right\},
\end{aligned}
\label{eq:supp_point_cloud_update}
\end{equation}
where $\uplus$ denotes observation accumulation without geometric
averaging or point merging. In practice, we retain the RGB-D
observations, camera poses, and validity masks associated with
individual trajectories, and instantiate their corresponding colored
3D points on demand during rendering.

Importantly, an RGB-D observation at time $t$ is accumulated only
into $\mathcal{P}_{i,t}$ and never into
$\mathcal{P}_{i,t'}$ for $t'\neq t$. Consequently, a moving object
is represented at different 3D locations in the corresponding
point-cloud snapshots over time, rather than being duplicated within
a single static point cloud.

\subsection{Point-Cloud Warping}
\label{sec:supp_point_cloud_warping}

Although the memory is conceptually represented as a sequence of
colored point clouds, its factorized implementation allows us to
render each previously observed RGB-D trajectory independently and
then composite the warped results. Consider rendering the target
frame at time $t$ along the $i$-th trajectory. Let
$\mathbf{I}_{t}^{j}$, $\mathbf{D}_{t}^{j}$,
$\mathbf{M}_{t}^{j}$, and $\mathbf{C}_{t}^{j}$ denote the RGB frame,
depth map, validity mask, and camera pose associated with a previously
observed trajectory $j\in\{0,\ldots,i-1\}$.

For a source pixel $\mathbf{p}$, its source-camera 3D position is
\begin{equation}
    \mathbf{x}_{j,t}(\mathbf{p})
    =
    \mathbf{D}_{t}^{j}(\mathbf{p})
    \mathbf{K}^{-1}
    \overline{\mathbf{p}}.
\label{eq:supp_source_backprojection}
\end{equation}
Following the implementation, the relative transformation from the
source camera $\mathbf{C}_{t}^{j}$ to the target camera
$\mathbf{C}_{t}^{i}$ is
\begin{equation}
    \mathbf{T}_{j\rightarrow i,t}
    =
    \mathbf{C}_{t}^{i}
    \left(
        \mathbf{C}_{t}^{j}
    \right)^{-1}.
\label{eq:supp_relative_camera_transform}
\end{equation}
The source point is transformed into the target camera coordinate
system according to
\begin{equation}
    \begin{bmatrix}
        \mathbf{x}_{j\rightarrow i,t}(\mathbf{p})\\
        1
    \end{bmatrix}
    =
    \mathbf{T}_{j\rightarrow i,t}
    \begin{bmatrix}
        \mathbf{x}_{j,t}(\mathbf{p})\\
        1
    \end{bmatrix},
\label{eq:supp_target_camera_point}
\end{equation}
where
\begin{equation}
    \mathbf{x}_{j\rightarrow i,t}(\mathbf{p})
    =
    \begin{bmatrix}
        x_{j\rightarrow i,t}(\mathbf{p})\\
        y_{j\rightarrow i,t}(\mathbf{p})\\
        z_{j\rightarrow i,t}(\mathbf{p})
    \end{bmatrix}.
\end{equation}
Its homogeneous coordinate on the target image plane is
\begin{equation}
    \overline{\mathbf{p}}_{j\rightarrow i,t}
    =
    \mathbf{K}
    \mathbf{x}_{j\rightarrow i,t}(\mathbf{p}),
\label{eq:supp_homogeneous_target_projection}
\end{equation}
and the corresponding continuous pixel coordinate is
\begin{equation}
    \mathbf{p}_{j\rightarrow i,t}
    =
    \begin{bmatrix}
        \overline{p}_{x}/\overline{p}_{z}\\
        \overline{p}_{y}/\overline{p}_{z}
    \end{bmatrix}.
\label{eq:supp_target_projection}
\end{equation}
The depth of this point in the target camera coordinate system is
\begin{equation}
    z_{j\rightarrow i,t}(\mathbf{p})
    =
    \left[
        \mathbf{x}_{j\rightarrow i,t}(\mathbf{p})
    \right]_{3}.
\label{eq:supp_transformed_depth}
\end{equation}

\paragraph{Depth-aware bilinear splatting.}
Since $\mathbf{p}_{j\rightarrow i,t}(\mathbf{p})$ generally lies at
a non-integer image coordinate, we distribute its RGB and depth values
to the four neighboring target pixels through bilinear splatting.
Let
$\mathcal{N}(\mathbf{p}_{j\rightarrow i,t}(\mathbf{p}))$ denote these
four neighboring pixels. For an integer target pixel
$\mathbf{q}=(q_x,q_y)$ in this neighborhood, the bilinear proximity
weight is
\begin{equation}
\begin{aligned}
    b_{j,t}(\mathbf{p},\mathbf{q})
    =
    \left(
        1-
        \left|
            p_{j\rightarrow i,t}^{x}(\mathbf{p})-q_x
        \right|
    \right)
    \left(
        1-
        \left|
            p_{j\rightarrow i,t}^{y}(\mathbf{p})-q_y
        \right|
    \right).
\end{aligned}
\label{eq:supp_bilinear_weight}
\end{equation}

We additionally apply depth-aware weighting to suppress contributions
from distant points when multiple source pixels are splatted onto the
same target neighborhood. Specifically, we define
\begin{equation}
    \ell_{j,t}(\mathbf{p})
    =
    \log
    \left(
        1+
        \operatorname{clip}
        \left(
            z_{j\rightarrow i,t}(\mathbf{p}),
            0,
            1000
        \right)
    \right).
\label{eq:supp_log_depth}
\end{equation}
Let $\mathcal{B}$ denote the collection of source pixels included in
the current batched warping operation. The normalization factor used
by the implementation is
\begin{equation}
    \ell_{\mathcal{B}}^{\max}
    =
    \max_{(j,t,\mathbf{p})\in\mathcal{B}}
    \ell_{j,t}(\mathbf{p}).
\label{eq:supp_max_log_depth}
\end{equation}
The complete splatting weight is then
\begin{equation}
\begin{aligned}
    w_{j,t}(\mathbf{p},\mathbf{q})
    =
    b_{j,t}(\mathbf{p},\mathbf{q})\,
    \mathbf{M}_{t}^{j}(\mathbf{p})\,
    \exp
    \left(
        -50
        \frac{
            \ell_{j,t}(\mathbf{p})
        }{
            \ell_{\mathcal{B}}^{\max}
        }
    \right).
\end{aligned}
\label{eq:supp_depth_aware_splatting_weight}
\end{equation}

For each source trajectory $j$, the accumulated splatting weight at
target pixel $\mathbf{q}$ is
\begin{equation}
    s_{j,t}^{i}(\mathbf{q})
    =
    \sum_{\mathbf{p}:\,
    \mathbf{q}\in
    \mathcal{N}(\mathbf{p}_{j\rightarrow i,t}(\mathbf{p}))}
    w_{j,t}(\mathbf{p},\mathbf{q}).
\label{eq:supp_accumulated_splatting_weight}
\end{equation}
For target pixels satisfying
$s_{j,t}^{i}(\mathbf{q})>0$, the RGB and depth warped from source
trajectory $j$ are computed as
\begin{equation}
    \mathbf{I}_{j\rightarrow i,t}(\mathbf{q})
    =
    \frac{
        \displaystyle
        \sum_{\mathbf{p}:\,
        \mathbf{q}\in
        \mathcal{N}(\mathbf{p}_{j\rightarrow i,t}(\mathbf{p}))}
        w_{j,t}(\mathbf{p},\mathbf{q})
        \mathbf{I}_{t}^{j}(\mathbf{p})
    }{
        s_{j,t}^{i}(\mathbf{q})
    },
\label{eq:supp_rgb_splatting}
\end{equation}
and
\begin{equation}
    \mathbf{D}_{j\rightarrow i,t}(\mathbf{q})
    =
    \frac{
        \displaystyle
        \sum_{\mathbf{p}:\,
        \mathbf{q}\in
        \mathcal{N}(\mathbf{p}_{j\rightarrow i,t}(\mathbf{p}))}
        w_{j,t}(\mathbf{p},\mathbf{q})
        z_{j\rightarrow i,t}(\mathbf{p})
    }{
        s_{j,t}^{i}(\mathbf{q})
    }.
\label{eq:supp_depth_splatting}
\end{equation}
The corresponding warped validity mask is
\begin{equation}
    \mathbf{M}_{j\rightarrow i,t}(\mathbf{q})
    =
    \mathbb{I}
    \left[
        s_{j,t}^{i}(\mathbf{q})>0
    \right].
\label{eq:supp_splatting_mask}
\end{equation}

\paragraph{Cross-trajectory z-buffer composition.}
After independently warping all previously observed trajectories, we
combine their results using a nearest-depth z-buffer. For each target
pixel $\mathbf{q}$, the set of valid warped sources is
\begin{equation}
    \mathcal{A}_{t}^{i}(\mathbf{q})
    =
    \left\{
        j
        \;\middle|\;
        \mathbf{M}_{j\rightarrow i,t}(\mathbf{q})>0.5,
        \quad
        \mathbf{D}_{j\rightarrow i,t}(\mathbf{q})>\tau_{0}
    \right\},
\label{eq:supp_valid_warped_sources}
\end{equation}
where $\tau_{0}=10^{-5}$. If multiple warped sources cover the same
target pixel, we select the source with the smallest warped depth:
\begin{equation}
    j_{t}^{i,*}(\mathbf{q})
    =
    \underset{
        j\in\mathcal{A}_{t}^{i}(\mathbf{q})
    }{\arg\min}
    \;
    \mathbf{D}_{j\rightarrow i,t}(\mathbf{q}).
\label{eq:supp_cross_trajectory_zbuffer}
\end{equation}

The resulting point-cloud rendering at target pixel $\mathbf{q}$ is
defined as
\begin{equation}
\left(
    \mathbf{I}_{\mathrm{pc},t}^{i}(\mathbf{q}),
    \mathbf{D}_{\mathrm{pc},t}^{i}(\mathbf{q}),
    \mathbf{M}_{\mathrm{pc},t}^{i}(\mathbf{q})
\right)
=
\begin{cases}
\left(
    \mathbf{I}_{j_{t}^{i,*}(\mathbf{q})\rightarrow i,t}(\mathbf{q}),
    \mathbf{D}_{j_{t}^{i,*}(\mathbf{q})\rightarrow i,t}(\mathbf{q}),
    1
\right),
&
\mathcal{A}_{t}^{i}(\mathbf{q})\neq\varnothing,\\[3pt]
\left(
    \mathbf{0},
    0,
    0
\right),
&
\mathcal{A}_{t}^{i}(\mathbf{q})=\varnothing.
\end{cases}
\label{eq:supp_cross_trajectory_composition}
\end{equation}

Accordingly, the elementary point-cloud rendering operator is
\begin{equation}
\left(
    \mathbf{I}_{\mathrm{pc},t}^{i},
    \mathbf{D}_{\mathrm{pc},t}^{i},
    \mathbf{M}_{\mathrm{pc},t}^{i}
\right)
=
\mathcal{R}
\left(
    \mathcal{P}_{i-1,t},
    \mathbf{C}_{t}^{i}
\right),
\label{eq:supp_point_cloud_rendering}
\end{equation}
where the subscript $\mathrm{pc}$ denotes the RGB, depth, and visibility outputs of point-cloud rendering, $\mathcal{R}$ consists of the depth-aware bilinear splatting
defined in
Eq.~\ref{eq:supp_source_backprojection}--
Eq.~\ref{eq:supp_splatting_mask}, followed by the cross-trajectory
z-buffer composition defined in
Eq.~\ref{eq:supp_valid_warped_sources}--
Eq.~\ref{eq:supp_cross_trajectory_composition}. Equivalently,
\begin{equation}
\begin{aligned}
    \mathcal{R}
    \left(
        \mathcal{P}_{i-1,t},
        \mathbf{C}_{t}^{i}
    \right)
    =
    \operatorname{ZBuf}
    \left(
        \left\{
        \mathcal{W}
        \left(
            \mathbf{I}_{t}^{j},
            \mathbf{D}_{t}^{j},
            \mathbf{M}_{t}^{j},
            \mathbf{C}_{t}^{j},
            \mathbf{C}_{t}^{i},
            \mathbf{K}
        \right)
        \right\}_{j=0}^{i-1}
    \right),
\end{aligned}
\label{eq:supp_explicit_rendering_operator}
\end{equation}
where $\mathcal{W}$ denotes the depth-aware bilinear forward-warping
operation for an individual historical trajectory.

The rendering operation is applied independently at every timestamp.
In particular, rendering the target frame at time $t$ involves only
$\mathcal{P}_{i-1,t}$ and $\mathbf{C}_{t}^{i}$. No point from
$\mathcal{P}_{i-1,t'}$, where $t'\neq t$, participates in the
rendering. Dynamic content is therefore represented by its evolving
geometry across temporally indexed point-cloud snapshots, allowing
static backgrounds and moving objects to be handled uniformly without
explicit dynamic--static decomposition.

\section{Additional Method Details}
\label{sec:supp_implementation}

\subsection{Noise Monitor}
\label{sec:supp_noise_monitor}

We provide the implementation details of the non-learned noise
monitor introduced in the main paper. Given a rendered depth frame,
we identify unreliable pixels using three complementary criteria:
invalid depth, large depth gradients, and high local depth variance.
We set the corresponding thresholds to
\begin{equation}
    \tau_{0}=10^{-5},
    \qquad
    \tau_{g}=0.03,
    \qquad
    \tau_{v}=10^{-4}.
\end{equation}

The horizontal and vertical depth derivatives are computed using
finite differences, and their magnitude is given by
\begin{equation}
    G_{t}^{i}(\mathbf{p})
    =
    \sqrt{
        \left(
            \nabla_x
            \widetilde{\mathbf{D}}_{t}^{i}(\mathbf{p})
        \right)^2
        +
        \left(
            \nabla_y
            \widetilde{\mathbf{D}}_{t}^{i}(\mathbf{p})
        \right)^2
    }.
\end{equation}
The local depth variance is computed over a $3\times3$ neighborhood
using
\begin{equation}
\begin{aligned}
    S_{t}^{i}(\mathbf{p})
    =
    \operatorname{Mean}_{\mathcal{N}_{3}(\mathbf{p})}
    \left[
        \left(
            \widetilde{\mathbf{D}}_{t}^{i}
        \right)^2
    \right]
    -
    \left(
        \operatorname{Mean}_{\mathcal{N}_{3}(\mathbf{p})}
        \left[
            \widetilde{\mathbf{D}}_{t}^{i}
        \right]
    \right)^2.
\end{aligned}
\end{equation}
A pixel is marked as unreliable if it satisfies any of the three
criteria. The resulting noise mask is simultaneously applied to the
rendered RGB, depth, and visibility conditions, ensuring that the
video diffusion model does not receive mutually inconsistent
conditioning signals. In our default setting, the cleaning operation
is applied once after point-cloud rendering. The monitor is fully
deterministic, introduces no learnable parameters, and incurs
negligible computational overhead.

\subsection{Cache-Guided Appearance Video Diffusion}
\label{sec:supp_appearance_diffusion}

Our cache-guided appearance model
$\mathcal{F}_{\mathrm{app}}$ is initialized from the pretrained
Wan2.1-Fun-1.3B-InP video diffusion model. As described in the main
paper, it takes the cleaned partial RGB video
$\overline{\mathbf{V}}^{i}$, partial depth video
$\overline{\mathbf{D}}^{i}$, and visibility mask
$\overline{\mathbf{M}}^{i}$ as conditions and synthesizes the
novel-view video $\mathbf{V}^{i}$ along the target camera trajectory.

\paragraph{Latent condition encoding.}
We first normalize the three conditioning videos to the input range
of the Wan VAE. Since the depth and visibility conditions are
single-channel signals, each of them is replicated over three
channels before VAE encoding. Let $\mathcal{E}_{\mathrm{vae}}$ denote
the frozen Wan VAE encoder. The condition latents are given by
\begin{equation}
\begin{aligned}
    \mathbf{z}_{\mathrm{rgb}}^{i}
    &=
    \mathcal{E}_{\mathrm{vae}}
    \left(
        \overline{\mathbf{V}}^{i}
    \right),\\
    \mathbf{z}_{\mathrm{dep}}^{i}
    &=
    \mathcal{E}_{\mathrm{vae}}
    \left(
        \operatorname{Rep}_{3}
        \left(
            \overline{\mathbf{D}}^{i}
        \right)
    \right),\\
    \mathbf{z}_{\mathrm{mask}}^{i}
    &=
    \mathcal{E}_{\mathrm{vae}}
    \left(
        \operatorname{Rep}_{3}
        \left(
            \overline{\mathbf{M}}^{i}
        \right)
    \right),
\end{aligned}
\label{eq:supp_app_condition_encoding}
\end{equation}
where $\operatorname{Rep}_{3}(\cdot)$ replicates a single-channel
video over the RGB channels. All three conditions are therefore
encoded into the same spatio-temporal latent space as the target
video.

\paragraph{Latent-space condition injection.}
Let $\mathbf{z}_{\tau}^{i}$ denote the noisy target RGB latent at
diffusion timestep $\tau$. We inject the cache-derived conditions
through channel-wise latent concatenation:
\begin{equation}
    \mathbf{h}_{\tau}^{i}
    =
    \operatorname{Concat}_{c}
    \left[
        \mathbf{z}_{\tau}^{i},
        \mathbf{z}_{\mathrm{rgb}}^{i},
        \mathbf{z}_{\mathrm{dep}}^{i},
        \mathbf{z}_{\mathrm{mask}}^{i}
    \right],
\label{eq:supp_app_latent_concatenation}
\end{equation}
where $\operatorname{Concat}_{c}$ denotes concatenation along the
channel dimension. The Wan VAE produces a 16-channel latent for each
video. Consequently, $\mathbf{h}_{\tau}^{i}$ contains 64 channels,
including 16 noisy target channels and 48 conditioning channels.

To accommodate this expanded input, we replace the original input
projection with a 3D convolutional patch embedding
$\mathcal{P}_{\mathrm{app}}$ that maps the 64-channel latent volume
to the hidden dimension of the pretrained Wan DiT:
\begin{equation}
    \mathbf{x}_{\tau}^{i}
    =
    \mathcal{P}_{\mathrm{app}}
    \left(
        \mathbf{h}_{\tau}^{i}
    \right).
\label{eq:supp_app_patch_embedding}
\end{equation}
The kernel size and stride of $\mathcal{P}_{\mathrm{app}}$ follow
the original spatio-temporal patchification configuration of the Wan
backbone.

\paragraph{Video diffusion backbone.}
The resulting tokens are processed by the original Wan DiT blocks.
Each block consists of spatio-temporal self-attention, text
cross-attention, and a feed-forward network. Three-dimensional rotary
position embeddings are applied along the temporal, height, and width
axes, while the diffusion timestep modulates the self-attention and
feed-forward layers. The caption of the input video is encoded by the
UMT5 text encoder and injected into all DiT blocks through
cross-attention.

\subsection{Cache-Consistent Video Depth Completion}
\label{sec:supp_depth_completion}

We provide additional architectural details of the cache-consistent
video depth completion model $\mathcal{F}_{\mathrm{dep}}$ introduced
in the main paper. The model is initialized from
Wan2.1-Fun-1.3B-InP and adopts a dual-path conditioning architecture
consisting of a geometry-anchored denoising branch and an auxiliary
RGB guidance branch.

\paragraph{Condition encoding.}
Following Eq.~\ref{eq:depth_to_disparity}, the cleaned partial cache
depth $\overline{\mathbf{D}}^{i}$ is converted into the normalized
partial disparity $\overline{\mathbf{U}}^{i}$. We independently
encode $\overline{\mathbf{U}}^{i}$ and the generated RGB video
$\mathbf{V}^{i}$ using the pretrained Wan VAE encoder
$\mathcal{E}_{\mathrm{vae}}$:
\begin{equation}
\begin{aligned}
    \mathbf{z}_{\mathrm{partial}}^{i}
    &=
    \mathcal{E}_{\mathrm{vae}}
    \left(
        \operatorname{Rep}_{3}
        \left(
            \overline{\mathbf{U}}^{i}
        \right)
    \right),\\
    \mathbf{z}_{\mathrm{rgb}}^{i}
    &=
    \mathcal{E}_{\mathrm{vae}}
    \left(
        \mathbf{V}^{i}
    \right),
\end{aligned}
\label{eq:supp_depth_condition_encoding}
\end{equation}
where $\operatorname{Rep}_{3}(\cdot)$ replicates the single-channel
partial disparity over three channels before VAE encoding. Both
$\mathbf{z}_{\mathrm{partial}}^{i}$ and
$\mathbf{z}_{\mathrm{rgb}}^{i}$ contain 16 latent channels.

\paragraph{Geometry-anchored denoising branch.}
Let $\mathbf{z}_{\tau}^{i}$ denote the noisy disparity latent at
diffusion timestep $\tau$. Following the Wan inpainting interface, we
concatenate it with the partial-disparity latent and a four-channel
inpainting indicator:
\begin{equation}
    \mathbf{h}_{\tau}^{i}
    =
    \operatorname{Concat}_{c}
    \left[
        \mathbf{z}_{\tau}^{i},
        \mathbf{m}_{\mathrm{inp}},
        \mathbf{z}_{\mathrm{partial}}^{i}
    \right],
\label{eq:supp_depth_main_input}
\end{equation}
where $\operatorname{Concat}_{c}$ denotes channel-wise concatenation
and $\mathbf{m}_{\mathrm{inp}}$ is set to one over the complete latent
volume. The resulting 36-channel tensor consists of 16 noisy disparity
channels, four inpainting-indicator channels, and 16 partial-disparity
channels.

The main-branch tokens are obtained through the original
spatio-temporal patch embedding of the Wan inpainting model:
\begin{equation}
    \mathbf{x}_{0}^{i}
    =
    \mathcal{E}_{\mathrm{patch}}^{\mathrm{dep}}
    \left(
        \mathbf{h}_{\tau}^{i}
    \right).
\label{eq:supp_depth_main_embedding}
\end{equation}
This branch directly incorporates the partial disparity into the main
denoising stream, thereby anchoring the prediction to the geometry
accumulated in the 4D memory.

\paragraph{Auxiliary RGB guidance branch.}
The generated RGB video provides complementary semantic, structural,
and temporal cues for regions not covered by the partial cache
disparity. Following our implementation, we replicate its 16-channel
latent three times along the channel dimension and process the
resulting 48-channel tensor using an additional spatio-temporal patch
embedding:
\begin{equation}
    \mathbf{g}_{0}^{i}
    =
    \mathcal{E}_{\mathrm{patch}}^{\mathrm{rgb}}
    \left(
        \operatorname{Concat}_{c}
        \left[
            \mathbf{z}_{\mathrm{rgb}}^{i},
            \mathbf{z}_{\mathrm{rgb}}^{i},
            \mathbf{z}_{\mathrm{rgb}}^{i}
        \right]
    \right).
\label{eq:supp_depth_rgb_embedding}
\end{equation}

The auxiliary branch contains six DiT blocks and progressively updates
the RGB guidance features at every fifth layer of the main Wan
backbone. Before each main DiT block, an RGB guidance feature is
projected through a learned linear layer and added to the main hidden
state. Let $\mathbf{x}_{\ell}^{i}$ denote the hidden state before the
$\ell$-th main DiT block, and let $\mathbf{r}_{\ell}^{i}$ denote the
projected RGB guidance feature injected at this layer. The feature
injection and backbone update are expressed as
\begin{equation}
\begin{aligned}
    \widehat{\mathbf{x}}_{\ell}^{i}
    &=
    \mathbf{x}_{\ell}^{i}
    +
    \mathbf{r}_{\ell}^{i},\\
    \mathbf{x}_{\ell+1}^{i}
    &=
    \mathcal{B}_{\ell}
    \left(
        \widehat{\mathbf{x}}_{\ell}^{i},
        \mathbf{e}_{\mathrm{text}},
        \tau
    \right),
\end{aligned}
\label{eq:supp_depth_feature_injection}
\end{equation}
where $\widehat{\mathbf{x}}_{\ell}^{i}$ is the hidden state after RGB
guidance injection, $\mathcal{B}_{\ell}$ denotes the $\ell$-th main
DiT block, and $\mathbf{e}_{\mathrm{text}}$ is the embedding of the
input-video caption. At layers
$\ell\in\{0,5,10,15,20,25\}$, the RGB guidance features are first
updated by the corresponding auxiliary DiT block; the remaining
layers receive learned linear projections of the initial RGB guidance
tokens. The main and auxiliary branches use the same diffusion
timestep conditioning, text context, and three-dimensional rotary
position embeddings.

\paragraph{Dense disparity decoding.}
After flow-matching denoising, the clean disparity latent
$\mathbf{z}_{0}^{i}$ is decoded using the Wan VAE decoder
$\mathcal{D}_{\mathrm{vae}}$. Since the decoder produces a
three-channel output, we average its channels and apply min--max
normalization to obtain the dense relative disparity:
\begin{equation}
    \widehat{\mathbf{U}}^{i}
    =
    \mathcal{N}
    \left(
        \frac{1}{3}
        \sum_{c=1}^{3}
        \left[
            \mathcal{D}_{\mathrm{vae}}
            \left(
                \mathbf{z}_{0}^{i}
            \right)
        \right]_{c}
    \right).
\label{eq:supp_depth_decoding}
\end{equation}
The predicted disparity $\widehat{\mathbf{U}}^{i}$ is subsequently
converted to metric depth through the two-stage global alignment
described below, followed by Poisson refinement to obtain the final
cache-consistent dense depth video $\mathbf{D}^{i}$.

\paragraph{Cache-supported region.}
Let $\overline{\mathbf{D}}^{i}$ denote the partial metric depth
rendered from the 4D memory. We define the cache-supported region as
\begin{equation}
    \Omega^{i}
    =
    \left\{
        (t,\mathbf{p})
        \;\middle|\;
        \overline{\mathbf{M}}_{t}^{i}(\mathbf{p})=1,
        \quad
        \overline{\mathbf{D}}_{t}^{i}(\mathbf{p})
        >
        \tau_{\mathrm{valid}}
    \right\},
\label{eq:supp_alignment_region}
\end{equation}
where $\tau_{\mathrm{valid}}=10^{-6}$. Only pixels that are both
visible in the cache rendering and associated with valid depth values
are used as geometric anchors. No additional mask erosion is applied
in our default configuration.

\paragraph{Global affine alignment.}
In the first stage, we normalize the valid values of the partial cache
depth over the complete video:
\begin{equation}
    \overline{\mathbf{D}}_{\mathrm{n}}^{i}
    =
    \mathcal{N}
    \left(
        \overline{\mathbf{D}}^{i};
        \Omega^{i}
    \right),
\label{eq:supp_partial_depth_normalization}
\end{equation}
where $\mathcal{N}(\cdot;\Omega^{i})$ denotes min--max normalization
over $\Omega^{i}$. To reduce the influence of outliers, we retain only
values between the $2\%$ and $98\%$ quantiles of
$\overline{\mathbf{D}}_{\mathrm{n}}^{i}$. Denoting the corresponding
inlier set by $\Omega_{\mathrm{in}}^{i}$, we estimate a global affine
transformation shared across all frames:
\begin{equation}
\begin{aligned}
    \left(
        a_{1}^{i},
        b_{1}^{i}
    \right)
    =
    \underset{a,b}{\arg\min}\quad
    &
    \sum_{(t,\mathbf{p})\in\Omega_{\mathrm{in}}^{i}}
    \left(
        a\widehat{\mathbf{U}}_{t}^{i}(\mathbf{p})
        +b
        -
        \overline{\mathbf{D}}_{\mathrm{n},t}^{i}(\mathbf{p})
    \right)^{2}
    \\
    &+
    \lambda_{\mathrm{ridge}}
    \left(
        a^{2}+b^{2}
    \right),
\end{aligned}
\label{eq:supp_stage1_alignment}
\end{equation}
where $\lambda_{\mathrm{ridge}}=10^{-4}$. If fewer than 100 inlier
pixels are available, we use the identity transformation. The aligned
normalized prediction is
\begin{equation}
    \mathbf{A}^{i}
    =
    \operatorname{clip}
    \left(
        a_{1}^{i}\widehat{\mathbf{U}}^{i}
        +b_{1}^{i},
        0,
        1
    \right).
\label{eq:supp_normalized_aligned_prediction}
\end{equation}

In the second stage, we recover the metric scale by fitting
$\mathbf{A}^{i}$ to the original partial cache depth:
\begin{equation}
    \left(
        a_{2}^{i},
        c_{2}^{i}
    \right)
    =
    \underset{a,c}{\arg\min}
    \sum_{(t,\mathbf{p})\in\Omega^{i}}
    \left(
        a\mathbf{A}_{t}^{i}(\mathbf{p})
        +c
        -
        \overline{\mathbf{D}}_{t}^{i}(\mathbf{p})
    \right)^{2}.
\label{eq:supp_stage2_metric_alignment}
\end{equation}
This regression uses all cache-supported pixels without quantile
trimming or ridge regularization. The resulting dense metric-depth
prediction is
\begin{equation}
    \mathbf{G}^{i}
    =
    a_{2}^{i}\mathbf{A}^{i}
    +
    c_{2}^{i}.
\label{eq:supp_aligned_metric_depth}
\end{equation}
Both affine transformations are shared across the complete video,
which avoids frame-dependent scale fluctuations. We do not directly
overwrite $\mathbf{G}^{i}$ in cache-supported regions; exact
consistency is imposed by the subsequent Poisson refinement.

\paragraph{Poisson refinement.}
We solve the metric-depth Poisson objective in
Eq.~\ref{eq:poisson_refinement} independently for each frame. The
solution is initialized using the partial cache depth in known regions
and the aligned prediction elsewhere:
\begin{equation}
    \mathbf{X}_{t}^{(0)}(\mathbf{p})
    =
    \begin{cases}
        \overline{\mathbf{D}}_{t}^{i}(\mathbf{p}),
        &
        (t,\mathbf{p})\in\Omega^{i},\\
        \mathbf{G}_{t}^{i}(\mathbf{p}),
        &
        (t,\mathbf{p})\notin\Omega^{i}.
    \end{cases}
\label{eq:supp_poisson_initialization}
\end{equation}

We discretize the Poisson equation using a four-connected
neighborhood. For an unknown pixel
$\mathbf{p}\notin\Omega_{t}^{i}$, one Jacobi iteration is
\begin{equation}
\begin{aligned}
    \mathbf{X}_{t}^{(k+1)}(\mathbf{p})
    =
    \frac{1}{4}
    \Bigg[
        \sum_{\mathbf{q}\in\mathcal{N}_{4}(\mathbf{p})}
        \mathbf{X}_{t}^{(k)}(\mathbf{q})
        +
        4\mathbf{G}_{t}^{i}(\mathbf{p}) -\sum_{\mathbf{q}\in\mathcal{N}_{4}(\mathbf{p})}
        \mathbf{G}_{t}^{i}(\mathbf{q})
    \Bigg],
\end{aligned}
\label{eq:supp_poisson_jacobi}
\end{equation}
where $\mathcal{N}_{4}(\mathbf{p})$ denotes the four-connected
neighborhood. After every iteration, cache-supported pixels are reset
to their metric cache-depth values:
\begin{equation}
    \mathbf{X}_{t}^{(k+1)}(\mathbf{p})
    =
    \overline{\mathbf{D}}_{t}^{i}(\mathbf{p}),
    \qquad
    \forall\,
    (t,\mathbf{p})\in\Omega^{i}.
\label{eq:supp_poisson_constraint}
\end{equation}

We use replicated padding at image boundaries and perform 800 Jacobi
iterations. For frames without cache-supported pixels,
$\mathbf{G}_{t}^{i}$ is retained unchanged. The final dense
metric-depth video is
\begin{equation}
    \mathbf{D}^{i}
    =
    \left\{
        \mathbf{X}_{t}^{(800)}
    \right\}_{t=1}^{T}.
\label{eq:supp_final_metric_depth}
\end{equation}
The two-stage alignment recovers the metric scale of the 4D memory,
while the Poisson refinement preserves the completed depth structure
and enforces exact consistency with cached geometry.

\subsection{Dual-Stream Warping}
\label{sec:supp_dual_stream_warping}

As described in the main paper, the appearance and geometry streams
are two logical views of the same accumulated 4D memory rather than
two physically duplicated point-cloud representations. Both streams
share the RGB-D observations, camera poses, and provenance masks of
the input video and previously generated trajectories. Their
difference lies only in the source-selection masks used during
point-cloud warping.

\paragraph{Stream-specific source masks.}
Let $\mathbf{M}_{t}^{j}$ denote the provenance mask stored for the
$j$-th generated trajectory at time $t$, where visible cache-supported
regions are marked as one. For the original input trajectory
$j=0$, all pixels are treated as valid in both streams. For historical
generated trajectories $j>0$, the source masks of the two streams are
defined as
\begin{equation}
\begin{aligned}
    \mathbf{S}_{\mathrm{app},t}^{0}
    &=
    \mathbf{S}_{\mathrm{geo},t}^{0}
    =
    \mathbf{1},\\
    \mathbf{S}_{\mathrm{app},t}^{j}
    &=
    \mathbf{1}
    -
    \operatorname{Dilate}_{3\times3}
    \left(
        \mathbf{M}_{t}^{j}
    \right),\\
    \mathbf{S}_{\mathrm{geo},t}^{j}
    &=
    \mathbf{1},
    \qquad j>0.
\end{aligned}
\label{eq:supp_dual_stream_masks}
\end{equation}
We apply one $3\times3$ dilation to the stored provenance mask before
inversion. Consequently, the appearance stream retains the original
input together with newly synthesized regions from previous
trajectories, whereas the geometry stream uses all valid RGB-D
observations.

\paragraph{Parallel point-cloud warping.}
Using the elementary warping operator $\mathcal{W}$ and
cross-trajectory z-buffer composition introduced in
Sec.~\ref{sec:supp_point_cloud_warping}, the two streams are rendered
independently:
\begin{equation}
\begin{aligned}
    \left(
        \mathbf{V}_{\mathrm{app}}^{i},
        \mathbf{D}_{\mathrm{app}}^{i},
        \mathbf{M}_{\mathrm{app}}^{i}
    \right)
    &=
    \operatorname{ZBuf}
    \left(
        \left\{
            \mathcal{W}
            \left(
                \mathbf{I}_{t}^{j},
                \mathbf{D}_{t}^{j},
                \mathbf{S}_{\mathrm{app},t}^{j},
                \mathbf{C}_{t}^{j},
                \mathbf{C}_{t}^{i},
                \mathbf{K}
            \right)
        \right\}_{j=0}^{i-1}
    \right)_{t=1}^{T},\\
    \left(
        \mathbf{V}_{\mathrm{geo}}^{i},
        \mathbf{D}_{\mathrm{geo}}^{i},
        \mathbf{M}_{\mathrm{geo}}^{i}
    \right)
    &=
    \operatorname{ZBuf}
    \left(
        \left\{
            \mathcal{W}
            \left(
                \mathbf{I}_{t}^{j},
                \mathbf{D}_{t}^{j},
                \mathbf{S}_{\mathrm{geo},t}^{j},
                \mathbf{C}_{t}^{j},
                \mathbf{C}_{t}^{i},
                \mathbf{K}
            \right)
        \right\}_{j=0}^{i-1}
    \right)_{t=1}^{T}.
\end{aligned}
\label{eq:supp_dual_stream_rendering}
\end{equation}
Both streams use the same depth-aware bilinear splatting and
nearest-depth z-buffer. In our implementation, the observations are
processed in chunks of 16 frames and eight source trajectories to
reduce peak memory consumption without changing the rendering result.

\paragraph{Depth-conflict detection.}
The rendered metric depths are converted to the bounded inverse-depth
representation
\begin{equation}
    \phi(d)
    =
    \frac{1}{d+1}.
\label{eq:supp_bounded_inverse_depth}
\end{equation}
We identify conflicts by comparing the quantized inverse depths of
the two streams:
\begin{equation}
\begin{aligned}
    \mathbf{Q}_{t}^{i}(\mathbf{p})
    =
    \mathbb{I}
    \Big[
        \big|
            \operatorname{round}
            \left(
                255\,
                \phi(
                    \mathbf{D}_{\mathrm{app},t}^{i}(\mathbf{p})
                )
            \right)
            -
            \operatorname{round}
            \left(
                255\,
                \phi(
                    \mathbf{D}_{\mathrm{geo},t}^{i}(\mathbf{p})
                )
            \right)
        \big|
        >
        15
    \Big].
\end{aligned}
\label{eq:supp_dual_stream_conflict}
\end{equation}
Thus, the discrepancy threshold corresponds approximately to
$\tau_{\mathrm{occ}}=15/255$ in the normalized inverse-depth domain.

\paragraph{Pixel-wise stream selection.}
At pixels without detected conflicts, we preserve the sharper
appearance-stream result. At conflicting pixels, we select the
geometry stream, whose denser point coverage provides more reliable
visibility reasoning:
\begin{equation}
\begin{aligned}
    \widetilde{\mathbf{V}}^{i}
    &=
    \left(
        1-\mathbf{Q}^{i}
    \right)
    \odot
    \mathbf{V}_{\mathrm{app}}^{i}
    +
    \mathbf{Q}^{i}
    \odot
    \mathbf{V}_{\mathrm{geo}}^{i},\\
    \widetilde{\mathbf{D}}^{i}
    &=
    \left(
        1-\mathbf{Q}^{i}
    \right)
    \odot
    \mathbf{D}_{\mathrm{app}}^{i}
    +
    \mathbf{Q}^{i}
    \odot
    \mathbf{D}_{\mathrm{geo}}^{i},\\
    \widetilde{\mathbf{M}}^{i}
    &=
    \left(
        1-\mathbf{Q}^{i}
    \right)
    \odot
    \mathbf{M}_{\mathrm{app}}^{i}
    +
    \mathbf{Q}^{i}
    \odot
    \mathbf{M}_{\mathrm{geo}}^{i}.
\end{aligned}
\label{eq:supp_dual_stream_selection}
\end{equation}
The resulting
$(\widetilde{\mathbf{V}}^{i},
\widetilde{\mathbf{D}}^{i},
\widetilde{\mathbf{M}}^{i})$
correspond exactly to the output of
$\mathcal{R}_{\mathrm{dual}}$ in the main paper.

\subsection{Geometry-Aware Video Refinement}
\label{sec:supp_geometry_refinement}

We provide the architectural details of the geometry-aware video
refinement model $\mathcal{F}_{\mathrm{ref}}$ introduced in the main
paper. The model is initialized from Wan2.1-Fun-1.3B-InP and follows
the same latent video diffusion formulation as
$\mathcal{F}_{\mathrm{app}}$. It complements the initial appearance
generation stage by using the recovered dense geometry
$\mathbf{D}^{i}$ to suppress residual texture flicker and
high-frequency artifacts before the generated RGB-D observations are
written into the 4D memory.

\paragraph{Geometry-aware condition encoding.}
The refinement model takes the cleaned partial RGB rendering
$\overline{\mathbf{V}}^{i}$ and completed dense depth video
$\mathbf{D}^{i}$ as its effective conditions. Both videos are encoded
using the pretrained Wan VAE encoder:
\begin{equation}
\begin{aligned}
    \mathbf{z}_{\mathrm{rgb,ref}}^{i}
    &=
    \mathcal{E}_{\mathrm{vae}}
    \left(
        \overline{\mathbf{V}}^{i}
    \right),\\
    \mathbf{z}_{\mathrm{dep,ref}}^{i}
    &=
    \mathcal{E}_{\mathrm{vae}}
    \left(
        \operatorname{Rep}_{3}
        \left(
            \mathbf{D}^{i}
        \right)
    \right),
\end{aligned}
\label{eq:supp_refinement_condition_encoding}
\end{equation}
where $\operatorname{Rep}_{3}(\cdot)$ replicates the single-channel
depth video over three channels before VAE encoding. The two
conditions are thus represented in the same spatio-temporal latent
space as the target RGB video.

\paragraph{Latent-space condition injection.}
Let $\mathbf{z}_{\tau,\mathrm{ref}}^{i}$ denote the noisy RGB latent
at diffusion timestep $\tau$. Following the Wan inpainting interface,
we concatenate it with the partial-RGB latent, dense-depth latent, and
a constant conditioning latent:
\begin{equation}
    \mathbf{h}_{\tau,\mathrm{ref}}^{i}
    =
    \operatorname{Concat}_{c}
    \left[
        \mathbf{z}_{\tau,\mathrm{ref}}^{i},
        \mathbf{z}_{\mathrm{rgb,ref}}^{i},
        \mathbf{z}_{\mathrm{dep,ref}}^{i},
        \mathbf{z}_{\mathrm{const}}
    \right],
\label{eq:supp_refinement_latent_input}
\end{equation}
where $\mathbf{z}_{\mathrm{const}}$ is obtained by encoding an all-one
video and preserves the conditioning layout of the pretrained
inpainting model without introducing additional scene information.
Since each latent contains 16 channels, the resulting input contains
64 channels.

The concatenated latent is mapped to the Wan DiT hidden dimension
through a 3D convolutional patch embedding:
\begin{equation}
    \mathbf{x}_{\tau,\mathrm{ref}}^{i}
    =
    \mathcal{E}_{\mathrm{patch}}^{\mathrm{ref}}
    \left(
        \mathbf{h}_{\tau,\mathrm{ref}}^{i}
    \right).
\label{eq:supp_refinement_patch_embedding}
\end{equation}
The resulting tokens are processed by the Wan DiT blocks using full
spatio-temporal self-attention, text cross-attention, diffusion
timestep modulation, and three-dimensional rotary position
embeddings. After flow-matching denoising, the clean RGB latent is
decoded by the Wan VAE decoder, producing the refined video
$\mathbf{V}_{\mathrm{ref}}^{i}$.

\paragraph{Memory update.}
The partial RGB rendering preserves appearance already established in
the 4D memory, while the completed depth provides dense structural
guidance across both previously observed and newly revealed regions.
Their combination allows $\mathcal{F}_{\mathrm{ref}}$ to improve the
appearance of $\mathbf{V}^{i}$ without modifying its recovered depth
$\mathbf{D}^{i}$.

When refinement is enabled, the refined video replaces the initial
appearance-generation result during memory update:
\begin{equation}
    \mathbf{V}_{\mathrm{upd}}^{i}
    =
    \begin{cases}
        \mathbf{V}_{\mathrm{ref}}^{i},
        & \text{with refinement},\\
        \mathbf{V}^{i},
        & \text{without refinement}.
    \end{cases}
\label{eq:supp_memory_update_video}
\end{equation}
The RGB-D pair
$(\mathbf{V}_{\mathrm{upd}}^{i},\mathbf{D}^{i})$ is subsequently
back-projected and fused into the 4D memory. Therefore, the refinement
stage not only improves the current video but also prevents residual
appearance artifacts from accumulating and propagating to subsequent
camera trajectories.

\section{Training and Inference Details}
\label{sec:supp_training_data}
\subsection{Training Data Construction}
\label{sec:training data construction}

We construct a unified training corpus containing 136K samples the from
MultiCamVideo dataset and 100K samples from the OpenVid dataset. Following
TrajectoryCrafter~\cite{yu2025trajectorycrafter}, we employ double
reprojection to synthesize target-aligned partial RGB-D conditions
from monocular videos. We use DepthCrafter~\cite{hu2025depthcrafter}
to estimate temporally consistent depth sequences for the selected
training clips.

\paragraph{Double reprojection.}
Let
$\mathbf{V}_{\mathrm{gt}}
=\{\mathbf{I}_{\mathrm{gt},t}\}_{t=1}^{T}$
denote a clean target video,
$\mathbf{D}_{\mathrm{gt}}
=\{\mathbf{D}_{\mathrm{gt},t}\}_{t=1}^{T}$
its DepthCrafter depth sequence, and
$\mathbf{C}_{\mathrm{gt}}
=\{\mathbf{C}_{\mathrm{gt},t}\}_{t=1}^{T}$
its camera trajectory. For every timestamp, we first construct the
colored point cloud
\begin{equation}
    \mathcal{P}_{\mathrm{gt},t}
    =
    \operatorname{BackProj}
    \left(
        \mathbf{I}_{\mathrm{gt},t},
        \mathbf{D}_{\mathrm{gt},t},
        \mathbf{C}_{\mathrm{gt},t};
        \mathbf{1}
    \right),
\label{eq:supp_training_initial_point_cloud}
\end{equation}
where $\operatorname{BackProj}$ is the RGB-D back-projection operator
introduced in the main paper.

We then sample a virtual camera trajectory
$\mathbf{C}^{\prime}
=\{\mathbf{C}_{t}^{\prime}\}_{t=1}^{T}$
and render the point-cloud sequence along this trajectory:
\begin{equation}
\left(
    \mathbf{I}_{t}^{\prime},
    \mathbf{D}_{t}^{\prime},
    \mathbf{M}_{t}^{\prime}
\right)
=
\mathcal{R}
\left(
    \mathcal{P}_{\mathrm{gt},t},
    \mathbf{C}_{t}^{\prime}
\right),
\qquad t=1,\ldots,T,
\label{eq:supp_first_reprojection}
\end{equation}
where $\mathcal{R}$ is the point-cloud rendering operator defined in
Sec.~\ref{sec:supp_point_cloud_warping}. This first reprojection
introduces view-dependent disocclusions, out-of-view content, and
sparse rendering holes.

We subsequently reconstruct a point cloud from each intermediate
RGB-D frame:
\begin{equation}
    \mathcal{P}_{t}^{\prime}
    =
    \operatorname{BackProj}
    \left(
        \mathbf{I}_{t}^{\prime},
        \mathbf{D}_{t}^{\prime},
        \mathbf{C}_{t}^{\prime};
        \mathbf{M}_{t}^{\prime}
    \right),
\label{eq:supp_intermediate_point_cloud}
\end{equation}
and render it back to the original target camera:
\begin{equation}
\left(
    \mathbf{I}_{\mathrm{par},t},
    \mathbf{D}_{\mathrm{par},t},
    \mathbf{M}_{\mathrm{par},t}
\right)
=
\mathcal{R}
\left(
    \mathcal{P}_{t}^{\prime},
    \mathbf{C}_{\mathrm{gt},t}
\right).
\label{eq:supp_second_reprojection}
\end{equation}
Collecting the resulting frames gives
\begin{equation}
\begin{aligned}
    \mathbf{V}_{\mathrm{par}}
    &=
    \left\{
        \mathbf{I}_{\mathrm{par},t}
    \right\}_{t=1}^{T},\\
    \mathbf{D}_{\mathrm{par}}
    &=
    \left\{
        \mathbf{D}_{\mathrm{par},t}
    \right\}_{t=1}^{T},\\
    \mathbf{M}_{\mathrm{par}}
    &=
    \left\{
        \mathbf{M}_{\mathrm{par},t}
    \right\}_{t=1}^{T}.
\end{aligned}
\label{eq:supp_partial_training_conditions}
\end{equation}

Although the second reprojection returns to the original camera
trajectory, points removed by occlusion or field-of-view changes
during the first reprojection cannot be recovered. Consequently,
$\mathbf{V}_{\mathrm{par}}$ and
$\mathbf{D}_{\mathrm{par}}$ contain realistic geometry-induced
missing regions while remaining spatially aligned with
$\mathbf{V}_{\mathrm{gt}}$ and
$\mathbf{D}_{\mathrm{gt}}$.

\paragraph{Appearance video diffusion.}
For training the cache-guided appearance model
$\mathcal{F}_{\mathrm{app}}$, we use the double-reprojected RGB-D
conditions and visibility mask as inputs, with the clean video as
supervision:
\begin{equation}
    \mathcal{F}_{\mathrm{app}}
    \left(
        \mathbf{V}_{\mathrm{par}},
        \mathbf{D}_{\mathrm{par}},
        \mathbf{M}_{\mathrm{par}}
    \right)
    \longrightarrow
    \mathbf{V}_{\mathrm{gt}}.
\label{eq:supp_appearance_training_pair}
\end{equation}
Here,
$(\mathbf{V}_{\mathrm{par}},
\mathbf{D}_{\mathrm{par}},
\mathbf{M}_{\mathrm{par}})$
plays the same role as
$(\overline{\mathbf{V}}^{i},
\overline{\mathbf{D}}^{i},
\overline{\mathbf{M}}^{i})$
during inference.

\paragraph{Video depth completion.}
To ensure exact consistency between the partial-depth input and its
supervision in known regions, we construct the partial depth by
masking the complete DepthCrafter depth using the double-reprojection
visibility mask:
\begin{equation}
    \mathbf{D}_{\mathrm{mask}}
    =
    \mathbf{M}_{\mathrm{par}}
    \odot
    \mathbf{D}_{\mathrm{gt}}.
\label{eq:supp_masked_training_depth}
\end{equation}
We then convert the masked and complete depths into normalized
disparities:
\begin{equation}
\begin{aligned}
    \mathbf{U}_{\mathrm{par}}
    &=
    \mathcal{N}
    \left(
        \frac{1}{
            \mathbf{D}_{\mathrm{mask}}+\epsilon
        };
        \mathbf{M}_{\mathrm{par}}
    \right),\\
    \mathbf{U}_{\mathrm{gt}}
    &=
    \mathcal{N}
    \left(
        \frac{1}{
            \mathbf{D}_{\mathrm{gt}}+\epsilon
        };
        \mathbf{1}
    \right),
\end{aligned}
\label{eq:supp_training_disparity}
\end{equation}
where $\mathcal{N}(\cdot;\mathbf M)$ denotes min--max normalization
over pixels selected by $\mathbf M$. The depth-completion model is
trained using
\begin{equation}
    \mathcal{F}_{\mathrm{dep}}
    \left(
        \mathbf{V}_{\mathrm{gt}},
        \mathbf{U}_{\mathrm{par}}
    \right)
    \longrightarrow
    \mathbf{U}_{\mathrm{gt}}.
\label{eq:supp_depth_training_pair}
\end{equation}
During training, $\mathbf{V}_{\mathrm{gt}}$ plays the role of the
generated RGB video $\mathbf{V}^{i}$ in
Eq.~\ref{eq:depth_completion}, while
$\mathbf{U}_{\mathrm{par}}$ corresponds to the partial disparity
$\overline{\mathbf{U}}^{i}$.

\paragraph{Geometry-aware refinement.}
For training the refinement model, we use the partial RGB video and
complete DepthCrafter depth as conditions, with the clean target video
as supervision:
\begin{equation}
    \mathcal{F}_{\mathrm{ref}}
    \left(
        \mathbf{V}_{\mathrm{par}},
        \mathbf{D}_{\mathrm{gt}}
    \right)
    \longrightarrow
    \mathbf{V}_{\mathrm{gt}}.
\label{eq:supp_refinement_training_pair}
\end{equation}
Here, $\mathbf{V}_{\mathrm{par}}$ corresponds to the partial RGB
condition $\overline{\mathbf{V}}^{i}$, while
$\mathbf{D}_{\mathrm{gt}}$ plays the role of the completed dense depth
$\mathbf{D}^{i}$ used during inference. This construction teaches the
refinement model to restore high-frequency appearance details while
remaining constrained by complete scene geometry.

\subsection{Training Configurations}
\label{sec:supp_training_config}

We train the cache-guided appearance generation model
$\mathcal{F}_{\mathrm{app}}$, cache-consistent video depth completion
model $\mathcal{F}_{\mathrm{dep}}$, and geometry-aware refinement
model $\mathcal{F}_{\mathrm{ref}}$ independently. All three models are
initialized from the pretrained Wan2.1-Fun-1.3B-InP checkpoint and
adopt the same training configuration. 

For each model, we optimize all parameters of the video DiT while
keeping the Wan VAE, UMT5 text encoder, and CLIP image encoder frozen.
We use AdamW with $\beta_{1}=0.9$, $\beta_{2}=0.999$, a learning rate
of $1\times10^{-4}$, and a weight decay of $1\times10^{-2}$. The
learning rate follows a cosine annealing schedule without warm-up and
decreases to $1\%$ of its initial value, i.e.,
$1\times10^{-6}$, over the complete training process.

All models are trained for five epochs using bfloat16 precision. Each
training sample contains 81 frames at a resolution of
$480\times832$. We use a per-GPU batch size of one on eight NVIDIA H20
GPUs, resulting in a global batch size of eight without additional
gradient accumulation. Each training sample is used once per epoch.
We enable distributed data parallelism with unused-parameter detection
and save the trainable video-DiT parameters every 1,000 optimization
steps. The module-specific inputs and supervision targets follow the
training-data construction described in
Sec.~\ref{sec:supp_training_data}.

\subsection{Inference Configurations}
\label{sec:supp_inference_config}

At inference time, Demiurge sequentially processes the prescribed
target camera trajectories. For each trajectory, we first render the
current 4D memory using dual-stream warping and remove unreliable
pixels using the noise monitor. The cleaned partial RGB, depth, and
visibility conditions are then passed to
$\mathcal{F}_{\mathrm{app}}$ to generate the initial novel-view RGB
video $\mathbf{V}^{i}$. Subsequently,
$\mathcal{F}_{\mathrm{dep}}$ predicts a dense relative disparity,
which is converted to metric depth through the two-stage global
alignment and Poisson refinement. Finally,
$\mathcal{F}_{\mathrm{ref}}$ uses the partial RGB rendering and
completed dense depth to produce the refined video
$\mathbf{V}_{\mathrm{ref}}^{i}$. The resulting RGB-D pair
$(\mathbf{V}_{\mathrm{ref}}^{i},\mathbf{D}^{i})$ is back-projected
and accumulated into the 4D memory before processing the next target
trajectory.

All three video diffusion modules generate 81-frame videos at a
resolution of $480\times832$. We use 50 flow-matching inference steps. All diffusion modules operate in
bfloat16 precision. We enable tiled Wan VAE encoding and decoding with
a tile size of $(30,52)$ and a tile stride of $(15,26)$ to reduce peak
memory consumption.

To control peak memory during multi-source rendering, we process at
most 16 consecutive frames and eight historical trajectories in each
rendering batch. This chunking only changes the execution order and
does not affect the z-buffer composition result. Rendered metric depth
is encoded as bounded inverse depth using
$\phi(d)=1/(d+1)$ before being used as a diffusion condition. The
geometric depth is recovered when the observation is subsequently
used for point-cloud construction or warping.

\section{Evaluation Details}
\label{sec:supp_evaluation}

\subsection{Metric Definitions}
\label{sec:supp_metrics}

All quantitative metrics are computed over the complete evaluation
set unless otherwise specified. The benchmark contains 128 input
videos and 14 target camera trajectories per input, resulting in
1,792 generated novel-view videos.

\paragraph{Fr\'echet Video Distance.}
We use Fr\'echet Video Distance (FVD) to evaluate the
distribution-level quality of generated videos. For every target
trajectory, we compare the 128 generated videos with the 128 input
videos. We use the pretrained I3D network as the feature extractor.
All 81 frames of each video are resized to $224\times224$, normalized
to $[-1,1]$, and jointly passed to I3D to obtain a video-level feature
vector.

Let $(\boldsymbol{\mu}_{g},\boldsymbol{\Sigma}_{g})$ and
$(\boldsymbol{\mu}_{r},\boldsymbol{\Sigma}_{r})$ denote the mean and
covariance of the generated and reference video features,
respectively. FVD is computed as
\begin{equation}
\begin{aligned}
    \operatorname{FVD}
    =
    \left\|
        \boldsymbol{\mu}_{g}
        -
        \boldsymbol{\mu}_{r}
    \right\|_{2}^{2}
    +
    \operatorname{Tr}
    \left(
        \boldsymbol{\Sigma}_{g}
        +
        \boldsymbol{\Sigma}_{r}
        -
        2
        \left(
            \boldsymbol{\Sigma}_{g}
            \boldsymbol{\Sigma}_{r}
        \right)^{1/2}
    \right).
\end{aligned}
\label{eq:supp_fvd}
\end{equation}
We compute FVD independently for each target trajectory and report
the average over all trajectories.

\paragraph{Q-Align.}
We use Q-Align~\cite{wu2023q} to assess the perceptual quality
and aesthetics of the generated results at both the video and image
levels. Specifically, we report four scores:
Q-Align-Video Quality, Q-Align-Video Aesthetics,
Q-Align-Image Quality, and Q-Align-Image Aesthetics.

For video-level evaluation, each 81-frame video is divided into
consecutive segments of eight frames to satisfy the model context
limit. Q-Align evaluates each segment independently, and the segment
scores are averaged to obtain the video-level score. If the final
segment contains only one frame, its last frame is repeated to form a
two-frame segment. For image-level evaluation, all 81 frames are
independently scored and then averaged. The reported values are
further averaged over all scenes and target trajectories.

\paragraph{Projection MSE.}
Projection MSE (P-MSE) evaluates the consistency between a generated
novel-view video and the content observable from the input video under
the same target camera trajectory. For each input video and target
trajectory, we first geometrically project the input RGB-D video onto
the target trajectory, obtaining a partial RGB rendering and its
visibility mask.

Let $\mathbf{Y}_{s,n,t}$ denote frame $t$ of the video generated for
scene $s$ and target trajectory $n$,
$\mathbf{R}_{s,n,t}$ the corresponding partial RGB projection of the
input video, and $\mathbf{M}_{s,n,t}$ its binary visibility mask.
All RGB values are normalized to $[0,1]$. The P-MSE of an individual
input--trajectory pair is
\begin{equation}
\operatorname{P\text{-}MSE}_{s,n}
=
\frac{
    \displaystyle
    \sum_{t}
    \sum_{\mathbf{p}}
    \mathbf{M}_{s,n,t}(\mathbf{p})
    \left\|
        \mathbf{Y}_{s,n,t}(\mathbf{p})
        -
        \mathbf{R}_{s,n,t}(\mathbf{p})
    \right\|_{2}^{2}
}{
    \displaystyle
    3
    \sum_{t}
    \sum_{\mathbf{p}}
    \mathbf{M}_{s,n,t}(\mathbf{p})
}.
\label{eq:supp_pmse_pair}
\end{equation}
The factor of three averages the squared error over the RGB channels.
The visibility mask is binarized at a grayscale threshold of
$127/255$, and P-MSE is evaluated only over its visible regions.

We use pair-macro aggregation as the main metric, assigning equal
weight to every input--trajectory pair:
\begin{equation}
    \operatorname{P\text{-}MSE}
    =
    \frac{1}{SN}
    \sum_{s=1}^{S}
    \sum_{n=1}^{N}
    \operatorname{P\text{-}MSE}_{s,n},
\label{eq:supp_pmse}
\end{equation}
where $S=128$ and $N=14$. This differs from pooling all visible
pixels globally, which would assign larger weights to pairs with
greater projected coverage.

\paragraph{Camera trajectory estimation.}
We use VGGT~\cite{wang2025vggt} to recover camera poses from the
generated videos. For each scene and sampled timestamp, the frames
generated along all 14 target trajectories are jointly processed to
estimate their camera-to-world matrices. We sample every tenth frame
from each 81-frame video. The estimated poses are compared against the
target camera trajectories used during generation.

The target trajectories progressively interpolate from the source
camera to their prescribed endpoints. Consequently, all target views
share the same camera pose at the first frame. Since this timestamp
contains no cross-view camera variation, we exclude it from camera
accuracy evaluation.

\paragraph{Rotation error.}
Let $\widehat{\mathbf{R}}_{t,n}$ and
$\mathbf{R}_{t,n}$ denote the estimated and target camera rotations
at sampled timestamp $t$ and trajectory $n$, respectively. VGGT and
our trajectory generator use different camera-axis conventions and
independent world-coordinate gauges. We therefore first determine a
global axis-convention rotation $\mathbf{D}$ from the 24
orientation-preserving signed axis permutations. For every sampled
timestamp, we additionally estimate a world-frame alignment
$\mathbf{W}_{t}$ through orthogonal Procrustes alignment.

The rotation error is defined as the mean geodesic distance:
\begin{equation}
\begin{aligned}
    \operatorname{R\text{-}err}
    =
    \frac{1}{|\mathcal{T}|N}
    \sum_{t\in\mathcal{T}}
    \sum_{n=1}^{N}
    \arccos
    \left(
        \frac{
            \operatorname{Tr}
            \left[
                \widehat{\mathbf{R}}_{t,n}
                \left(
                    \mathbf{W}_{t}
                    \mathbf{R}_{t,n}
                    \mathbf{D}
                \right)^{\mathsf T}
            \right]
            -1
        }{2}
    \right),
\end{aligned}
\label{eq:supp_rotation_error}
\end{equation}
where $\mathcal{T}$ is the set of non-degenerate sampled timestamps.
We report R-err in radians. The global convention alignment
$\mathbf{D}$ is selected by exhaustive search to minimize the mean
rotation error, while each $\mathbf{W}_{t}$ is solved in closed form.

\paragraph{Translation error.}
Let $\widehat{\mathbf{c}}_{t,n}$ and
$\mathbf{c}_{t,n}$ denote the camera centers extracted from the
estimated and target camera-to-world matrices, respectively. Since
monocular camera reconstruction is ambiguous up to scale, rotation,
and translation, we independently align the estimated camera centers
to the target centers at each timestamp using Umeyama Sim(3)
alignment:
\begin{equation}
\left(
    s_{t},
    \mathbf{R}_{t}^{\mathrm{sim}},
    \mathbf{b}_{t}
\right)
=
\underset{s,\mathbf{R},\mathbf{b}}{\arg\min}
\sum_{n=1}^{N}
\left\|
    \mathbf{c}_{t,n}
    -
    \left(
        s\mathbf{R}
        \widehat{\mathbf{c}}_{t,n}
        +
        \mathbf{b}
    \right)
\right\|_{2}^{2}.
\label{eq:supp_sim3_alignment}
\end{equation}
Following the implementation, translation error is computed by
summing the squared errors over trajectories and averaging over
timestamps:
\begin{equation}
\begin{aligned}
    \operatorname{T\text{-}err}
    =
    \frac{1}{|\mathcal{T}|}
    \sum_{t\in\mathcal{T}}
    \sum_{n=1}^{N}
    \left\|
        \mathbf{c}_{t,n}
        -
        \left(
            s_{t}
            \mathbf{R}_{t}^{\mathrm{sim}}
            \widehat{\mathbf{c}}_{t,n}
            +
            \mathbf{b}_{t}
        \right)
    \right\|_{2}^{2}.
\end{aligned}
\label{eq:supp_translation_error}
\end{equation}
Both R-err and T-err are first computed per scene and then averaged
over all 128 evaluation scenes.

\paragraph{View Sync.}
View Sync evaluates the visual consistency among the novel-view videos
generated from the same input sequence. For each scene, we select the
temporally aligned final frame from each of the 14 generated videos
and compute dense pairwise correspondences using
RoMa~\cite{edstedt2024roma}. This results in
$\binom{14}{2}=91$ view pairs per scene.

Before matching, all frames are resized to $480\times832$. For
baselines that concatenate the generated result with an auxiliary
rendering, we retain only the generated region before resizing. RoMa
uses a coarse resolution of 560 and produces dense correspondences on
an upsampled grid of $864\times1152$.

Let $\gamma_{s,n,m}(\mathbf{p})$ denote the confidence assigned by
RoMa to the correspondence at pixel $\mathbf{p}$ between trajectories
$n$ and $m$ of scene $s$. We count the high-confidence
correspondences as
\begin{equation}
    \operatorname{VS}_{s,n,m}
    =
    \sum_{\mathbf{p}\in\Omega_{\mathrm{RoMa}}}
    \mathbb{I}
    \left[
        \gamma_{s,n,m}(\mathbf{p})
        >
        \tau_{\mathrm{sync}}
    \right],
\label{eq:supp_view_sync_pair}
\end{equation}
where $\Omega_{\mathrm{RoMa}}$ denotes the dense output grid and
$\tau_{\mathrm{sync}}=0.7$. Since RoMa performs symmetric matching,
we retain only one correspondence direction to avoid duplicate
counting.

The final View Sync score is the average number of high-confidence
correspondences over all scenes and trajectory pairs:
\begin{equation}
    \operatorname{ViewSync}
    =
    \frac{1}{
        S\binom{N}{2}
    }
    \sum_{s=1}^{S}
    \sum_{1\leq n<m\leq N}
    \operatorname{VS}_{s,n,m},
\label{eq:supp_view_sync}
\end{equation}
where $S=128$ and $N=14$. A higher View Sync score indicates stronger
cross-view consistency among videos generated along different camera
trajectories. The optional fundamental-matrix verification is used
only for diagnostic analysis and is not included in the reported
View Sync score.

\paragraph{3DGS Reconstruction Consistency.}
We evaluate whether the multi-trajectory videos generated from the same input can be jointly explained by a coherent 3D representation. Specifically, we map all generated videos onto a common 81-frame reference timeline and select three normalized time points, $\mathcal{T}=\{1/3,2/3,1\}$. At each time point, we extract the temporally aligned frames from the 14 generated videos and independently reconstruct a 3D Gaussian representation using AnySplat~\cite{jiang2025anysplat}. Different time points are reconstructed separately to avoid mixing dynamic scene content into a single static representation.

The reconstructed Gaussians are rendered back to the 14 input viewpoints using the camera poses and intrinsics estimated by AnySplat. We then compare each rendered image with its corresponding generated frame using PSNR, SSIM, and LPIPS. For a scene $s$ and time point $p$, the reconstruction score is
\begin{equation}
    \mathcal{E}_{s,p}^{m}
    =
    \frac{1}{N}
    \sum_{i=1}^{N}
    m\left(
        \hat{\mathbf I}_{s,p}^{i},
        \mathbf I_{s,p}^{i}
    \right),
\end{equation}
where $N=14$, $\mathbf I_{s,p}^{i}$ is the frame extracted from the $i$-th generated trajectory, $\hat{\mathbf I}_{s,p}^{i}$ is its 3DGS rendering, and $m$ denotes PSNR, SSIM, or LPIPS. The final result is obtained by first averaging over the three time points within each scene and then averaging equally over all 128 scenes:
\begin{equation}
    \mathcal{E}^{m}
    =
    \frac{1}{S}
    \sum_{s=1}^{S}
    \frac{1}{|\mathcal T|}
    \sum_{p\in\mathcal T}
    \mathcal{E}_{s,p}^{m},
\end{equation}
where $S=128$. Higher PSNR and SSIM and lower LPIPS indicate that the generated views can be more faithfully represented by a shared 3D structure, reflecting stronger cross-view appearance and geometric consistency.

\section{Key capabilities}
\label{appendix: Key capabilities}

We next qualitatively examine three key capabilities of Demiurge that underpin navigable 4D world generation: coherent spatiotemporal lattice construction, large-parallax video rerendering, and geometry-faithful 3D/4D reconstruction.

\subsection{Coherent spatiotemporal lattice generation}
\label{Coherent spatiotemporal lattice generation}

\begin{figure*}[ht]
    \centering
    \includegraphics[width=0.85\linewidth]{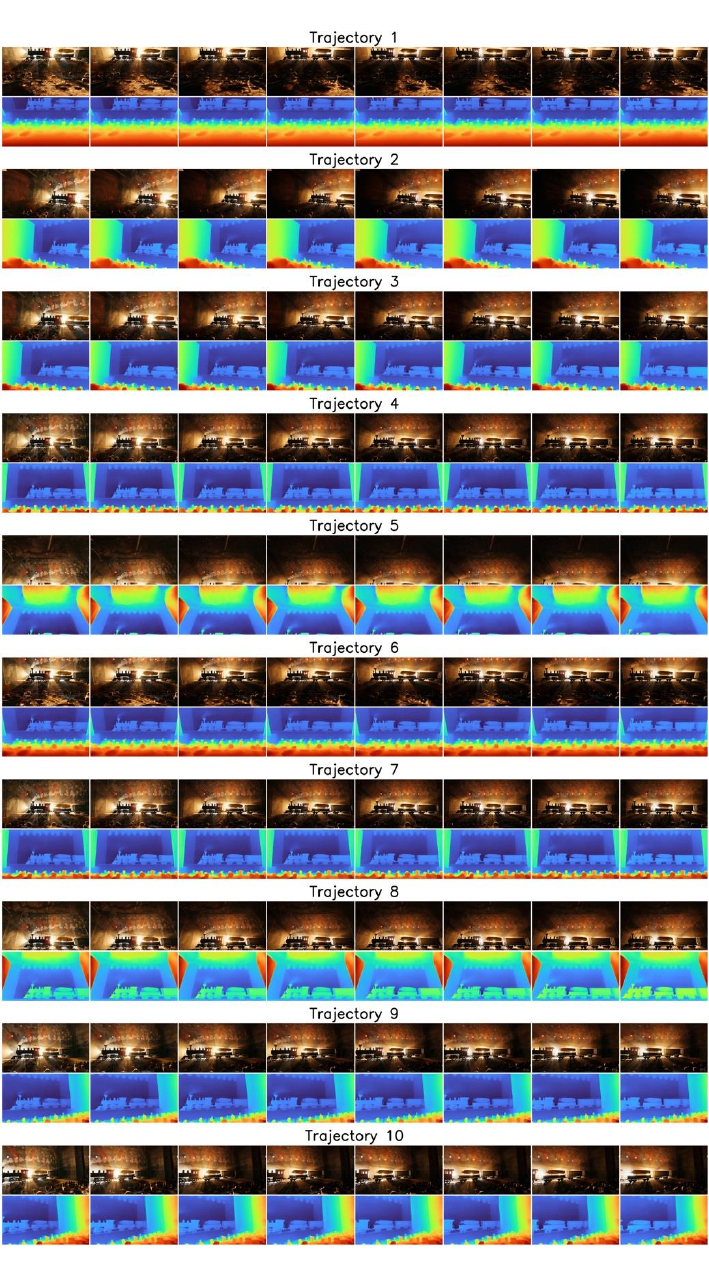}
    \caption{\textbf{Coherent spatiotemporal lattice generation (Case 1).} Rows correspond to different camera trajectories and columns to temporally aligned frames; each sample shows the generated RGB view and its recovered depth.}
    \label{fig:ablation_Key_capabilities_fig1}
\end{figure*}

\begin{figure*}[ht]
    \centering
    \includegraphics[width=0.85\linewidth]{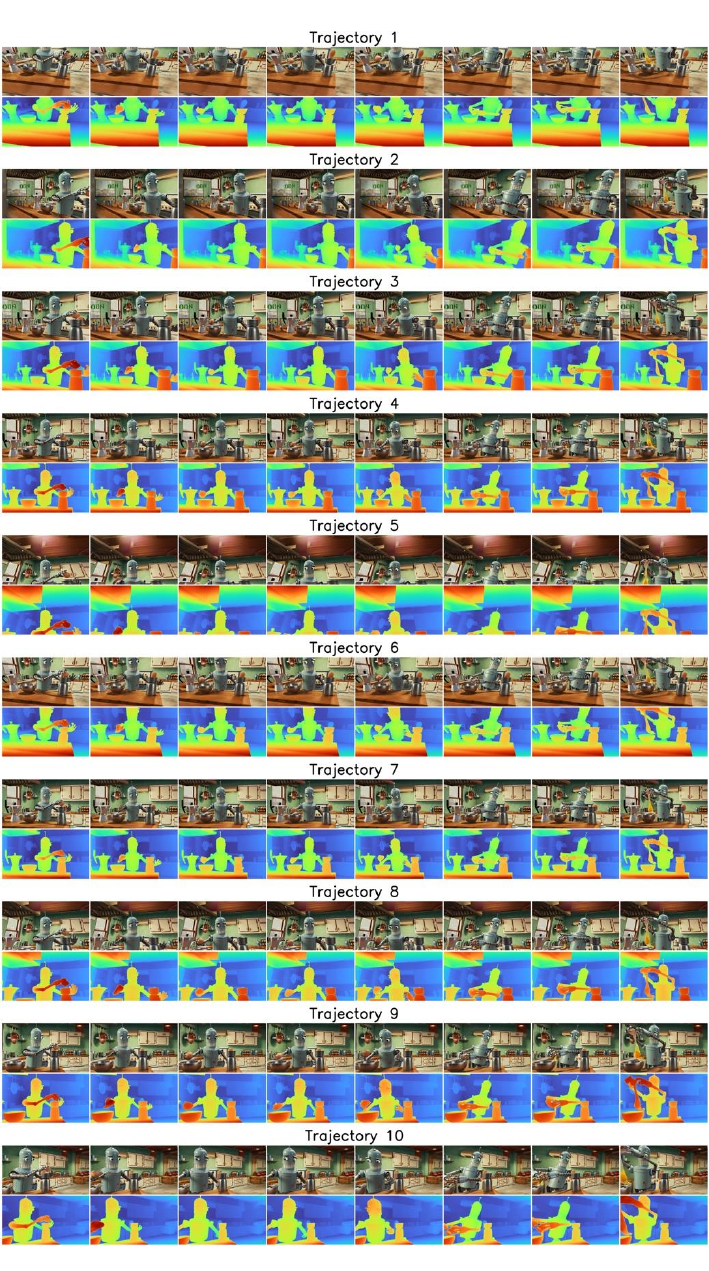}
    \caption{\textbf{Coherent spatiotemporal lattice generation (Case 2).} Rows correspond to different camera trajectories and columns to temporally aligned frames; each sample shows the generated RGB view and its recovered depth.}
    \label{fig:ablation_Key_capabilities_fig2}
\end{figure*}

Figs.~\ref{fig:ablation_Key_capabilities_fig1} and
\ref{fig:ablation_Key_capabilities_fig2} illustrate the spatiotemporal lattices generated along ten camera trajectories. Each row corresponds to a target
trajectory, while the columns show temporally aligned frames, jointly spanning the view and time dimensions. Demiurge preserves scene appearance, dynamic content, and recovered geometry across this lattice, demonstrating coherent world expansion beyond independently generated camera paths.

\subsection{Large-parallax video rerendering}
\label{Large-parallax video rerendering}

\begin{figure*}[ht]
    \centering
    \includegraphics[width=0.85\linewidth]{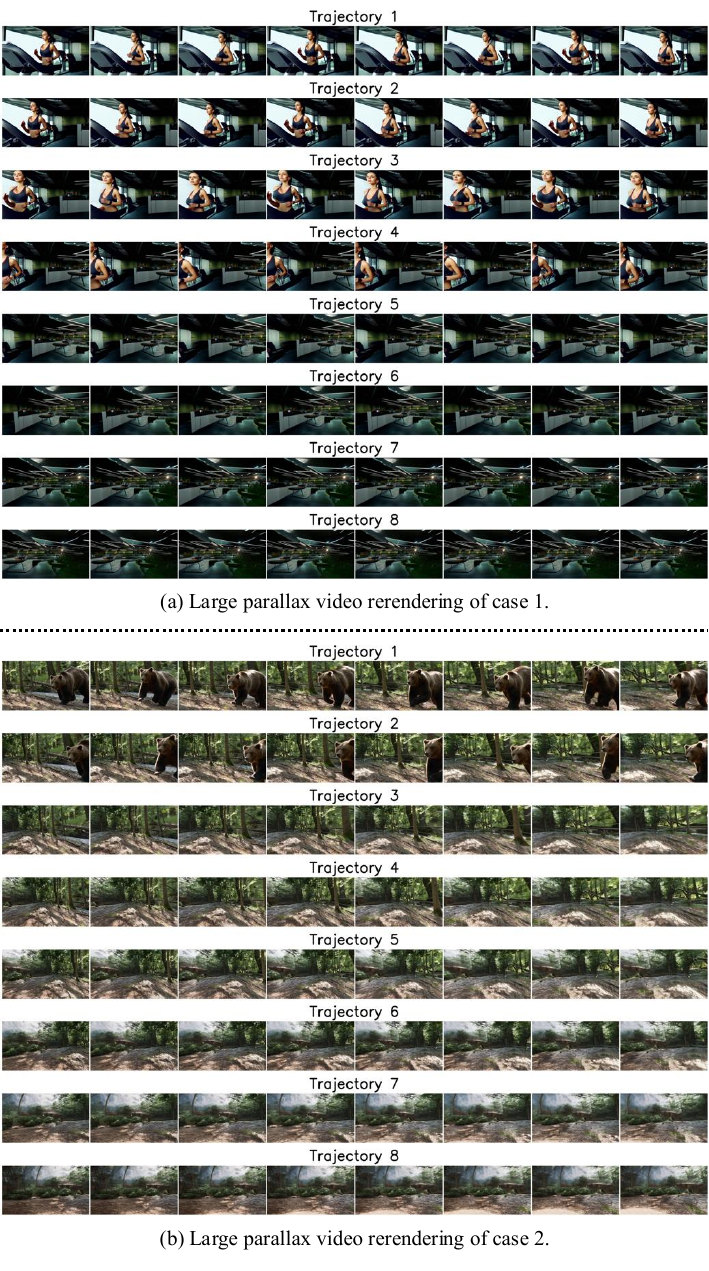}
    \caption{\textbf{Large-parallax video rerendering.} Demiurge progressively rotates the camera by up to $180^\circ$ rightward \textbf{(a)} and leftward \textbf{(b)}, while maintaining consistent scene content across eight successive trajectories. Sampled frames in each row are ordered chronologically.}
    \label{fig:ablation_Key_capabilities_fig3}
\end{figure*}

As shown in Fig.~\ref{fig:ablation_Key_capabilities_fig3}, Demiurge supports progressive camera rotations of up to $180^\circ$ in both directions. Despite the diminishing overlap with the input view, it consistently expands newly
revealed regions while preserving scene identity, dynamics, and spatial continuity across successive trajectories.

\subsection{Geometry-faithful outputs for 3DGS reconstruction}
\label{Geometry-faithful outputs for 3DGS reconstruction}

\begin{figure*}[ht]
    \centering
    \includegraphics[width=0.80\linewidth]{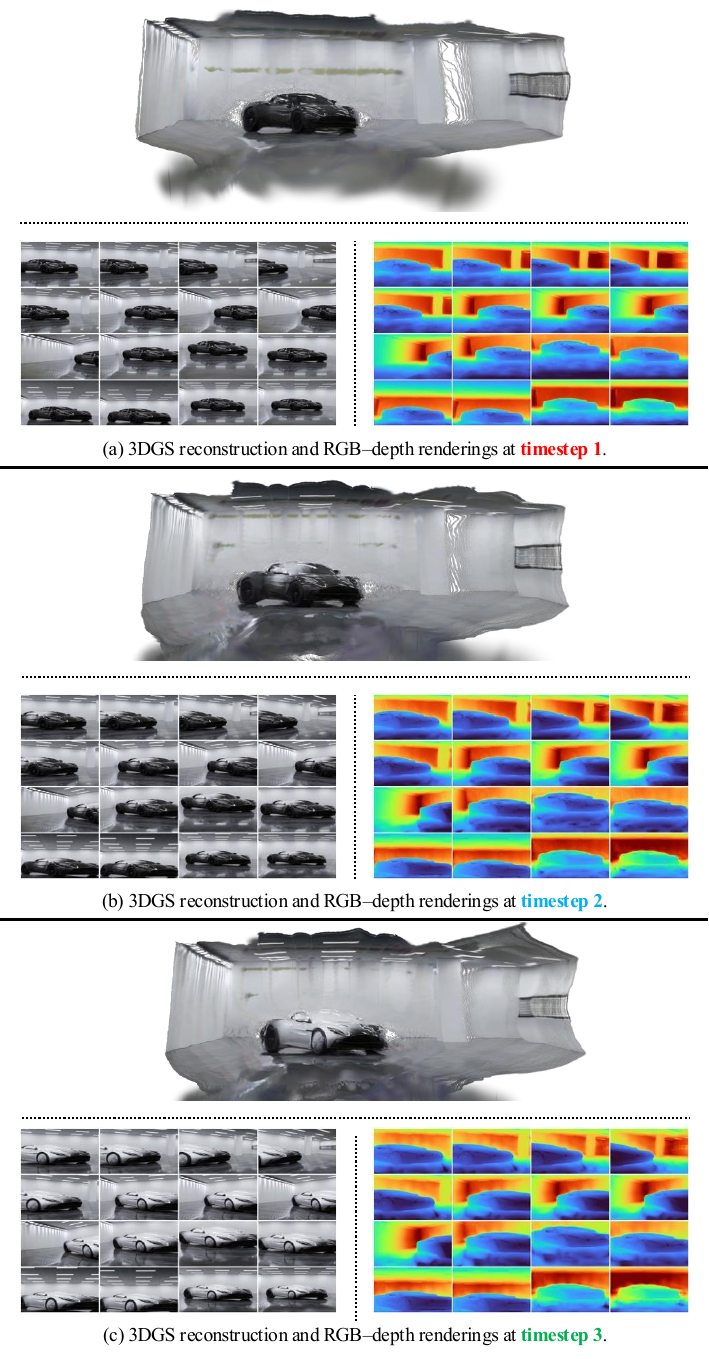}
    \caption{\textbf{Geometry-faithful 3DGS reconstruction (Case 1).} At each time step, we show the reconstructed 3DGS (top) and its multi-view RGB (bottom left) and depth renderings (bottom right).}

    \label{fig:ablation_Key_capabilities_fig4}
\end{figure*}

\begin{figure*}[ht]
    \centering
    \includegraphics[width=0.80\linewidth]{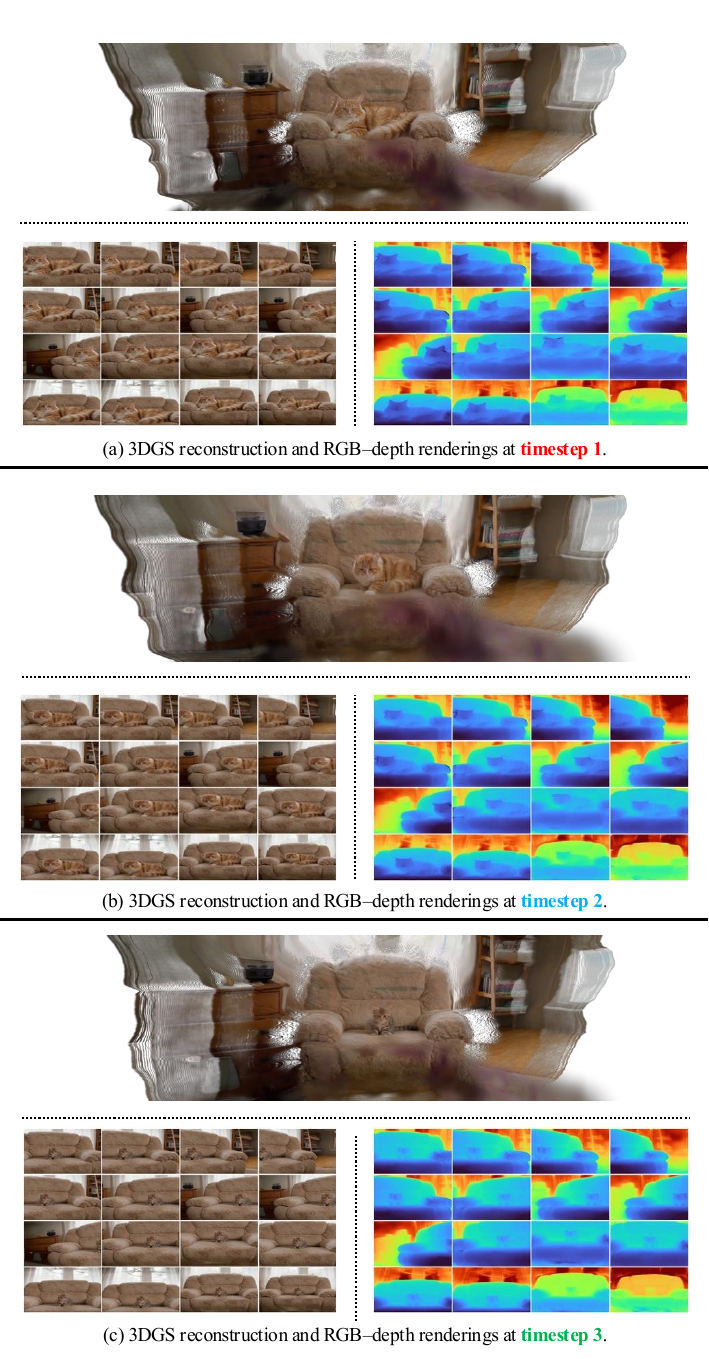}
    \caption{\textbf{Geometry-faithful 3DGS reconstruction (Case 2).} At each time step, we show the reconstructed 3DGS (top) and its multi-view RGB (bottom left) and depth renderings (bottom right).}
    \label{fig:ablation_Key_capabilities_fig5}
\end{figure*}

As shown in Figs.~\ref{fig:ablation_Key_capabilities_fig4} and
\ref{fig:ablation_Key_capabilities_fig5}, we independently reconstruct a 3DGS from the generated multi-view observations at three time steps using AnySplat~\cite{jiang2025anysplat}. The resulting representations preserve coherent scene geometry and produce consistent RGB and depth renderings across views, demonstrating that Demiurge generates geometrically compatible observations rather than merely plausible individual videos.

\section{More qualitative comparisons}
\label{appendix: more qualitative comparisons}

\begin{figure*}[t]
    \centering
    \includegraphics[width=0.72\linewidth]{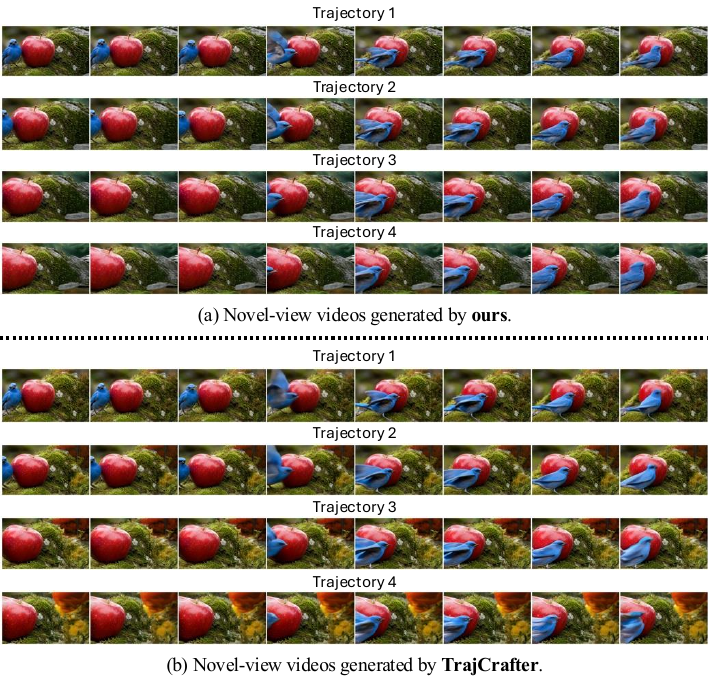}
    \caption{\textbf{Additional comparison with TrajCrafter.} Demiurge better preserves object appearance and scene geometry across different camera trajectories, with less content drift and viewpoint-dependent distortion.}
    \label{fig:ablation_more_comparison_fig1}
\end{figure*}

We provide additional qualitative comparisons with all competing methods in Figs.~\ref{fig:ablation_more_comparison_fig1}--%
\ref{fig:ablation_more_comparison_fig6}. In each figure, rows correspond to different target camera trajectories, while sampled frames are arranged chronologically from left to right. These examples cover diverse indoor and outdoor scenes, dynamic foreground objects, and substantial viewpoint changes.

Although the competing methods generally produce plausible videos along an individual trajectory, their independently generated views often exhibit appearance drift, object deformation, inconsistent background structure, or insufficient response to the prescribed camera motion. Such discrepancies become increasingly evident when comparing the same object or region across different trajectories. In contrast, Demiurge consistently preserves object identity, fine-grained appearance, scene layout, and temporal dynamics while revealing geometrically plausible content under novel viewpoints. These results further demonstrate the benefit of explicitly consolidating and reusing historical appearance and geometry through the evolving 4D memory.

\begin{figure*}[t]
    \centering
    \includegraphics[width=0.72\linewidth]{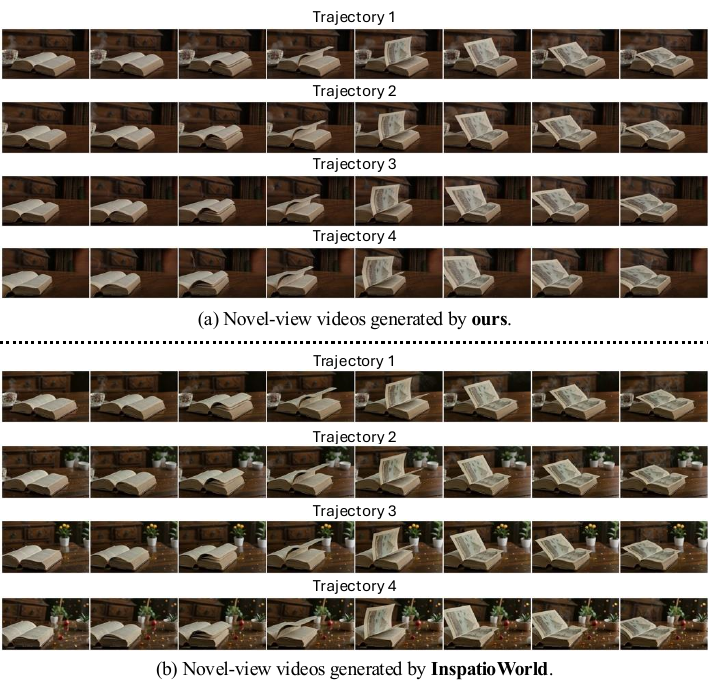}
    \caption{\textbf{Additional comparison with InSpatioWorld.} Demiurge maintains more consistent foreground details and background layout across trajectories, whereas InSpatioWorld introduces noticeable content changes.}
    \label{fig:ablation_more_comparison_fig2}
\end{figure*}

\begin{figure*}[t]
    \centering
    \includegraphics[width=0.72\linewidth]{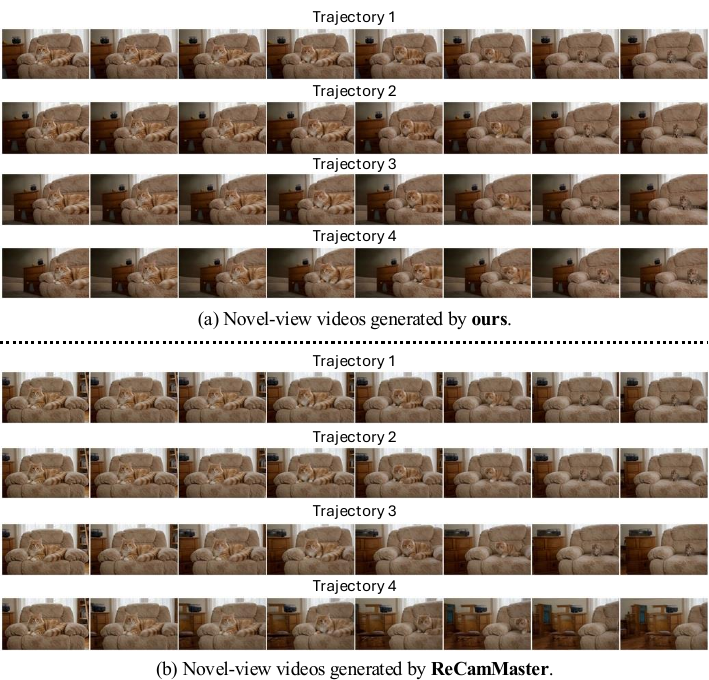}
    \caption{\textbf{Additional comparison with ReCamMaster.} Demiurge preserves object identity and spatial structure under changing viewpoints, while ReCamMaster exhibits cross-trajectory appearance and geometry inconsistencies.}
    \label{fig:ablation_more_comparison_fig3}
\end{figure*}

\begin{figure*}[t]
    \centering
    \includegraphics[width=0.72\linewidth]{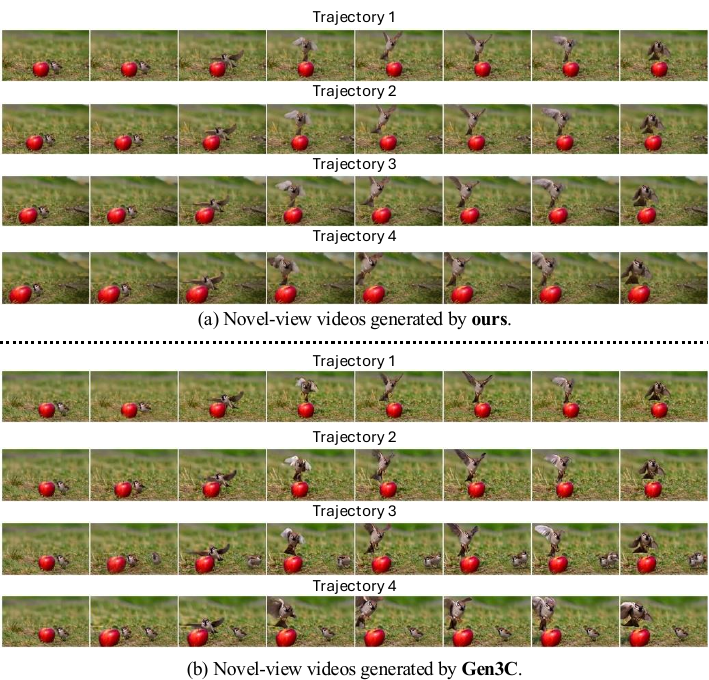}
    \caption{\textbf{Additional comparison with GEN3C.} Demiurge produces more consistent object appearance, dynamics, and spatial relationships across multiple target trajectories.}
    \label{fig:ablation_more_comparison_fig4}
\end{figure*}

\begin{figure*}[t]
    \centering
    \includegraphics[width=0.72\linewidth]{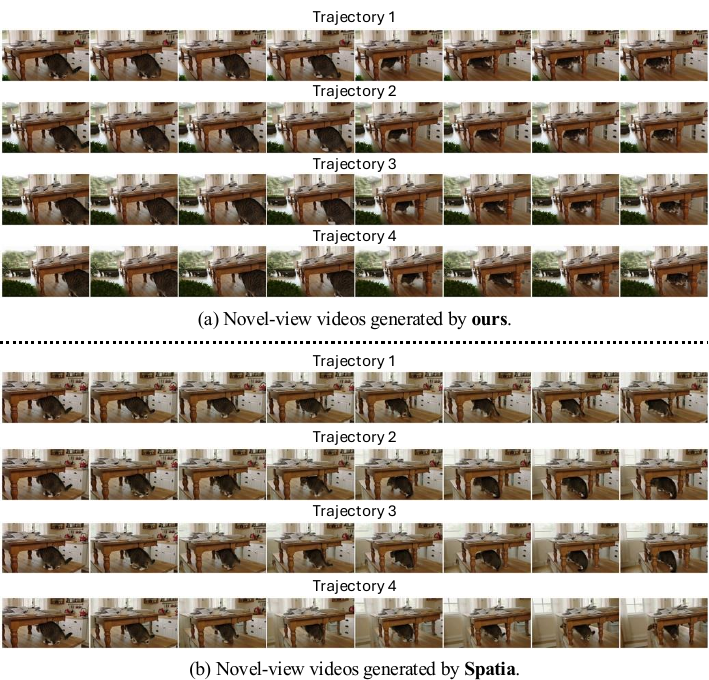}
    \caption{\textbf{Additional comparison with Spatia.} Demiurge follows the prescribed camera trajectories while maintaining coherent scene content, whereas Spatia exhibits weaker viewpoint variation and cross-trajectory consistency.}
    \label{fig:ablation_more_comparison_fig5}
\end{figure*}

\begin{figure*}[t]
    \centering
    \includegraphics[width=0.72\linewidth]{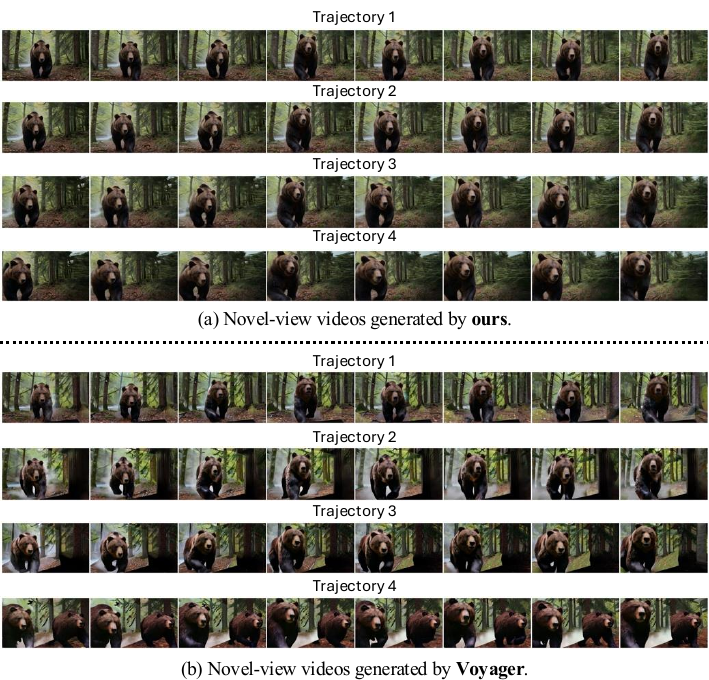}
    \caption{\textbf{Additional comparison with Voyager.} Demiurge better preserves the identity, scale, and geometry of the dynamic foreground object across substantial camera motions.}
    \label{fig:ablation_more_comparison_fig6}
\end{figure*}

\section{Additional ablation results}
\label{appendix:Additional ablation results}

\begin{figure*}[t]
    \centering
    \includegraphics[width=0.9\linewidth]{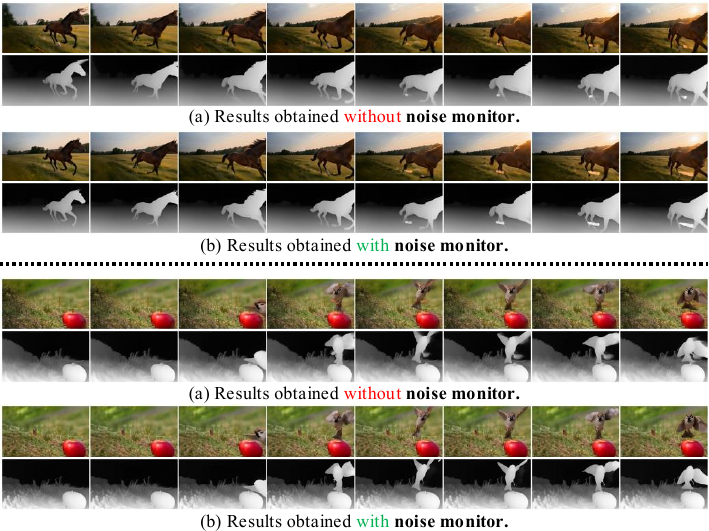}
    \caption{\textbf{Additional ablation of the noise monitor.} The noise monitor filters unreliable cache projections, reducing boundary artifacts and stabilizing the recovered geometry over time.}
    \label{fig:ablation_ablation_fig1}
\end{figure*}

We provide additional qualitative ablations in Figs.~\ref{fig:ablation_ablation_fig1}--%
\ref{fig:ablation_ablation_fig4}. In each example, sampled frames are arranged chronologically from left to right, with the corresponding depth maps shown below.

\begin{figure*}[t]
    \centering
    \includegraphics[width=0.9\linewidth]{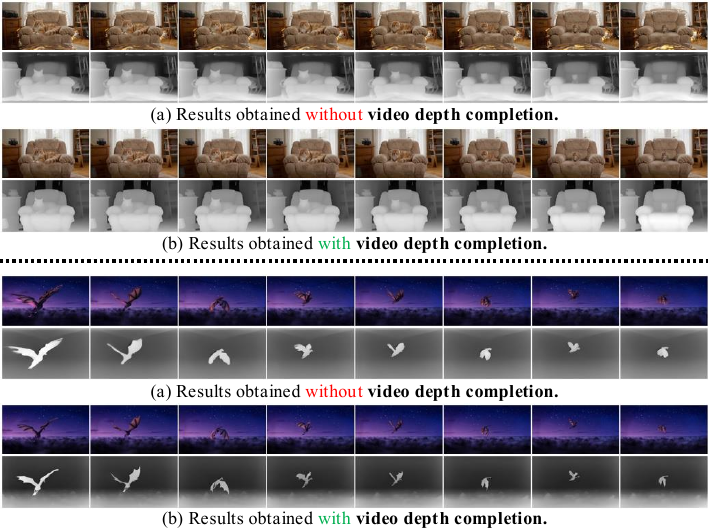}
    \caption{\textbf{Additional ablation of video depth completion.} Cache-consistent depth completion produces more coherent object geometry and temporal structure than direct monocular depth estimation.}
    \label{fig:ablation_ablation_fig2}
\end{figure*}

\begin{figure*}[t]
    \centering
    \includegraphics[width=0.9\linewidth]{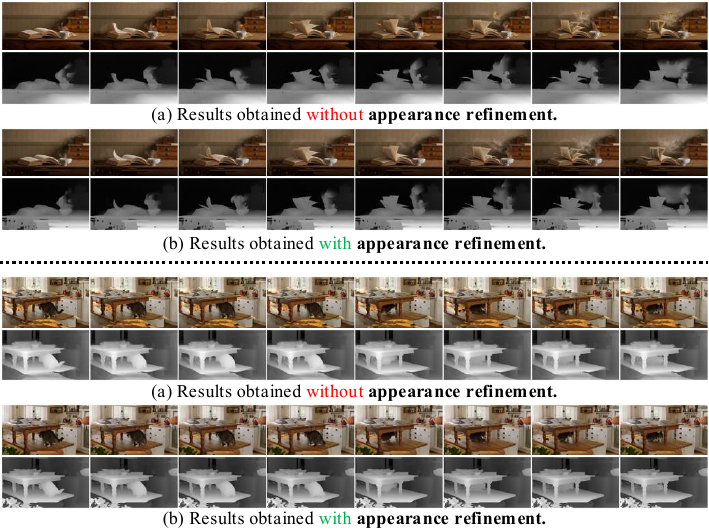}
    \caption{\textbf{Additional ablation of geometry-aware refinement.} Refinement suppresses residual appearance artifacts and texture flicker while preserving the recovered geometry.}
    \label{fig:ablation_ablation_fig3}
\end{figure*}

\begin{figure*}[t]
    \centering
    \includegraphics[width=0.95\linewidth]{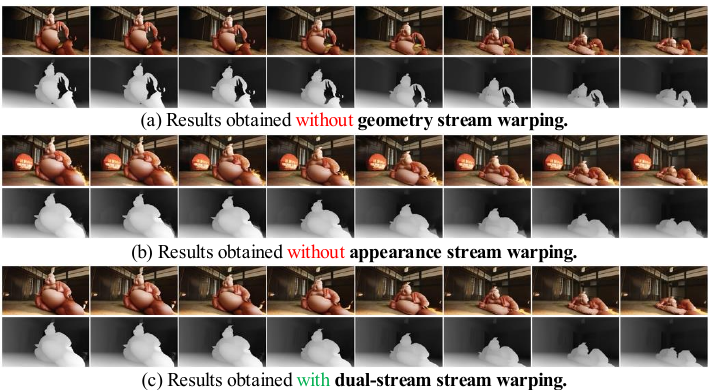}
    \caption{\textbf{Additional ablation of dual-stream warping.} The appearance stream preserves visual details, while the geometry stream provides reliable occlusion reasoning; combining them yields consistent appearance and geometry.}
    \label{fig:ablation_ablation_fig4}
\end{figure*}

Without the noise monitor, unreliable point-cloud projections near depth discontinuities propagate into the generated RGB video and recovered geometry, causing local boundary artifacts. Filtering these projections produces cleaner
appearance and more stable object geometry, as shown in
Fig.~\ref{fig:ablation_ablation_fig1}. Fig.~\ref{fig:ablation_ablation_fig2}
further demonstrates the importance of cache-consistent video depth completion. Direct depth estimation yields temporally unstable shapes and geometric discontinuities, whereas our module anchors observed regions to the cached geometry and consistently completes newly revealed regions. As shown in Fig.~\ref{fig:ablation_ablation_fig3}, geometry-aware refinement suppresses residual appearance artifacts and texture flicker while preserving the recovered scene structure. Finally, Fig.~\ref{fig:ablation_ablation_fig4} verifies the complementary roles of dual-stream warping. Removing the geometry stream weakens occlusion reasoning
and introduces erroneous projections, while removing the appearance stream causes visual degradation from repeatedly updated content. Their combination preserves stable appearance while retaining dense geometry for reliable visibility reasoning.

\section{Compatibility with Existing Method}
\label{appendix: Compatibility with Existing Method}

\begin{figure*}[t]
    \centering
    \includegraphics[width=1\linewidth]{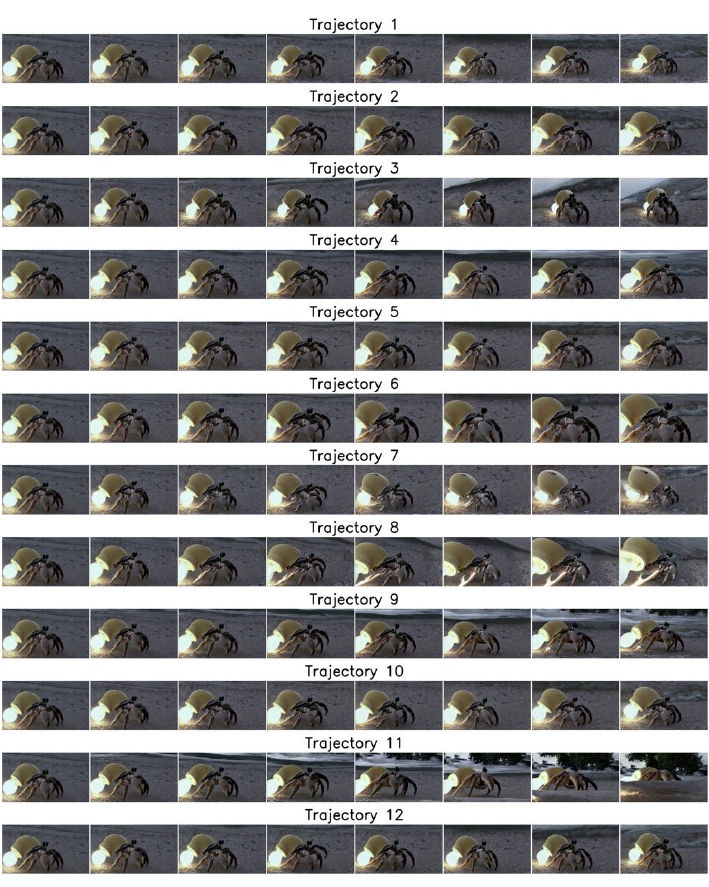}
    \caption{\textbf{Compatibility with GEN3C (Case 1, RGB).} Replacing our appearance generator with GEN3C still enables coherent RGB generation across twelve target trajectories. Frames in each row are arranged chronologically.}
    \label{fig:ablation_comptable_fig1_rgb}
\end{figure*}

To examine the modularity of Demiurge, we replace our cache-guided appearance video diffusion model with the existing GEN3C generator while retaining the remaining depth recovery and memory-update pipeline. As shown in Figs.~\ref{fig:ablation_comptable_fig1_rgb}--%
\ref{fig:ablation_comptable_fig2_depth}, the resulting system generates coherent RGB videos across multiple trajectories and recovers temporally stable depth for subsequent memory updates. This demonstrates that our framework is
not tied to a specific appearance generator: the explicit 4D memory and generate--recover--update cycle can serve as a general consistency layer for existing methods. Consequently, improvements in the underlying appearance generator can be incorporated to further enhance visual quality without redesigning the overall world-building pipeline.

\begin{figure*}[t]
    \centering
    \includegraphics[width=1\linewidth]{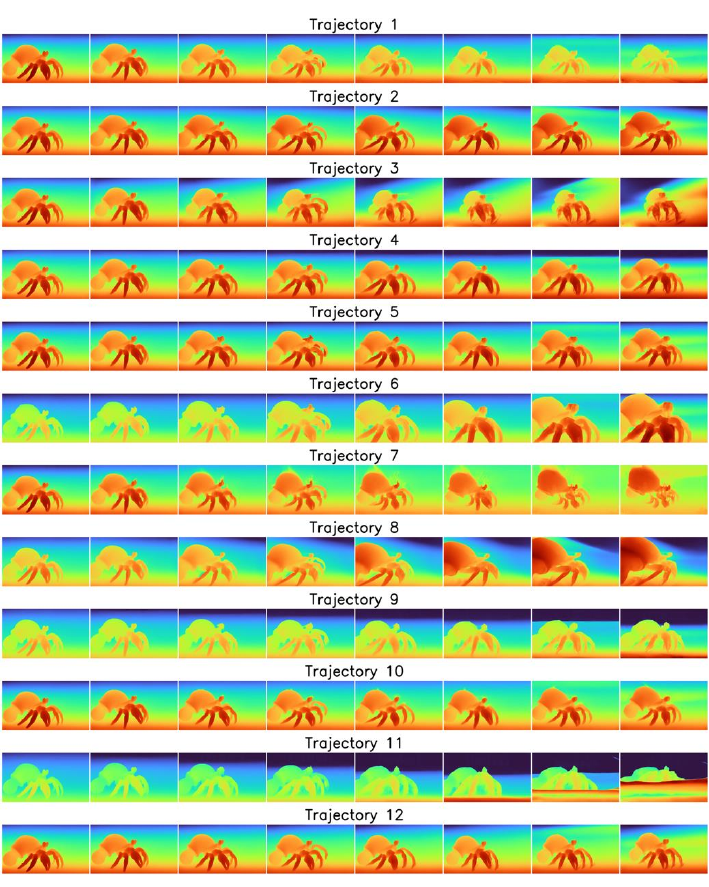}
    \caption{\textbf{Compatibility with GEN3C (Case 1, depth).} Our depth completion module recovers temporally coherent geometry from the GEN3C-generated videos for subsequent 4D memory updates.}
    \label{fig:ablation_comptable_fig1_depth}
\end{figure*}

\begin{figure*}[t]
    \centering
    \includegraphics[width=1\linewidth]{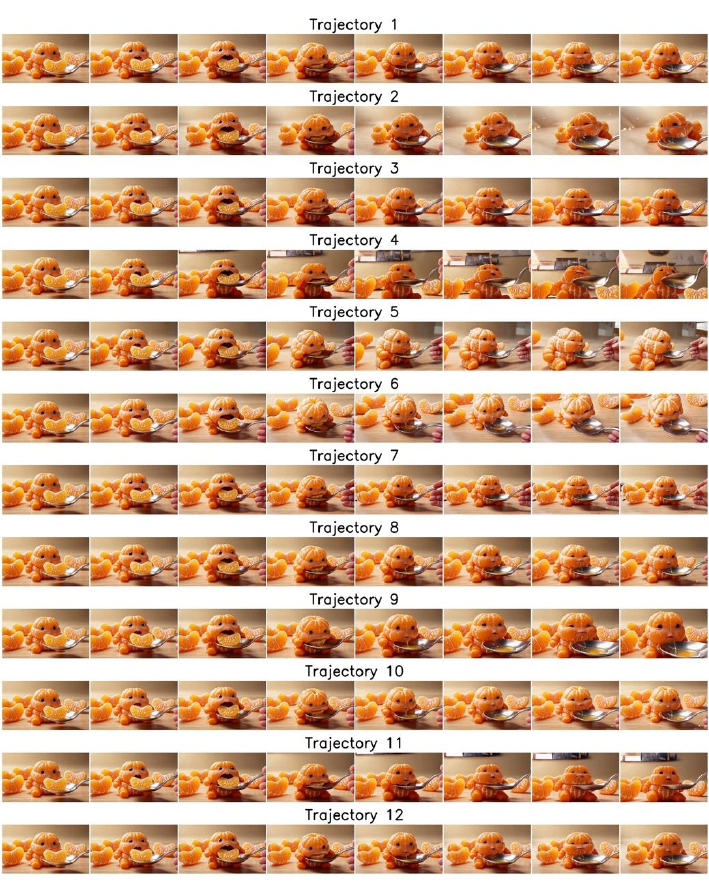}
    \caption{\textbf{Compatibility with GEN3C (Case 2, RGB).} The integrated framework preserves the dynamic subject and scene appearance across twelve target trajectories.}
    \label{fig:ablation_comptable_fig2_rgb}
\end{figure*}

\begin{figure*}[t]
    \centering
    \includegraphics[width=1\linewidth]{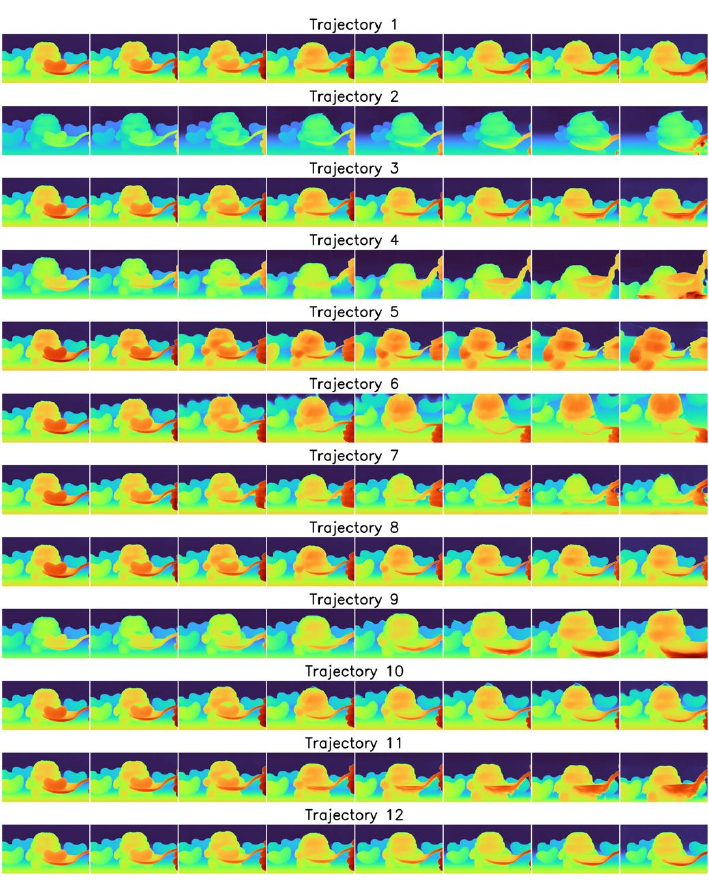}
    \caption{\textbf{Compatibility with GEN3C (Case 2, depth).} Consistent depth recovery across trajectories demonstrates that the remaining Demiurge pipeline can operate with an existing appearance generator.}
    \label{fig:ablation_comptable_fig2_depth}
\end{figure*}

\section{Additional visualizations}
\label{appendix: Additional visualizations}

\begin{figure*}[ht]
    \centering
    \includegraphics[width=0.85\linewidth]{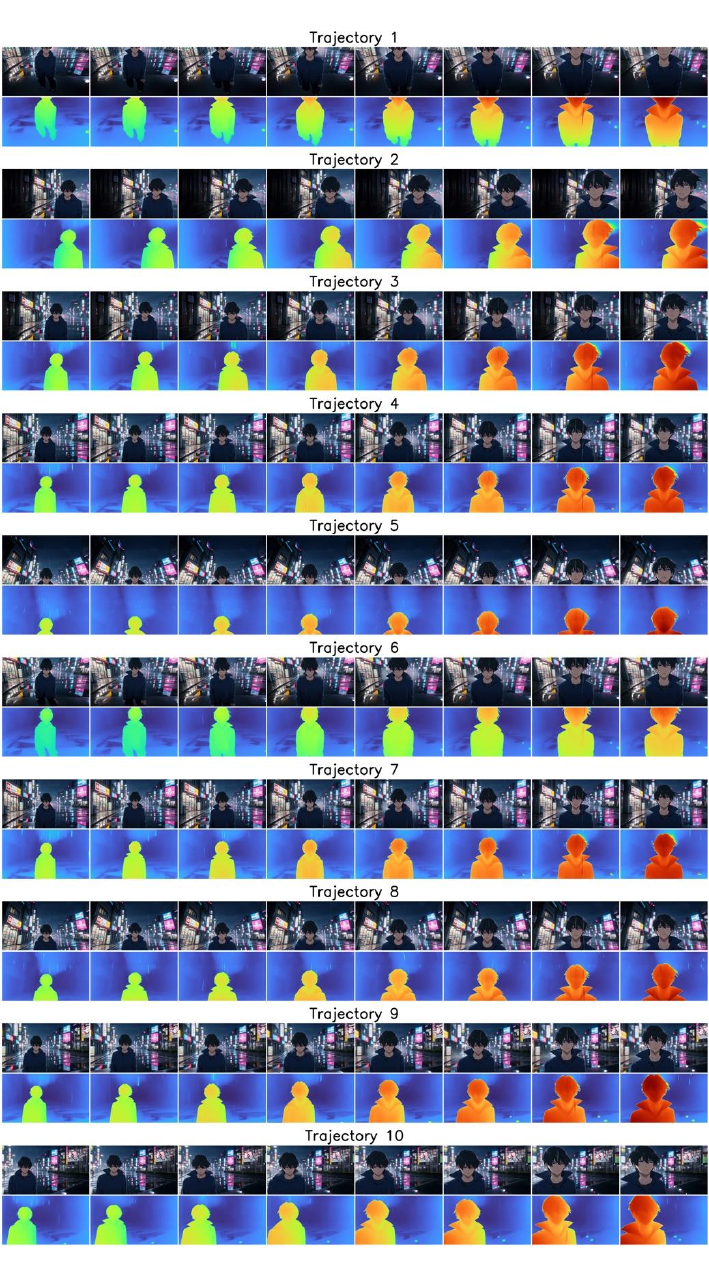}
    \caption{\textbf{Additional visualization (Case 1).} Generated RGB frames and recovered depths across ten target camera trajectories. Each row corresponds to one trajectory, with sampled frames arranged chronologically from left to right.}
    \label{fig:appendix_more_visulization_fig2}
\end{figure*}

We present additional results on diverse indoor and outdoor dynamic scenes in Figs.~\ref{fig:appendix_more_visulization_fig2}--%
\ref{fig:appendix_more_visulization_fig14}. In each figure, rows correspond to different target camera trajectories and columns show temporally aligned samples arranged from left to right. The generated RGB frames and their
recovered depths are shown in the upper and lower rows, respectively. Across different viewpoints and time steps, Demiurge consistently preserves scene appearance, dynamic content, and geometric structure, further demonstrating
its ability to construct coherent multi-trajectory observations for dynamic scenes.

\begin{figure*}[ht]
    \centering
    \includegraphics[width=0.85\linewidth]{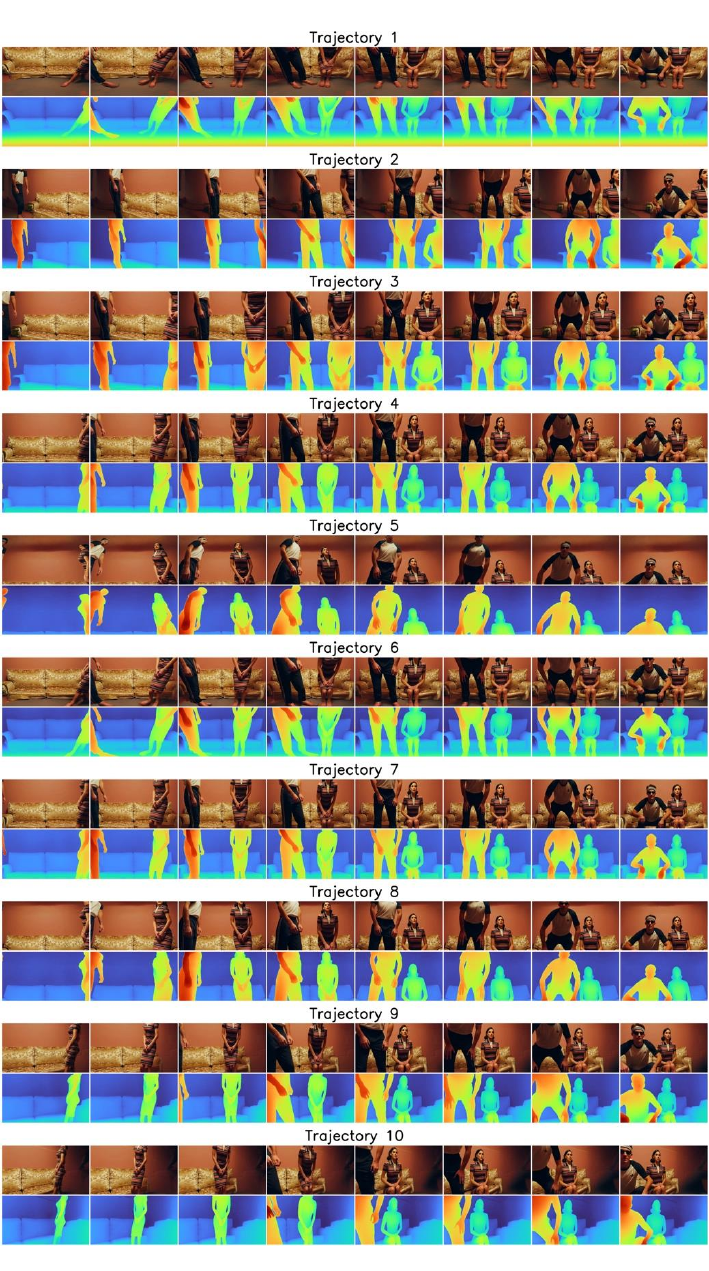}
    \caption{\textbf{Additional visualization (Case 2).} Generated RGB frames and recovered depths across ten target camera trajectories. Each row corresponds to one trajectory, with sampled frames arranged chronologically from left to right.}
    \label{fig:appendix_more_visulization_fig4}
\end{figure*}

\begin{figure*}[ht]
    \centering
    \includegraphics[width=0.85\linewidth]{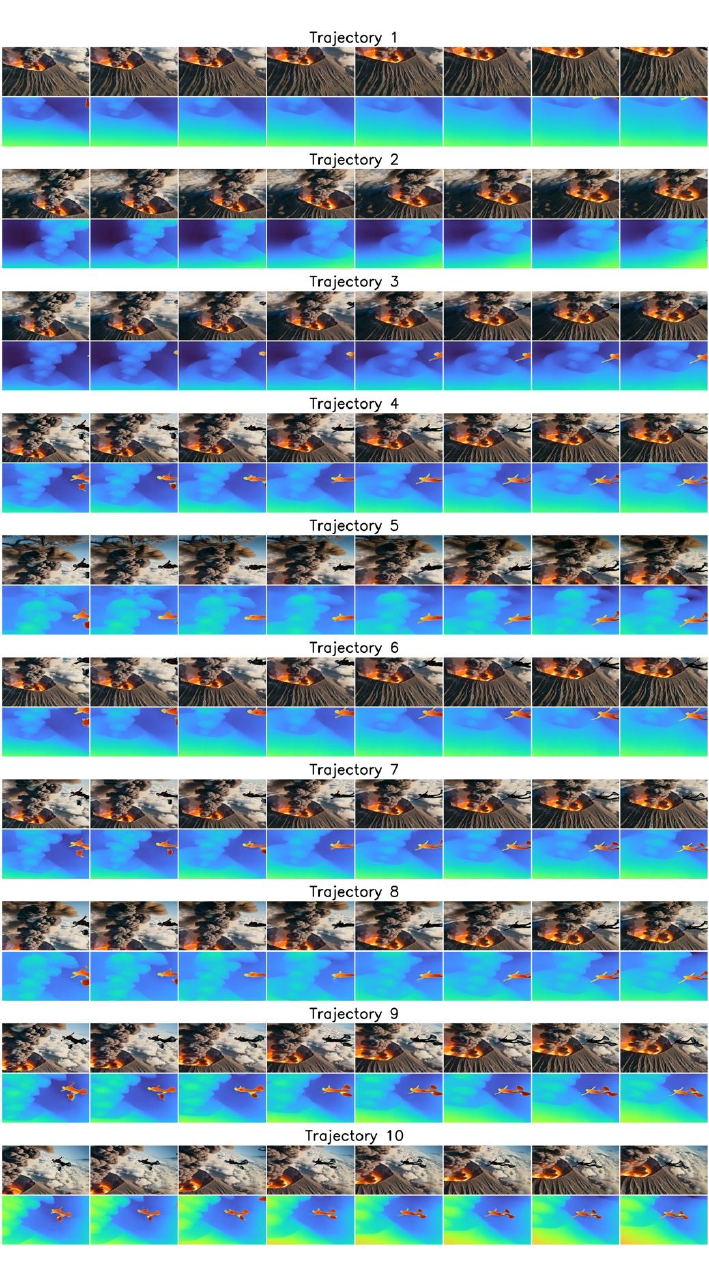}
    \caption{\textbf{Additional visualization (Case 3).} Generated RGB frames and recovered depths across ten target camera trajectories. Each row corresponds to one trajectory, with sampled frames arranged chronologically from left to right.}
    \label{fig:appendix_more_visulization_fig5}
\end{figure*}

\begin{figure*}[ht]
    \centering
    \includegraphics[width=0.85\linewidth]{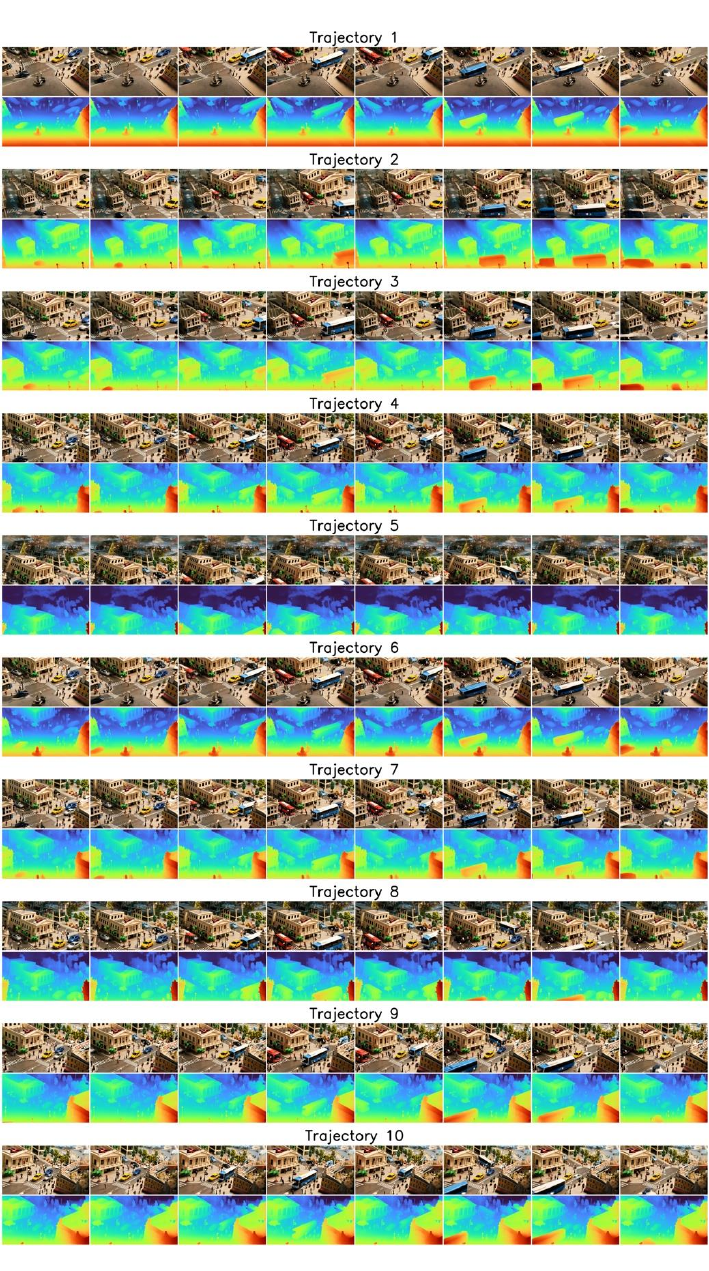}
    \caption{\textbf{Additional visualization (Case 4).} Generated RGB frames and recovered depths across ten target camera trajectories. Each row corresponds to one trajectory, with sampled frames arranged chronologically from left to right.}
    \label{fig:appendix_more_visulization_fig6}
\end{figure*}

\begin{figure*}[ht]
    \centering
    \includegraphics[width=0.85\linewidth]{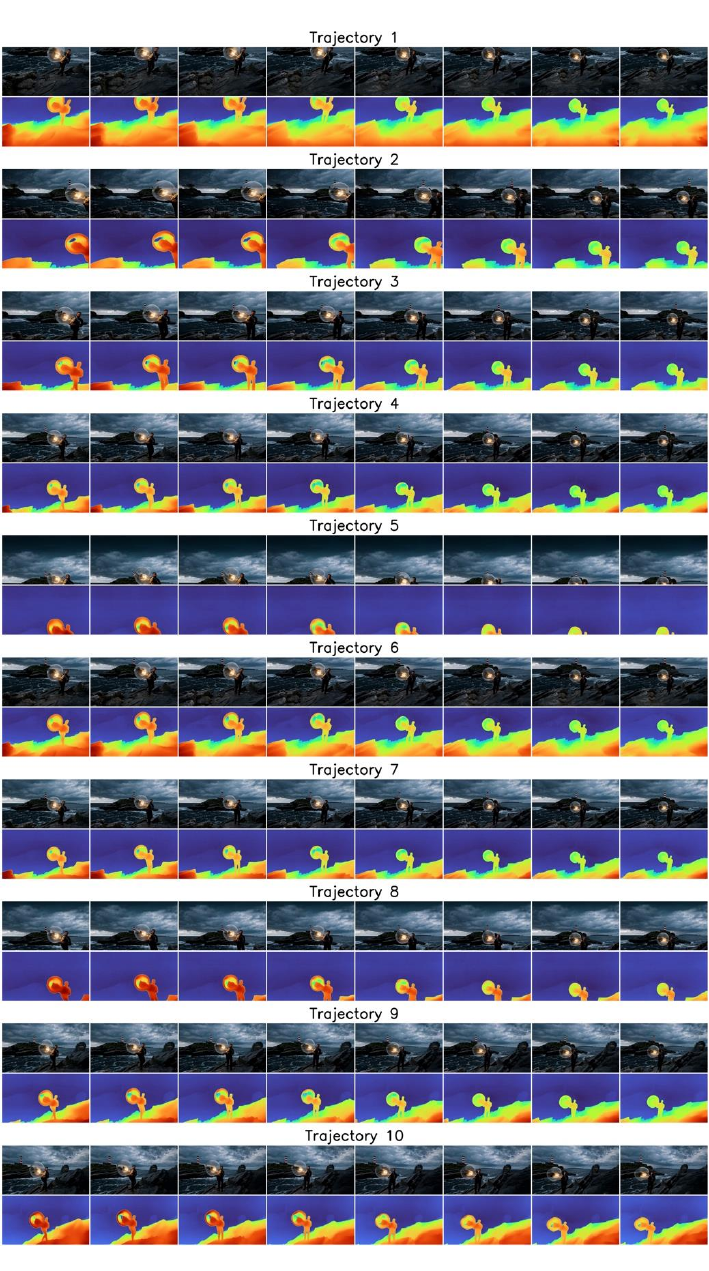}
    \caption{\textbf{Additional visualization (Case 5).} Generated RGB frames and recovered depths across ten target camera trajectories. Each row corresponds to one trajectory, with sampled frames arranged chronologically from left to right.}
    \label{fig:appendix_more_visulization_fig7}
\end{figure*}

\begin{figure*}[ht]
    \centering
    \includegraphics[width=0.85\linewidth]{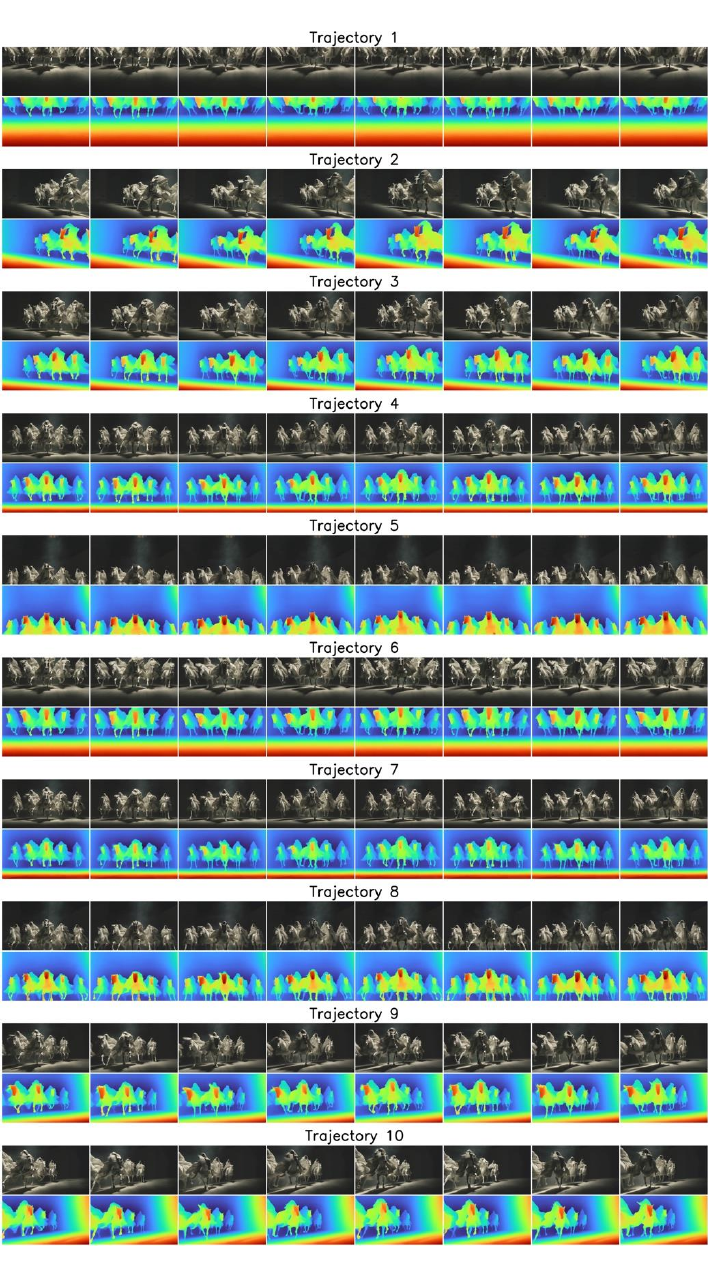}
    \caption{\textbf{Additional visualization (Case 6).} Generated RGB frames and recovered depths across ten target camera trajectories. Each row corresponds to one trajectory, with sampled frames arranged chronologically from left to right.}
    \label{fig:appendix_more_visulization_fig8}
\end{figure*}

\begin{figure*}[ht]
    \centering
    \includegraphics[width=0.85\linewidth]{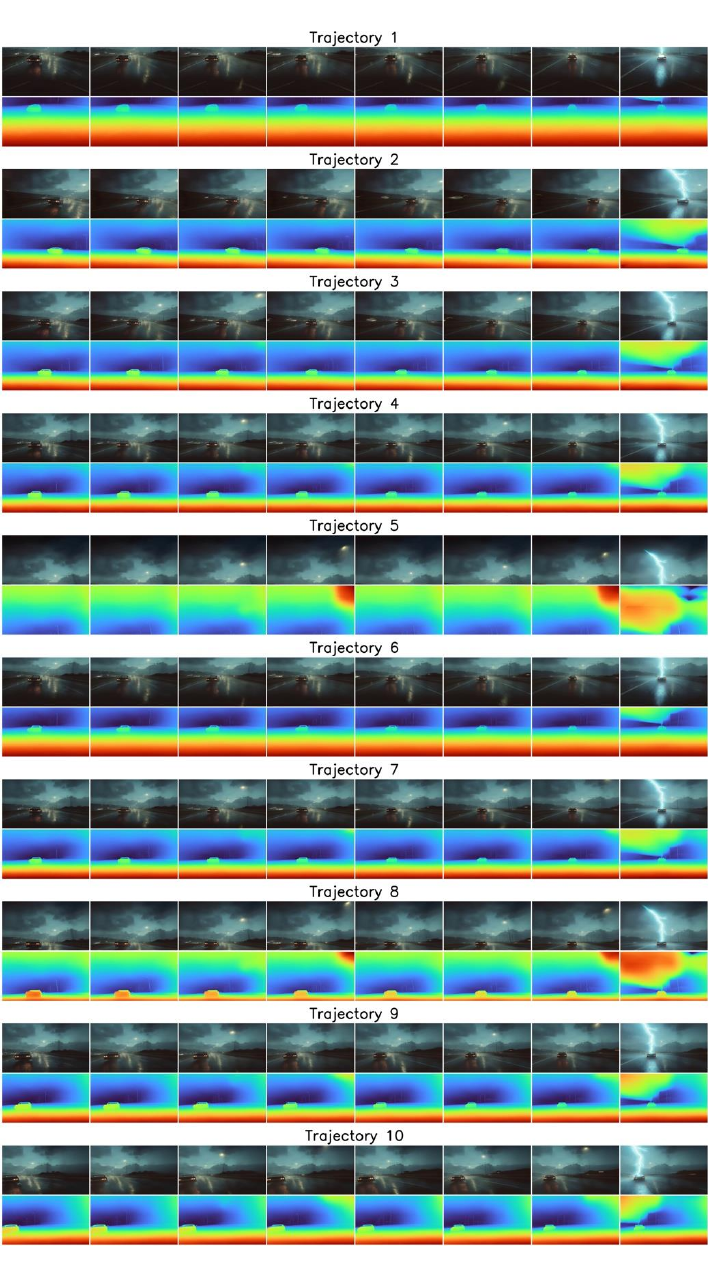}
    \caption{\textbf{Additional visualization (Case 7).} Generated RGB frames and recovered depths across ten target camera trajectories. Each row corresponds to one trajectory, with sampled frames arranged chronologically from left to right.}
    \label{fig:appendix_more_visulization_fig10}
\end{figure*}

\begin{figure*}[ht]
    \centering
    \includegraphics[width=0.85\linewidth]{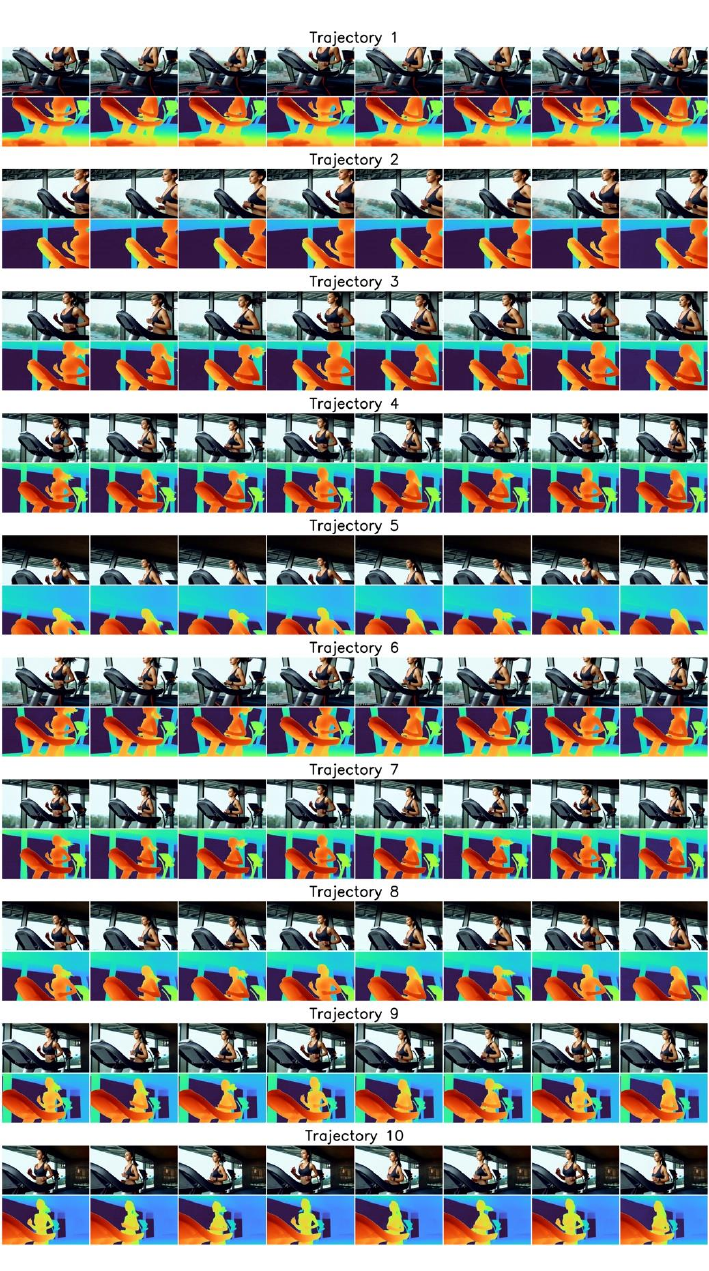}
    \caption{\textbf{Additional visualization (Case 8).} Generated RGB frames and recovered depths across ten target camera trajectories. Each row corresponds to one trajectory, with sampled frames arranged chronologically from left to right.}
    \label{fig:appendix_more_visulization_fig12}
\end{figure*}

\begin{figure*}[ht]
    \centering
    \includegraphics[width=0.85\linewidth]{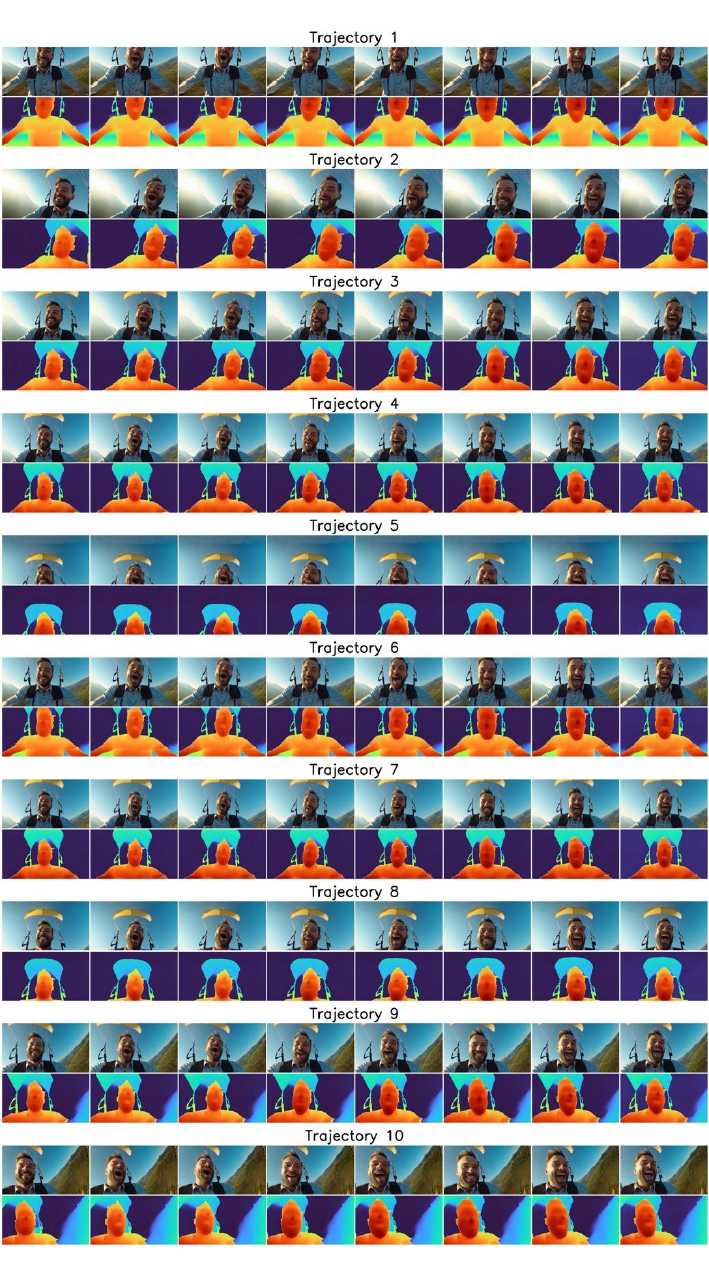}
    \caption{\textbf{Additional visualization (Case 9).} Generated RGB frames and recovered depths across ten target camera trajectories. Each row corresponds to one trajectory, with sampled frames arranged chronologically from left to right.}
    \label{fig:appendix_more_visulization_fig14}
\end{figure*}

\section{Runtime Analysis}
\label{appendix:runtime_Analysis}
Under our default inference setting, the appearance video diffusion, video
depth completion, and optional geometry-aware refinement modules each require
approximately 1.5 minutes to process an 81-frame video at a resolution of
$480\times832$. With refinement enabled, generating one target trajectory takes
approximately 5 minutes end to end. Since target trajectories are processed
autoregressively, the total inference time scales approximately linearly with
their number, i.e., about $5N$ minutes for $N$ trajectories. Refinement can be
disabled when faster inference is preferred, reducing the diffusion inference
time to approximately 3 minutes per trajectory. Although the current implementation is not optimized for speed, Demiurge remains stable over substantially longer generation horizons than those used
in our benchmark. In particular, we have generated up to 100 consecutive camera trajectories without noticeable error accumulation, demonstrating the long-horizon scalability of the evolving 4D memory.

\section{Limitations and future work}
\label{Limitations and future work}

The primary limitation of Demiurge is its inference latency. The current pipeline comprises several sequentially coupled modules, including appearance generation, video depth completion, optional geometry-aware refinement, and 4D memory update. Their intermediate RGB, depth, and geometric representations must be exchanged and synchronized before the next autoregressive step, preventing straightforward parallel execution. Consequently, the full pipeline requires approximately five minutes to generate one target-trajectory video, which limits its applicability to interactive scenarios.

An important direction for future work is to decouple latency-sensitive RGB generation from background geometry recovery and memory maintenance through an asynchronous cloud--edge architecture. Since users only consume the final RGB video, a fast path could immediately query the latest 4D cache and relevant video memories to synthesize the requested view, while a slower background path
performs depth completion, optional refinement, back-projection, and memory
updates asynchronously. The fast path would support responsive rendering within the currently cached region, whereas the background path would continuously consolidate newly generated observations and expand the renderable world. Such a fast--slow design could substantially reduce perceived latency without discarding the geometric consistency provided by the complete generate--recover--update cycle.

Moreover, developing a more efficient representation of the evolving 4D memory
is equally important. Our current time indexed point-cloud representation prioritizes explicitness and geometric interpretability, but repeated
observations may introduce redundant points and increase storage, retrieval, and rendering costs as the number of trajectories grows. Future work could jointly align and compress historical observations into a hierarchical spatiotemporal representation, merging geometrically and temporally compatible content while preserving dynamic boundaries and newly revealed regions. Confidence-aware fusion, adaptive spatial discretization, and compact 4D primitives may further reduce memory overhead and accelerate memory queries, enabling longer and larger-scale world expansion.

\stopcontents

\end{document}